\documentclass{zlab}

\newcommand{\templateoption}{option1}

\usepackage[T1]{fontenc}
\usepackage{tgpagella}
\usepackage{mathpazo}
\usepackage{inconsolata}
\usepackage[strings]{underscore}
\usepackage{xspace}
\usepackage{multirow} 

\usepackage{graphicx,amsmath,amssymb,hyperref}
\usepackage{subcaption}
\usepackage{longtable}
\usepackage{listings}
\usepackage{tcolorbox}
\tcbuselibrary{listings,skins,breakable,theorems}
\usepackage{adjustbox}
\usepackage[authoryear,round]{natbib}
\usepackage[table]{xcolor}
\usepackage{wrapfig}
\usepackage{placeins}
\usepackage{tikz}
\usepackage{pdfpages}
\usepackage{helvet}
\usepackage{etoc}
\usepackage[hang,flushmargin]{footmisc}

\definecolor{verocolor}{HTML}{1976D2}
\definecolor{linkred}{HTML}{ED1C24}
\definecolor{linkmagenta}{HTML}{C2185B}

\definecolor{reviewcolor}{HTML}{D81B60}

\newcommand{\model}{\textcolor{verocolor}{\textbf{Vero}}\xspace}
\newcommand{\modelQthreefive}{\textcolor{verocolor}{\textbf{Vero-Qwen35-9B}}\xspace}
\newcommand{\modelQthreeT}{\textcolor{verocolor}{\textbf{Vero-Qwen3T-8B}}\xspace}
\newcommand{\modelQthreeI}{\textcolor{verocolor}{\textbf{Vero-Qwen3I-8B}}\xspace}
\newcommand{\modelQtwofive}{\textcolor{verocolor}{\textbf{Vero-Qwen25-7B}}\xspace}
\newcommand{\modelMi}{\textcolor{verocolor}{\textbf{Vero-MiMo-7B}}\xspace}
\newcommand{\modelsuite}{\textcolor{verocolor}{\textbf{VeroEval}}\xspace}
\newcommand{\verodataset}{\textcolor{verocolor}{\textbf{Vero-600K}}\xspace}
\newcommand{\verodatasetlarge}{\textcolor{verocolor}{\textbf{Vero-1.6M}}\xspace}
\newcommand{\numdatasets}{59}

\setlist[itemize]{itemsep=0.1pt}

\hypersetup{
  colorlinks=true,
  urlcolor=verocolor,
  linkcolor=linkred,
  citecolor=verocolor,
  filecolor=black,
  pdfborder={0 0 0}
}

\definecolor{link}{HTML}{ED1C24}
\providecommand{\equationname}{Equation}
\providecommand{\sectionname}{Section}

\newcommand{\figref}[1]{%
  \figurename~\hyperref[#1]{\textcolor{link}{\ref*{#1}}}%
}
\newcommand{\tabref}[1]{%
  \tablename~\hyperref[#1]{\textcolor{link}{\ref*{#1}}}%
}
\newcommand{\eqrefc}[1]{%
  \equationname~\hyperref[#1]{\textcolor{link}{\ref*{#1}}}%
}
\newcommand{\secrefc}[1]{%
  \sectionname~\hyperref[#1]{\textcolor{link}{\ref*{#1}}}%
}
\newcommand{\appref}[1]{Appendix~\ref{#1}}

\newcommand{\iconfallback}[1]{\raisebox{0.1pt}{\scriptsize\texttt{#1}}\xspace}
\newcommand{\projecticon}{%
  \IfFileExists{figures/logos/project_page_icon.png}
    {\raisebox{-1.2pt}{\includegraphics[height=1.05em]{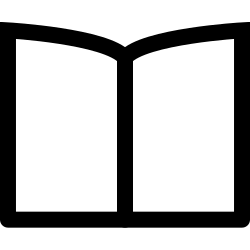}}\xspace}
    {\iconfallback{BK}}%
}
\newcommand{\github}{%
  \IfFileExists{figures/logos/github_logo.pdf}
    {\raisebox{-1.3pt}{\includegraphics[height=1.05em]{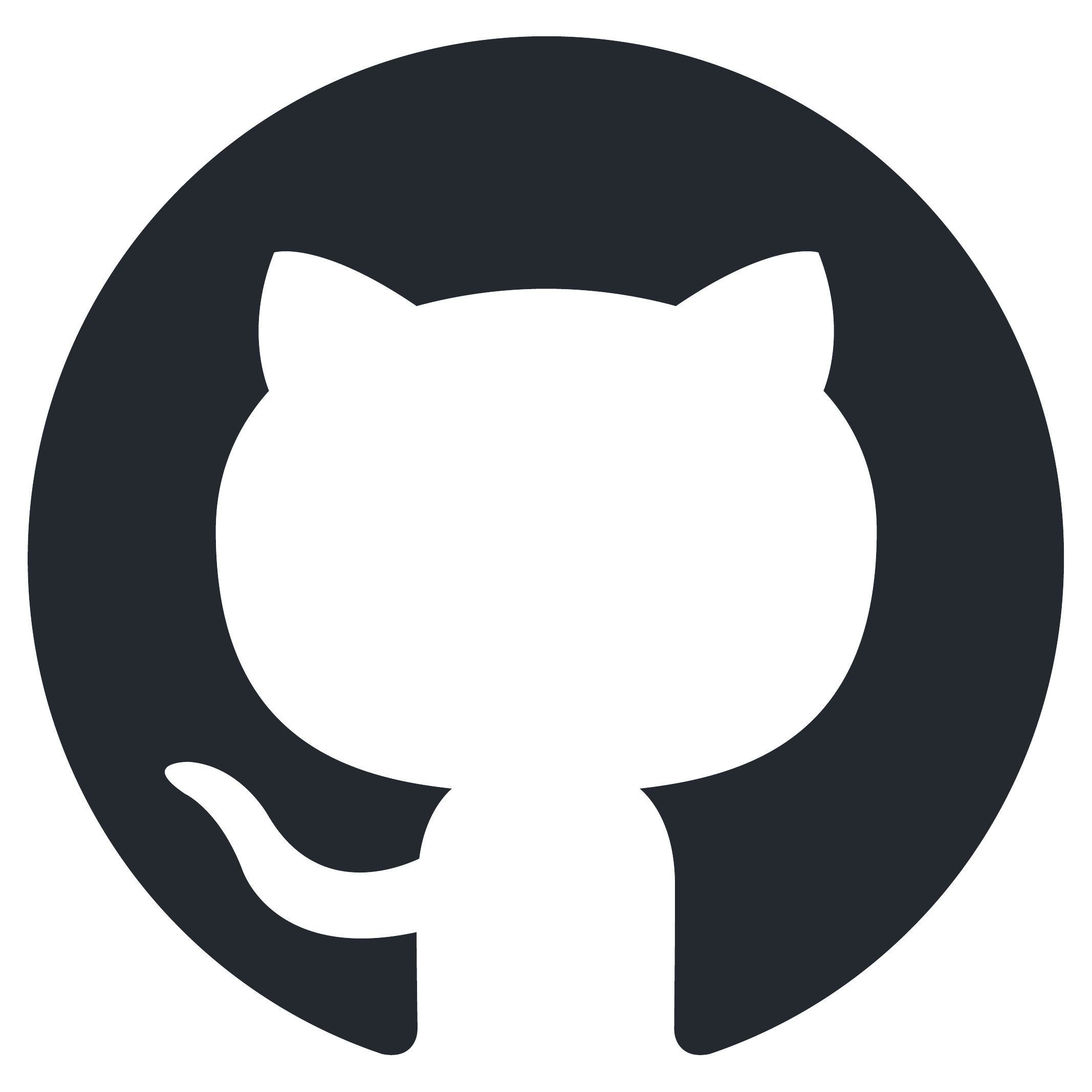}}\xspace}
    {\iconfallback{GH}}%
}
\newcommand{\hf}{%
  \IfFileExists{figures/logos/huggingface_logo.pdf}
    {\raisebox{-1.3pt}{\includegraphics[height=1.05em]{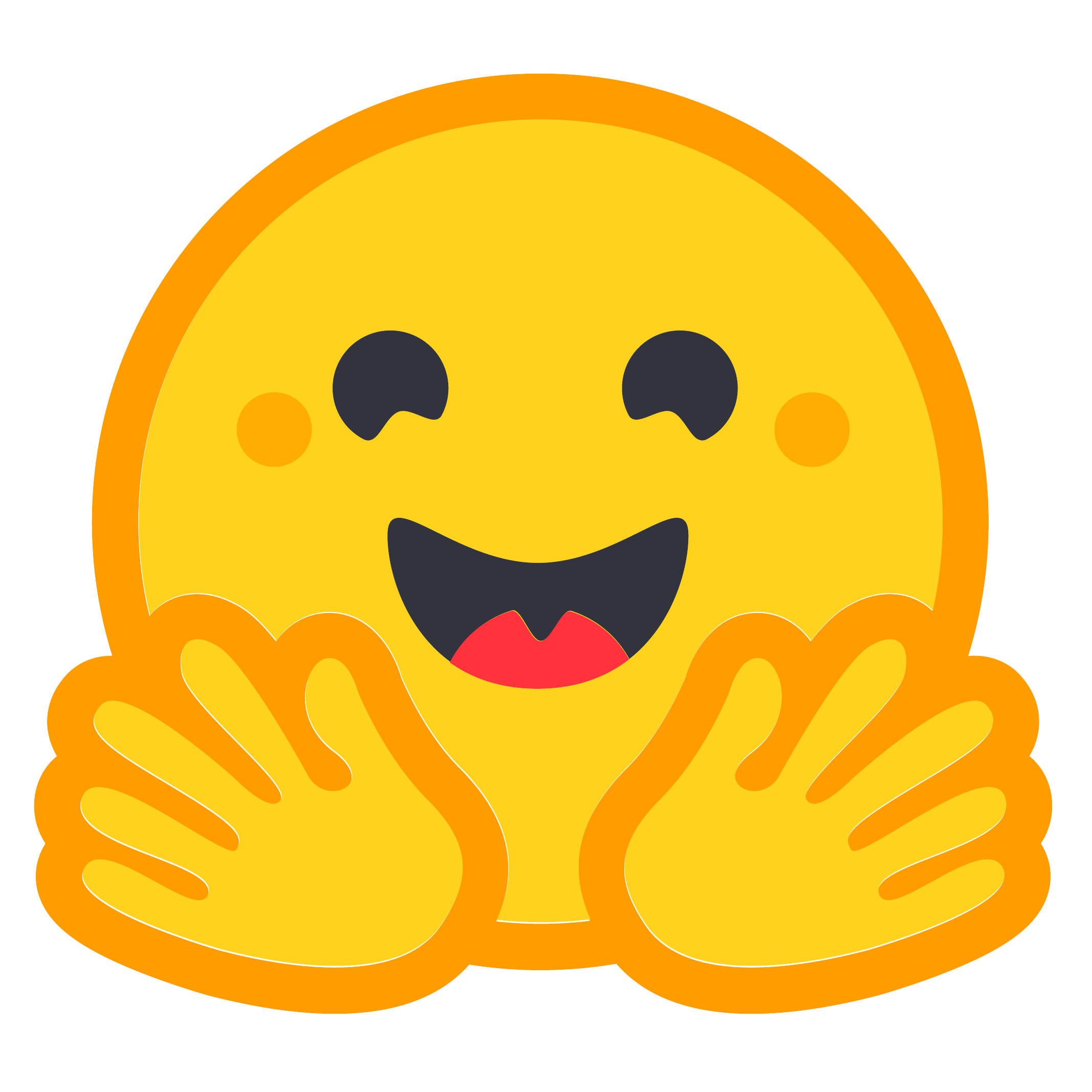}}\xspace}
    {\iconfallback{HF}}%
}
\newcommand{\dbicon}{%
  \IfFileExists{figures/logos/data_icon.pdf}
    {\raisebox{-1.3pt}{\includegraphics[height=1.05em]{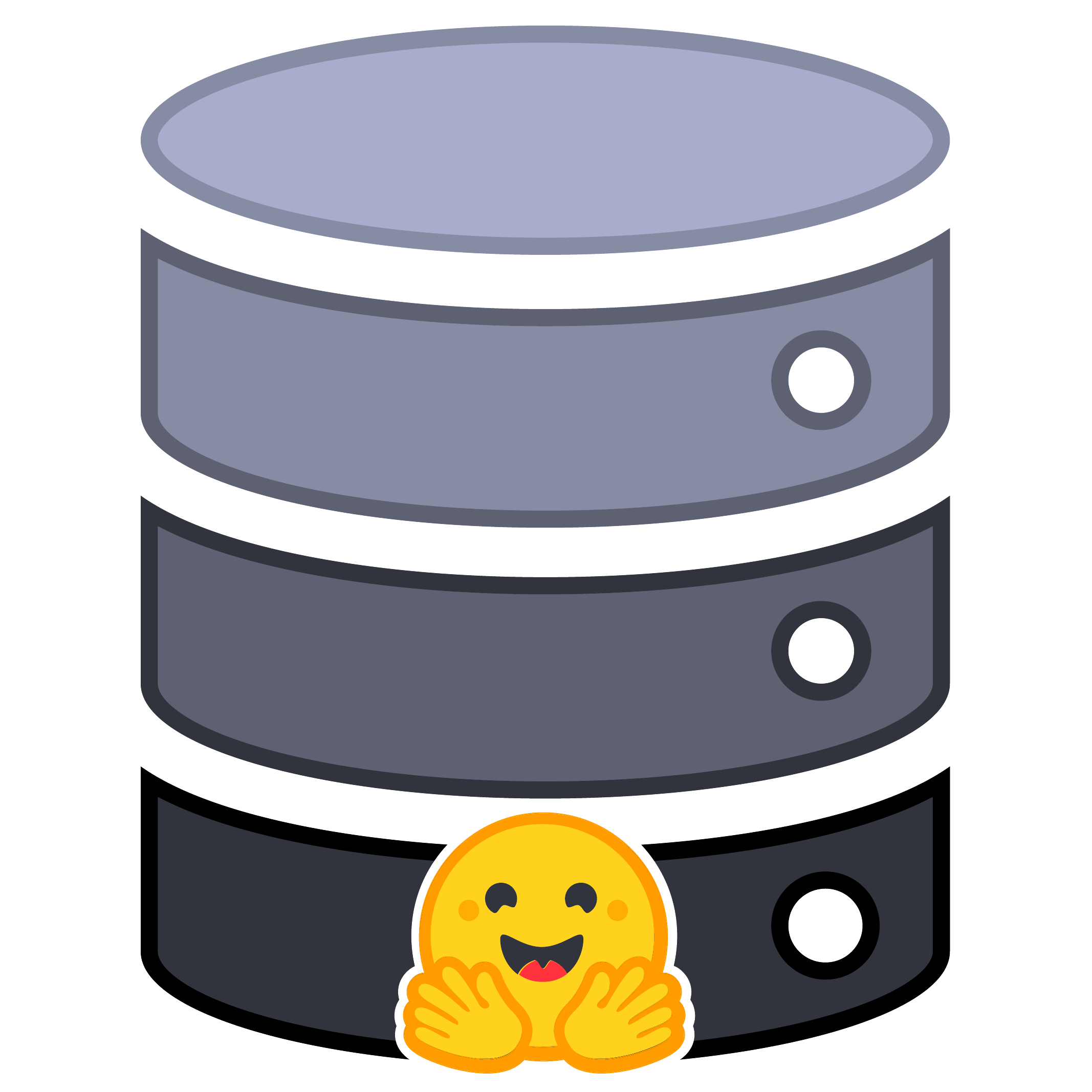}}\xspace}
    {\iconfallback{DB}}%
}
\newcommand{\worldwideweb}{\projecticon}

\makeatletter
\fancypagestyle{firststyle}{

  \fancyhead[L]{}
  \fancyhead[C]{}
  \fancyhead[R]{}
  \fancyfoot[L]{{\fontsize{8.5}{10}\selectfont \the\correspondingauthor}}
  \fancyfoot[R]{}
}
\makeatother

\DeclareTextFontCommand{\textbf}{\bfseries}

\title{%
  {\fontsize{16}{22}\selectfont \model{}}{\fontsize{14}{20}\selectfont : An Open RL Recipe for General Visual Reasoning\par}
  \vspace{-0.5cm}
}

\author{
    \vspace{.2cm}
    \parbox{\textwidth}{\centering
        Gabriel Sarch$^{*}$ \quad
        Linrong Cai$^{*}$ \quad
        Qunzhong Wang \quad
        Haoyang Wu \\
        \vspace{-0.5em}\Authfont
        Danqi Chen \quad
        Zhuang Liu$^{\dagger}$
    }
    \\
    \vspace{.0cm}
    {\normalfont\fontsize{11}{15}\selectfont {Princeton University}\hspace{.1cm}}
        \vspace{.3cm}
}

\correspondingauthor{\shortstack[l]{\, $^{*}$ Project Leads\, \hspace{1em} $^{\dagger}$ Corresponding Author}}

\makeatletter
\renewcommand{\abscontent}{}%
\makeatother

\newenvironment{abstractblock}{%
  {\centering\large\bfseries Abstract\par}
  \vspace{0.1em}
  \begin{list}{}{%
      \setlength{\leftmargin}{2em}
      \setlength{\rightmargin}{2em}
      \setlength{\topsep}{0pt}
      \setlength{\parsep}{0pt}
  }
  \item[]
}{%
  \end{list}
  \par\normalfont\vspace{1em}
}

\begin{document}

\begingroup
\makeatletter
\let\raggedright\centering
\makeatother

\maketitle
\endgroup

\newcommand{\abstractcontent}{
What does it take to build a visual reasoner that works across charts, science, spatial understanding, and open-ended tasks? The strongest vision-language models (VLMs) suggest that broad visual reasoning is within reach, yet their closed data and reinforcement learning (RL) pipelines make their gains difficult to study, reproduce, or extend. We introduce \model{}, a family of \emph{fully open} VLMs that match or exceed existing open-weight models across diverse visual reasoning tasks. We scale RL data and rewards across six broad task categories, constructing \verodataset, a 600K-sample dataset from \numdatasets{} datasets, and designing task-routed rewards that handle heterogeneous answers. Across \modelsuite{}, our 30-benchmark suite, \verodataset{} outperforms existing RL datasets under controlled comparisons. Applied to five starting models, \model{} variants gain 2.9--5.4 points on average over their initial models. Notably, \modelQthreeI{}, trained on the Instruct model, surpasses Qwen3-VL-8B-Thinking by 3.8 points on average without additional distillation. Systematic ablations reveal that different task categories elicit distinct reasoning patterns and that broad gains depend on learning them jointly rather than in isolation. All data, code, and models are publicly available.
}

\newcommand{\teaserfigure}[2]{
    \begin{figure}[#1]
        \vspace{-0.2cm}
        \centering
        \includegraphics[width=#2]{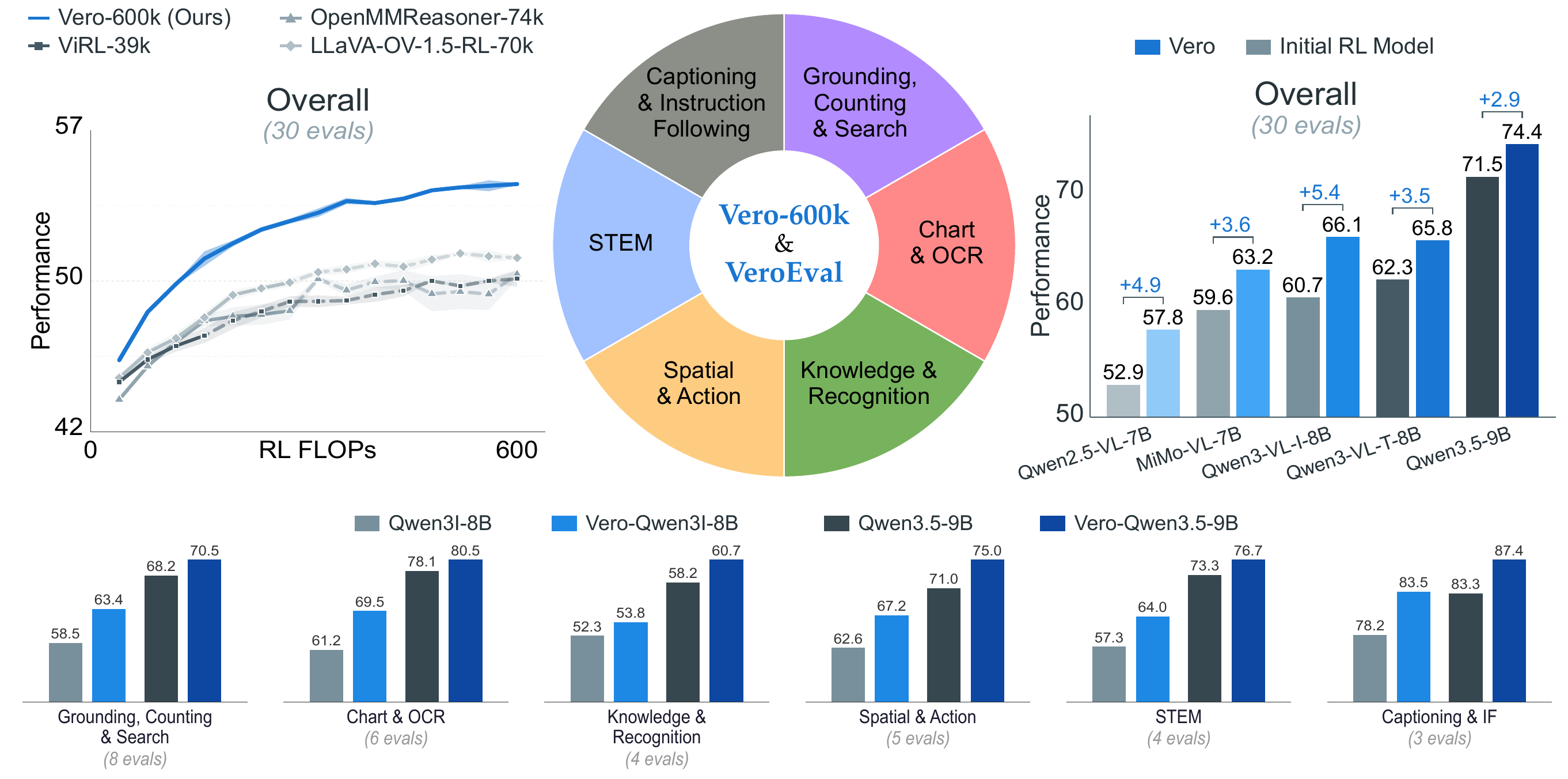}
        \caption{\textbf{\model{} achieves state-of-the-art performance across six task categories using a fully open RL recipe.} \textbf{Top left:} Training curves versus RL FLOPs for \verodataset compared with existing open RL datasets, all finetuned from Qwen2.5-VL-7B-Instruct; dashed lines indicate training beyond one epoch. Reported is mean $\pm$ sem (3 seeds). \textbf{Top center:} Summary of the broad task categories targeted in \verodataset and \modelsuite{}. \textbf{Top right:} Overall performance of \model{} (Avg@5) against each initial RL model across the 30 \modelsuite{} benchmarks. \textbf{Bottom row:} Same as top right, but displaying per-category scores for \model{} trained on Qwen3-VL-8B-Instruct and Qwen3.5-9B.
        }
        \vspace{-0.4cm}
        \label{fig:teaser}
    \end{figure}%
}

\newcommand{\linkstablestyleone}{%
    \begin{center}
        \small
        \hf\,\href{https://huggingface.co/collections/zlab-princeton/vero}{\textbf{Models}}
        \qquad
        \dbicon\,\href{https://huggingface.co/collections/zlab-princeton/vero}{\textbf{Data: 600K \& 1.6M}}
        \qquad
        \github\,\href{https://github.com/zlab-princeton/vero}{\textbf{Code}}
        \qquad
        \worldwideweb\,\href{https://vero-reasoning.github.io}{\textbf{Project Page}}
    \end{center}%
}
    
\newcommand{\linkstablestyletwo}{%
    \noindent\small
    \textbf{Website:} \url{https://vero-reasoning.github.io}\\
    \textbf{Code:}    \url{https://github.com/zlab-princeton/vero}\\
    \textbf{Data:}    \url{https://huggingface.co/collections/zlab-princeton/Vero}\\
    \textbf{Models:}  \url{https://huggingface.co/collections/zlab-princeton/Vero}
}

\makeatletter
\ifthenelse{\equal{\templateoption}{option1}}{
    \vspace{-1.5cm}
    \linkstablestyleone
    \vspace{0.5cm}
    \teaserfigure{!ht}{\textwidth}
    \vspace{0.1cm}
    \begin{abstractblock}
    \vspace{-0.7em}
    \abstractcontent
    \end{abstractblock}
    \newpage
}{
    \ifthenelse{\equal{\templateoption}{option2}}{
        \vspace{-1.00cm}
        \begin{abstractblock}
        \abstractcontent
        \end{abstractblock}
        \vspace{0.0cm}
        \teaserfigure{!bh}{\textwidth}
        \newpage
    }{
        \begin{abstractblock}
        \abstractcontent
        \end{abstractblock}
        \vspace{0.5cm}
        \vspace{-0.5cm}

    }
}
\makeatother

\section{Introduction}
\label{sec:intro}

Vision-language models (VLMs) are increasingly expected to reason across a wide range of visual tasks, from chart and scientific interpretation to spatial understanding and open-ended questions. Reinforcement learning (RL) has emerged as a key driver of this progress, with methods such as PPO~\citep{schulman2017proximal} and GRPO~\citep{grpo} enabling models to learn from their own generations through reward signals. Recent multimodal models such as GPT-5~\citep{gpt5}, Qwen3-VL~\citep{qwen3vl}, and Kimi K2.5~\citep{kimik25} demonstrate that RL drives substantial improvement in multimodal reasoning.

Yet the strongest existing visual reasoning models are products of proprietary RL pipelines with non-public data and undisclosed reward designs. Models such as Qwen3-VL~\citep{qwen3vl} release weights and are widely adopted, but do not release RL training code or datasets. Accompanying technical reports often omit detailed ablations of design choices, making it difficult to systematically study what drives performance. Meanwhile, fully open efforts such as OpenMMReasoner~\citep{openmmreasoner} and VL-Rethinker~\citep{vlrethinker} focus primarily on visual math, covering only a narrow subset of visual tasks. However, as we show in Sections~\ref{sec:cross_gen} and \ref{sec:behavioral-analysis}, training on a single task category does not generalize to other visual capabilities, in both task performance and chain-of-thought behavior. 
More broadly, applying RL across heterogeneous visual reasoning tasks is challenging, as diverse task mixtures induce interference, weak transfer, and optimization imbalance unless the training distribution and rewards are carefully designed~\citep{teh2017distral,schaul2019ray,hessel2019multi}.
This leaves a central question: \emph{what does it take to train a broadly capable visual reasoner?}

\begin{figure*}[t]
    \centering
    \includegraphics[width=1.0\linewidth]{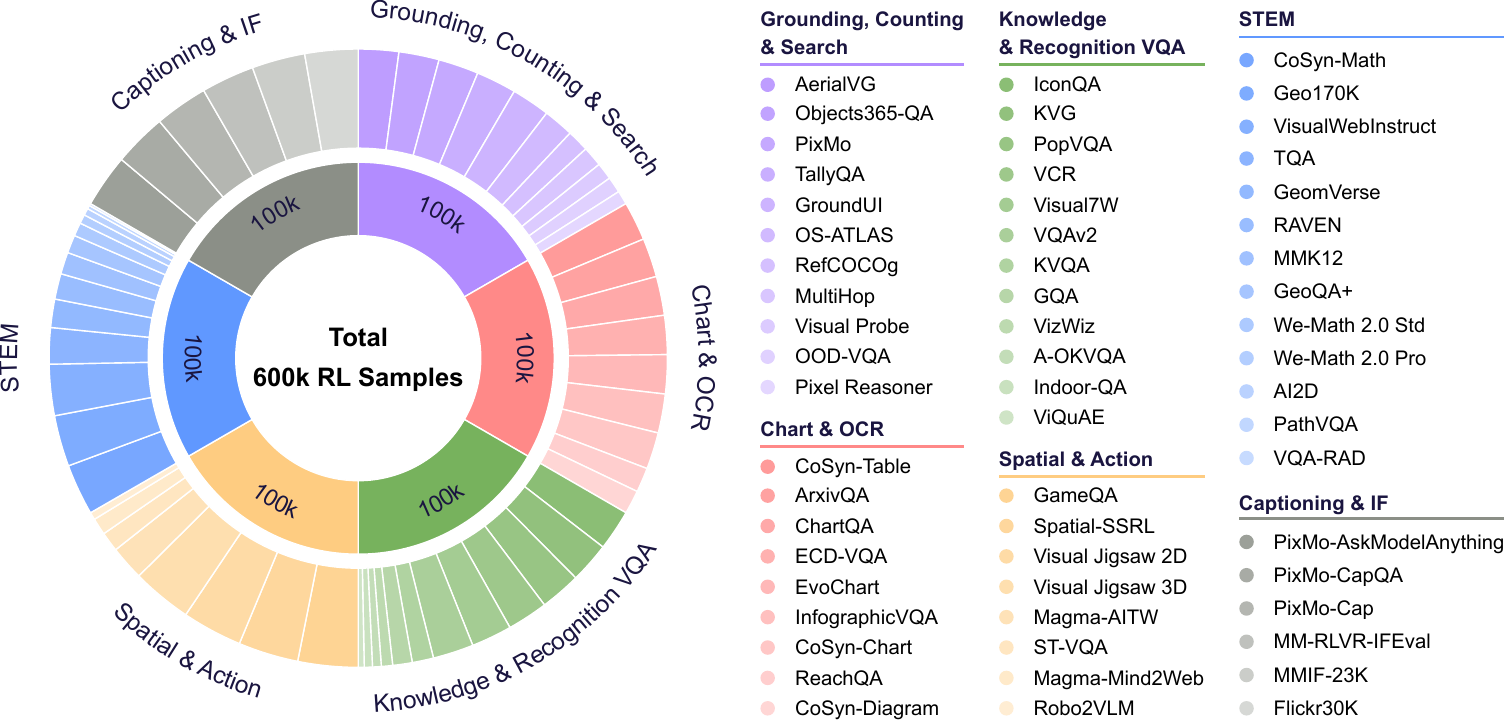}
    \caption{\textbf{Composition of \verodataset.} The inner ring shows six task categories, each allocated 100K samples (600K total), and the outer ring shows their \numdatasets{} constituent datasets. The categories represent real-world use cases and cover distinct visual reasoning capabilities (Sections~\ref{sec:cross_gen} and~\ref{sec:behavioral-analysis}). Categories are uniformly sampled to balance learning across tasks.}
    \label{fig:training-datasets}
\end{figure*}

We show that a single-stage RL recipe with diverse and high-quality data suffices. We introduce \model{}, a family of fully open VLMs trained with RL on top of existing models to perform strongly across diverse visual tasks. Our recipe centers on careful dataset selection, sample filtering, and task-balanced mixing: we build \verodataset, a 600K-sample training set from \numdatasets{} datasets spanning six core task categories (Chart \& OCR; STEM; Spatial \& Action; Knowledge \& Recognition; Grounding, Counting \& Search; and Captioning \& Instruction Following), and pair it with task-routed reward functions.\footnote{We also release \verodatasetlarge{}, a 1.6M extension built with the same process, while all experiments use \verodataset{} unless specified.} No additional warm start, no staged RL, and no proprietary data. Alongside the training data, we assemble \modelsuite{}, a comprehensive evaluation suite of 30 benchmarks spanning all six categories. Figure~\ref{fig:training-datasets} summarizes the composition of \verodataset.

Through systematic ablations of dataset selection, sample filtering, mixture strategies, and reward design, we find that data diversity is the critical ingredient. Different task categories elicit qualitatively distinct reasoning patterns that transfer poorly in isolation: for example, STEM tasks trigger elevated backtracking while grounding tasks suppress introspective behaviors in favor of directed visual search. Producing a generally capable model therefore requires broad task coverage, so that the model learns these distinct reasoning patterns jointly rather than in isolation. We additionally find that (1) uniform mixture weighting across task categories outperforms schemes based on accuracy, reasoning length, or image size; (2) multi-task training necessitates an expressive, task-routed reward design; and (3) including open-ended tasks is necessary to preserve visual chat ability during RL.

\model{} achieves strong overall performance across model families (Figure~\ref{fig:teaser}). Training on six different initial models yields consistent improvements, with gains of +2.9 to +5.4 points over five post-trained models, averaged over 30 benchmarks. Applied directly to the pretrained-only Qwen3.5-9B-Base, our recipe yields +12.9 points, reaching 73.0 overall, second only to \modelQthreefive{}, without any SFT or distillation warm start.
\modelQthreefive{}, trained from Qwen35-9B, reaches 74.4 overall and improves over its initial model by +2.9, with gains on 25 of 30 benchmarks and all six category averages. Among 8B models, \modelQthreeI{} reaches 66.1 overall and outperforms Qwen3-VL-8B-Thinking by +3.8 overall and on 25 of 30 benchmarks, while \modelQthreeT{} improves over Qwen3-VL-8B-Thinking by +3.5 overall and on 22 of 30 benchmarks. Compared with distillation warm start baselines, \modelQtwofive{} exceeds OpenMMReasoner-7B by +5.6 overall despite OpenMMReasoner using a 874K example teacher distillation warm start. We release all data, code, and models to facilitate future research.

\section{Related Work}
\label{sec:related}

\paragraph{Vision-language models.}
Vision-language models excel on multimodal tasks, including proprietary systems such as GPT-5~\citep{gpt5} and Gemini~\citep{team2023gemini,team2024gemini,comanici2025gemini}, open-weight families such as  Qwen~\citep{qwen25vl,qwen3vl}, GLM~\citep{glmv}, and Kimi~\citep{kimivl}, and fully open releases of data, code, and weights such as Molmo~\citep{deitke_molmo_2024,molmo2} and LLaVA~\citep{liu2023visual,an2025llava}. These models are expected to handle a wide range of tasks. 
While little is publicly known about proprietary model post-training, recent open-weight models have explored techniques such as RL with curriculum sampling~\citep{glmv} and mixed on-policy RL~\citep{mimovl}, yet the factors that drive their performance across diverse tasks remain unclear. Our work targets this gap by providing a fully open multi-domain RL recipe for general visual understanding.

\paragraph{Reasoning and thinking for VLMs.}
Chain-of-thought reasoning enables models to use additional test-time compute through step-by-step problem decomposition~\citep{wei2022chain,mmcot}.
The two dominant approaches for training reasoning models are distillation, where a strong teacher generates reasoning traces for supervised fine-tuning~\citep{llava_cot,mulberry,sarchgrounded}, and reinforcement learning, which optimizes against outcome-based rewards without requiring a fixed teacher~\citep{deepseekr1}. 
Recent works apply RL to visual reasoning~\citep{perceptionr1,vlrethinker,openmmreasoner,feng2025onethinker}, but primarily in narrow domains, leaving the effect of RL-trained reasoning on broad visual understanding underexplored. We show that RL with careful reward and data design consistently outperforms narrowly trained baselines across diverse visual task categories.

\paragraph{RL recipes and training data design for VLMs.}
Several works provide recipes for RL-based visual reasoning training. OpenMMReasoner~\citep{openmmreasoner} combines teacher distillation and GSPO~\citep{gspo} over multimodal reasoning benchmarks, OneThinker~\citep{feng2025onethinker} uses a distillation warm start before RL, VL-Rethinker~\citep{vlrethinker} addresses training instability via selective sample replay and forced rethinking, and Perception-R1~\citep{perceptionr1} designs discriminative rewards for perceptual tasks. These efforts primarily target visual math or narrow perceptual domains and provide only limited ablations of dataset selection, sample filtering, and reward design. Our recipe centers on \verodataset, which spans six task categories with 600K data points from \numdatasets{} datasets, includes a routed reward system, and provides systematic ablations of design choices, all released publicly to support open VLM research.

\section{\model{}}
\label{sec:method}

\subsection{Task Categories}
\label{subsec:task-definitions}

We consider the problem of training a Vision-Language Model (VLM) $\pi_\theta$ via reinforcement learning to maximize expected reward across a diverse set of visual reasoning tasks. Given a visual input $v$ (an image or set of images) and a text query $q$, the model generates a structured response $y = (z, a) \sim \pi_\theta(\cdot \mid v, q)$, where $z$ denotes the reasoning or thinking content and $a$ denotes the final answer. We verify the final answer $a$ against the ground-truth answer $y^*$. The RL training objective is:
\begin{equation}
    \max_\theta \; \mathbb{E}_{(v, q, y^*) \sim \mathcal{D}} \; \mathbb{E}_{(z, a) \sim \pi_\theta(\cdot \mid v, q)} \left[ R(a, y^*) \right],
\end{equation}
where $\mathcal{D}$ is the training data distribution. A central challenge is constructing $\mathcal{D}$ to span a broad range of visual reasoning capabilities, so that the resulting policy generalizes across diverse tasks.

Figure~\ref{fig:training-datasets} provides an overview of our training data composition. We organize our training data into six task categories, each targeting a distinct visual reasoning capability. This taxonomy is motivated by two observations. First, we find empirically (Section~\ref{sec:cross_gen} and Section~\ref{sec:behavioral-analysis}) that training on any single category fails to transfer reliably to others and elicits distinct chain-of-thought behaviors, suggesting that these categories exercise different reasoning strategies and skills. Second, while existing VLM evaluation frameworks organize benchmarks along similar axes, e.g., Qwen2.5-VL~\citep{qwen25vl} separates document understanding, mathematical reasoning, and grounding, and Kimi K2.5~\citep{kimik25} distinguishes reasoning from perception, these categorizations are typically adopted by convention rather than validated empirically, and they are designed for evaluation rather than training. Our taxonomy refines and extends these axes to cover a broader set of visual reasoning tasks, and we validate its effectiveness for multi-task RL (Section~\ref{sec:cross_gen}).

Concretely, we define six categories:
\textbf{STEM} (13 datasets) covers mathematical diagram reasoning, scientific figure interpretation, and medical image understanding, with answers that are typically numeric or symbolic.
\textbf{Spatial \& Action} (8 datasets) targets embodied reasoning, UI navigation, and 3D spatial understanding, requiring reasoning about spatial transformations and action sequences.
\textbf{Knowledge \& Recognition} (12 datasets) spans visual question answering that combines object, scene, and entity recognition with external or commonsense knowledge.
\textbf{Chart \& OCR} (9 datasets) focuses on extracting and reasoning over structured information in documents, charts, tables, and infographics.
\textbf{Grounding, Counting \& Search} (11 datasets) requires spatially localizing objects via bounding boxes, counting entity instances, and searching among visual distractors.
\textbf{Captioning \& Instruction Following} (6 datasets) encompasses open-ended image description and following prompt instructions. 

\begin{figure*}[t]
    \centering
    \includegraphics[width=1.0\linewidth]{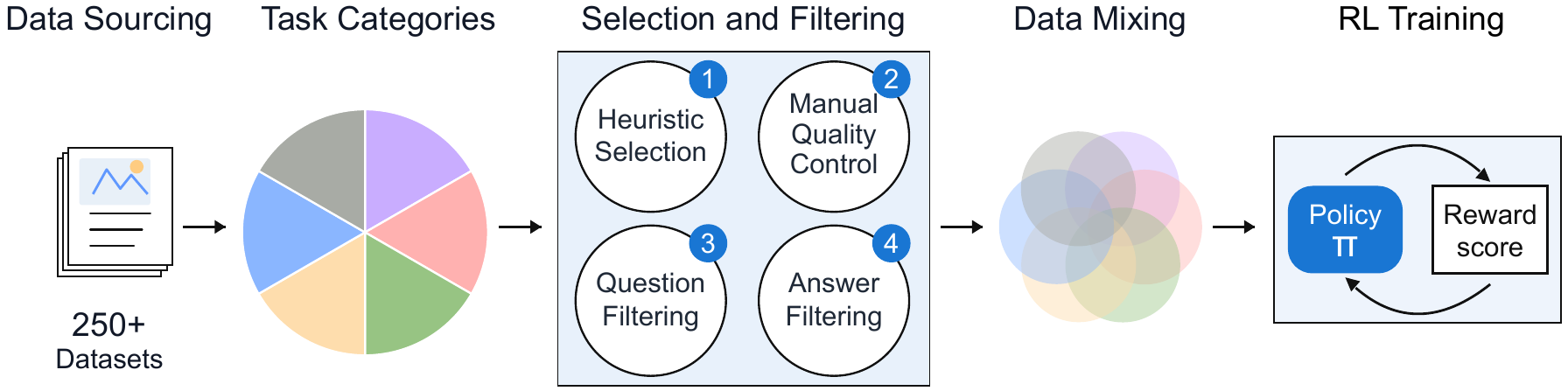}
            \caption{\textbf{\verodataset data curation pipeline.} Starting from over 250 candidate datasets, we assign each to one of six task categories and apply multi-stage selection and filtering: heuristic screening (size, resolution, answer format), manual quality control, LLM-based question filtering for ambiguity and verifiability, and answer filtering for stable reward computation. The retained data are combined into a uniformly weighted mixture across task categories and used for on-policy RL training with task-routed rewards.}
    \label{fig:curation-pipeline}
\end{figure*}

\begin{figure*}[t]
    \centering
    \includegraphics[width=1.0\linewidth]{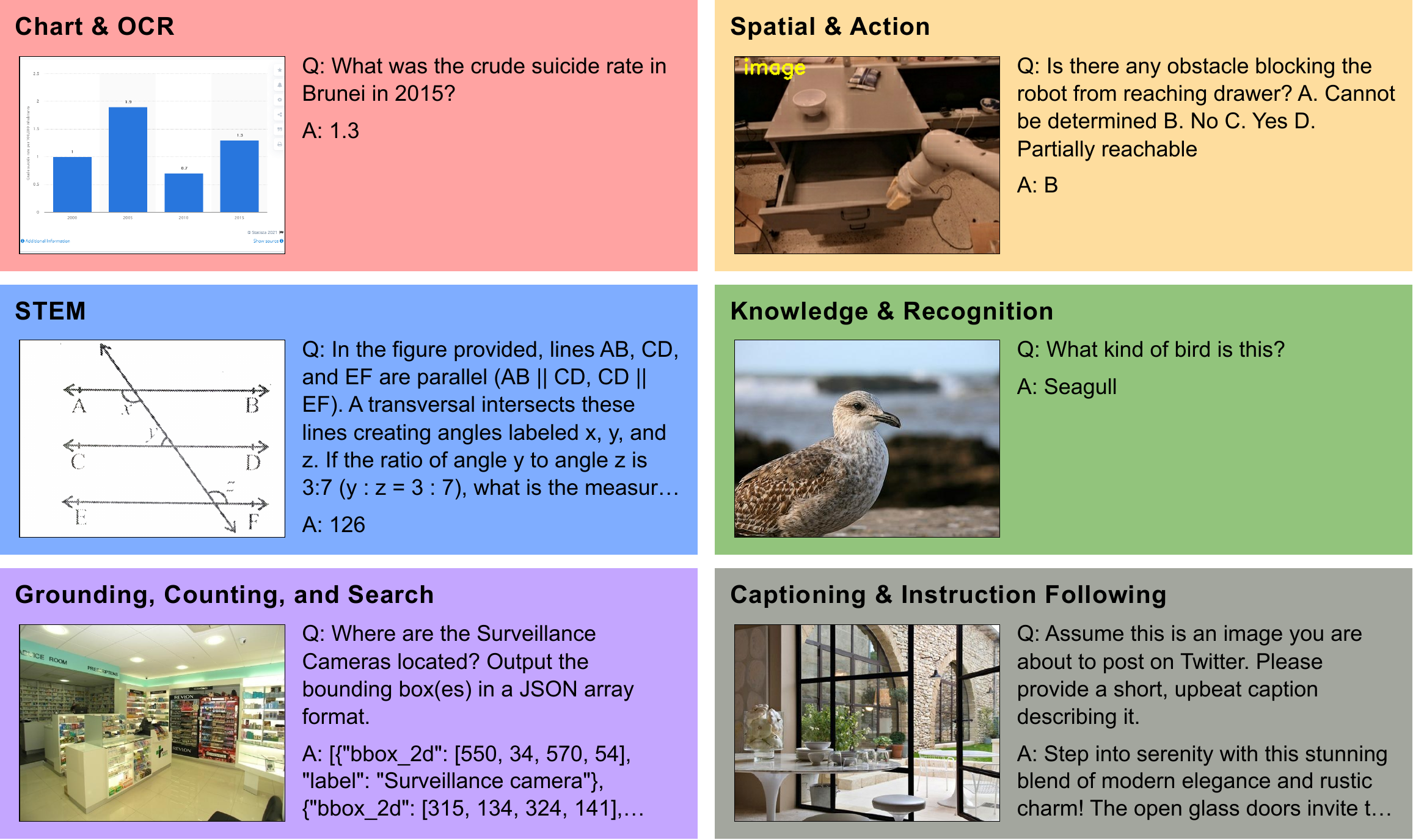}
            \caption{\textbf{Examples from each task category illustrate the breadth of \verodataset{}.} We show representative samples of our training data from the six categories, highlighting the diversity of visual inputs, question formats, and answer types covered by our training set.}
    \label{fig:examples}
\end{figure*}

\subsection{\verodataset{}: Sourcing, Filtering, and Mixtures of Broad Tasks}
\label{subsec:data}

\newcommand{\rotmidf}[1]{\raisebox{-0.9ex}{\rotatebox{90}{\scriptsize\shortstack[l]{\strut #1}}}}

We construct \verodataset, a multi-task RL training set of 600K samples from \numdatasets{} datasets spanning six task categories (Section~\ref{subsec:task-definitions}).
Figure~\ref{fig:curation-pipeline} summarizes the \verodataset data curation pipeline, and Figure~\ref{fig:examples} shows representative examples from all six task categories.

\paragraph{Step 1. Dataset sourcing and selection.} We start from over 250 candidate datasets drawn from instruction-tuning and RL collections (e.g., FineVision~\citep{finevision}) and recently released task-specific sources (e.g., Visual Jigsaw~\citep{wu2025visualjigsaw}), then apply dataset-level filtering. Each dataset is assigned to the task category that best reflects its primary skill, based on manual inspection and its utility in prior work.

\newcommand{\rot}[1]{\rotatebox{90}{\scriptsize\shortstack[c]{\strut #1}}}
\newcommand{\vnum}[1]{\raisebox{0.45ex}{#1}}
\definecolor{bestcellblue}{HTML}{EBF5FD}
\newcommand{\bestcell}[1]{\cellcolor{bestcellblue}\textbf{#1}}
\newcommand{\bestcellv}[1]{\cellcolor{bestcellblue}\vnum{\textbf{#1}}}

\paragraph{Heuristic selection.} We discard datasets with fewer than 1K examples, average image resolution below 200K pixels (retaining five low-resolution datasets for question quality), or binary questions to mitigate guessing.

\paragraph{Manual selection.} For each candidate, we inspect $\sim$50 examples against three criteria: \textbf{correctness} ($<$5\% annotation error rate in image--question--answer triples), \textbf{unambiguity} (each question admits a single verifiable answer), and \textbf{verifiability} (the answer format is compatible with our reward functions). Of $\sim$100 datasets passing heuristic screening, \numdatasets{} were retained. For a small number of datasets, we additionally rewrite questions to fix prompt clarity (e.g., GameQA, Magma) or drop high-error subsets.

Figure~\ref{fig:dataset-filtering-ablation} (Appendix) compares dataset selection strategies and shows our filtering has significant gains compared to taking a random subset from the candidate pool or sampling from the STEM or Chart subsets of the FineVision~\citep{finevision} training set. On STEM, our selected mixture achieves an average benchmark gain of +4.6, compared to +2.9 for FineVision and $-$0.2 for random sampling from all candidates. On Chart \& OCR, the gains are +3.4, +1.9, and +2.5, respectively.

\smallskip
\paragraph{Step 2. Data filtering.}
After dataset-level filtering, many individual examples remain ambiguous, unanswerable, or incompatible with our rewards. We apply additional steps to filter individual prompts.
\smallskip

\paragraph{Question filtering.} We use Qwen3-VL-235B-A22B-Instruct~\citep{qwen3vl} to remove ambiguous, image-irrelevant, or unverifiable questions. The model scores each datapoint on five criteria: (1)~\emph{relevance}, whether the image depicts what the question refers to; (2)~\emph{ambiguity}, whether the question is too vague or not a genuine question; (3)~\emph{language}, whether the question is in English; (4)~\emph{verifiability}, whether a single objectively correct answer can be derived from visible content; and (5)~\emph{numeric precision}, whether the required precision is visually unambiguous. Any triggered criterion removes the datapoint. We provide details in \appref{app:filtering-prompts}.

\paragraph{Answer filtering.} We normalize ground-truth answers using text-only Qwen3-235B-A22B-Instruct~\citep{qwen3vl} to ensure stable reward computation. 
Numeric answers are stripped of units and currency symbols, converted to decimal form, and evaluated as expressions. Samples with unsupported notation are filtered. Multiple-choice answers are normalized to a single canonical letter. 
Except for the captioning and instruction following task category, samples with multi-value answers, non-reducible symbolic expressions, or ambiguous descriptions requiring semantic matching are removed. A full list of answer filtering rules is in \appref{app:answer-filtering-details}.
We do not apply it to Spatial \& Action or Grounding tasks, since answers are already standardized for all datasets in these categories.

Effects of filtering are mixed across categories (Table~\ref{tab:filtering-ablations}): question filtering yields a clear gain on Spatial \& Action (+1.9 pts) but slightly hurts Grounding, Counting \& Search ($-0.6$ pts), while answer filtering substantially improves Knowledge \& Recognition (+2.1 pts) but is flat or marginally negative on other categories. Despite this variance, we apply both steps to all applicable task categories, as they generally help remove ambiguous samples from noisy datasets and the largest gains outweigh the small regressions.

\begin{table*}[t]
    \centering
    \begin{minipage}[b]{0.43\linewidth}
        \centering
        \adjustbox{max width=\linewidth}{        \small
        \setlength{\tabcolsep}{5pt}
        \begin{tabular}{l ccccc}
        \toprule
        & \rotmidf{Chart \&\\[-3pt]OCR} & \rotmidf{STEM} & \rotmidf{Spatial \&\\[-3pt]Action} & \rotmidf{Knowl. \&\\[-3pt]Recog.} & \rotmidf{Grnd.,\\[-3pt]Cnt. \&\\[-2pt]Search} \\
        \midrule
        Unfiltered & 60.0 & 45.4 & 56.3 & 62.5 & \bestcell{54.7} \\
        Q.~Filtering & \bestcell{60.1} & 43.6 & \bestcell{58.2} & 63.0 & 54.1 \\
        A.~Canonic. & 59.9 & \bestcell{45.5} & -- & \bestcell{64.6} & -- \\
        \bottomrule
        \end{tabular}}
    \end{minipage}
    \hfill
    \begin{minipage}[b]{0.55\linewidth}
        \centering
        \adjustbox{max width=\linewidth}{        \small
        \setlength{\tabcolsep}{2.5pt}
        \begin{tabular}{l cccccc}
        \toprule
        & \rotmidf{Chart \&\\[-3pt]OCR} & \rotmidf{STEM} & \rotmidf{Spatial \&\\[-3pt]Action} & \rotmidf{Knowl. \&\\[-3pt]Recog.} & \rotmidf{Grnd.,\\[-3pt]Cnt. \&\\[-2pt]Search} & \rotmidf{Bench.\\[-3pt]Avg.} \\
        \midrule
        equal ratios                          & \bestcell{+8.6} & +6.2 & \bestcell{+5.6} & +1.8 & +5.6 & \bestcell{+5.8} \\
        ratio $\propto$ $(1 - \text{acc.})^\alpha$ & +6.8 & \bestcell{+6.5} & +4.3 & \bestcell{+2.4} & +5.2 & +5.2 \\
        ratio $\propto$ $\text{area}^\alpha$       & +7.0 & +5.3 & +4.1 & +1.4 & \bestcell{+6.2} & +5.2 \\
        ratio $\propto$ $\text{length}^\alpha$     & +7.5 & +6.4 & +4.5 & +1.7 & +3.8 & +4.8 \\
        w/o Knowl.\ \& Recog.                & +6.4 & \bestcell{+6.5} & +4.8 & +1.9 & +4.7 & +4.9 \\
        \bottomrule
        \end{tabular}}
    \end{minipage}

    \vspace{18pt}

    \begin{minipage}[t]{0.42\linewidth}
        \captionof{table}{\textbf{Filtering generally helps remove ambiguous samples from noisy datasets.} Effect of question filtering and answer filtering on Qwen2.5-VL-7B-Instruct. This table shows category average scores on \modelsuite{}.}
        \label{tab:filtering-ablations}
    \end{minipage}
    \hfill
    \begin{minipage}[t]{0.55\linewidth}
        \captionof{table}{\textbf{Equal task ratios perform best overall.} Weighting schemes: equal ratios (uniform), difficulty-weighted by inverse accuracy, image-area-weighted by mean input resolution, reasoning-length-weighted by mean chain-of-thought length, and ablation dropping Knowledge \& Recognition. Values are absolute score changes ($\Delta$) on \modelsuite{} when using Qwen3-VL-8B-Instruct as the initial model for RL.}
        \label{tab:data-mixtures}
    \end{minipage}
\end{table*}

\smallskip
\paragraph{Step 3. Data mixtures.}
In our multi-task RL setting, the task category sampling distribution governs how training signal is allocated across skills. We investigate four task category weighting schemes (uniform, difficulty-weighted, image-size-weighted, reasoning-length-weighted), where the number of samples per batch is determined by a ratio proportional to the metric (e.g., difficulty). Uniform sampling achieves the highest benchmark average gain (+5.8 pts over the base model), outperforming alternative schemes. Alternatives yield gains on individual categories but at the cost of others (Table~\ref{tab:data-mixtures}). We use uniform task category weighting, as it achieves the best overall performance.
\smallskip

\paragraph{\modelsuite{} evaluation suite.} We introduce \modelsuite{}, a challenging evaluation suite for broad visual reasoning. We curate a suite of 30 benchmarks spanning the six visual reasoning categories defined in Section~\ref{sec:method}, with three to eight benchmarks per category. We select benchmarks according to three criteria: (i)~\emph{difficulty}: we favor benchmarks on which current frontier models have room for improvement, while retaining established benchmarks (e.g., ChartQA, ScreenSpot) for comparability with prior work; (ii)~\emph{annotation quality}: we include only benchmarks with well-defined evaluation protocols and reliable ground-truth labels; and (iii)~\emph{intra-category diversity}: within each category we include benchmarks that test complementary sub-skills (e.g., within Chart \& OCR: chart reasoning, infographic understanding, and scientific figure interpretation). The full benchmark list appears in Appendix Table~\ref{tab:eval-benchmark-table}.

\subsection{Training \model{} with Reinforcement Learning}
\label{subsec:rl}

\paragraph{Algorithmic details.} At its core, RL maximizes the expected reward of the model's response $y$ given a visual input $v$ and query $q$. Our RL algorithm builds on Group Relative Policy Optimization (GRPO)~\citep{grpo} and integrates algorithmic advances from GSPO~\citep{gspo}, among others~\citep{dapo}.

GSPO~\citep{gspo} replaces the independent per-token importance ratios of GRPO with a \emph{sequence-level} ratio. For each response $y_i$ in a group of $G$ rollouts, the sequence-average log-probability difference $\bar{\Delta}_i = \frac{1}{|y_i|}\sum_{t} (\log \pi_\theta(y_{i,t}) - \log \pi_{\theta_\text{old}}(y_{i,t}))$ is used to form a token-level ratio $s_{i,t}(\theta) = \exp(\operatorname{sg}(\bar{\Delta}_i) + \log \pi_\theta(y_{i,t}) - \operatorname{sg}(\log \pi_\theta(y_{i,t})))$, where $\operatorname{sg}$ denotes stop-gradient. The GSPO objective is then:

\begin{equation}
\mathcal{J}(\theta) =
\frac{1}{G}\sum_{i=1}^{G} \frac{1}{|y_i|} \sum_{t=1}^{|y_i|}
\min\!\Bigl(
  s_{i,t}(\theta)\,A_i,\;
  \operatorname{clip}\!\bigl(s_{i,t}(\theta),\,1{-}\varepsilon_\text{low},\,1{+}\varepsilon_\text{high}\bigr)\,A_i
\Bigr),
\end{equation}

where $A_i = (r_i - \mu_g)/(\sigma_g + \epsilon)$ is the normalized group advantage; note that in our setting $A_{i,t} = A_i$ for all $t$, i.e., the advantage is constant across tokens within a response. We adopt asymmetric clip-higher~\citep{dapo} ($\varepsilon_\text{high} > \varepsilon_\text{low}$), remove the KL penalty~\citep{dapo,liu2025understanding} to allow less-restricted updates, and apply a soft overlong penalty~\citep{dapo} that linearly ramps before the context limit.

\paragraph{Reward formulation.} 
The total reward for a response $y$ is ($\alpha = 0.2$):
\vspace{-0.25em}
\begin{equation}
R(y, y^*) = (1 - \alpha)\,R_\text{acc}(y, y^*) + \alpha\,R_\text{fmt}(y)
+ R_\text{overlong}(y),
\end{equation}
\vspace{-0.25em}
\paragraph{Overlong penalty.}
To discourage excessively long responses, we use the soft penalty from \citet{dapo} as a linear ramp in the buffer zone $[L_\text{max} - B,\; L_\text{max}]$ ($B = 2048$, $L_\text{max} = \texttt{max\_tokens}$, and $\lambda = 1.0$):
\begin{equation}
R_\text{overlong}(y) = \min\!\bigl(-\tfrac{|y| - (L_\text{max} - B)}{B}\,\lambda,\; 0\bigr),
\end{equation}
\vspace{-2em}

\paragraph{Format reward.}
$R_\text{fmt}$ requires the response to follow the format \texttt{<think>}$\ldots$\texttt{</think>}\allowbreak\texttt{<answer>}$\ldots$\texttt{</answer>} with non-empty think content; responses that violate this structure receive $R_\text{fmt} = 0$. Given valid structure, $R_\text{fmt} = 1$ by default. For discrete symbolic answer types (string match, multiple choice, numeric, list match, counting, ordering, search, web action), a single valid \verb|\boxed{...}| in the answer block is additionally required for $R_\text{fmt} = 1$; its absence or the presence of multiple \verb|\boxed{...}| expressions reduces $R_\text{fmt}$ to $0.5$.

\begin{figure*}[t]
    \centering
    \includegraphics[width=1.0\linewidth]{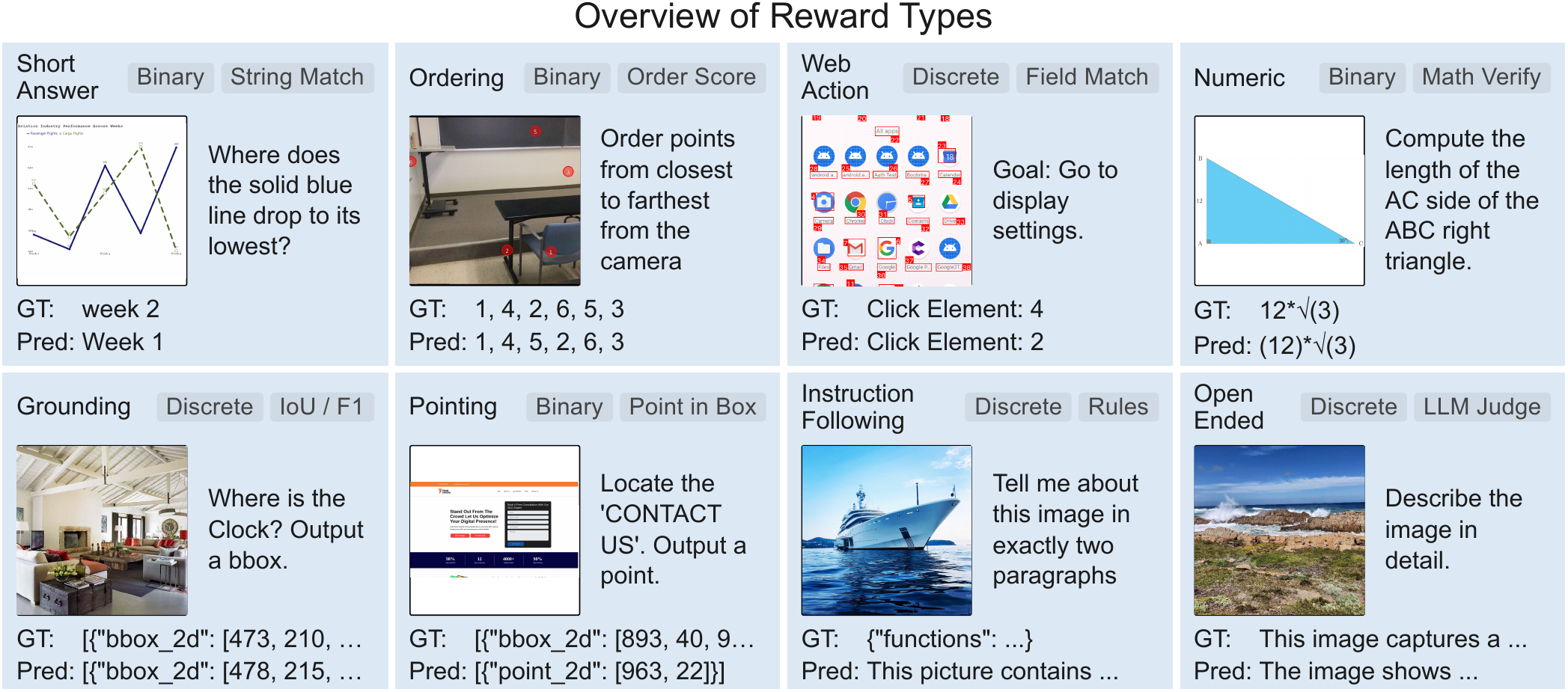}
    \caption{\textbf{Accuracy verifiers for our multi-task reward.} Each card illustrates a verifier used in \model{}, with an example visual question, ground-truth answer, and model prediction. Verifiers include binary rewards (string match, multiple choice (not displayed), list string match (not displayed), numeric, ordering, point-in-box) and graded rewards (IoU/F1 for grounding, field match for web actions, rule-based checks for instruction following, and LLM-as-judge for open-ended responses). This task-routed design enables accurate reward computation across diverse answer formats.}
    \label{fig:rewards}
    \vspace{-0.5em}
\end{figure*}

\paragraph{Multi-task reward.}
For $R_\text{acc}(y, y^*)$, we detail below the ten reward functions corresponding to the task types in our dataset (Figure~\ref{fig:rewards}). We show in Section~\ref{sec:ablations} that our reward design outperforms simple alternatives.
\vspace{-1em}
\begin{itemize}[leftmargin=*,itemsep=1pt]
    \item \textbf{String match} ($\in\{0,1\}$): normalized exact-string equality.
    \item \textbf{Multiple choice} ($\in\{0,1\}$): extracts a single letter (A--Z) and compares it to the predicted letter.
    \item \textbf{Numeric} ($\in\{0,1\}$): symbolic parsing via \textsc{math-verify}~\citep{mathverify}, with optional tolerance.
    \item \textbf{List string match} ($\in\{0,1\}$): any-match across a set of strings, handling synonym-equivalent answers.
    \item \textbf{Ordering} ($\in[0,1]$): full reward for exact list order and partial reward (discounted by a factor of 0.2) for correct set with wrong order. Adapted from Visual Jigsaw~\citep{wu2025visualjigsaw}.
    \item \textbf{Web action} ($\in[0,1]$): weighted match over structured JSON fields (\texttt{ACTION}, \texttt{MARK}, \texttt{VALUE}), with score equal to the fraction of non-null gold fields correctly predicted. Adapted from ViGoRL~\citep{sarchgrounded}.
    \item \textbf{Grounding} ($\in[0,1]$): optimal Hungarian matching of predicted and ground-truth bounding boxes, scoring IoU/F1 with threshold 0.5. Bounding box coordinates are normalized to the $[0, 1000]$ range (Qwen-style). Adapted from Perception-R1~\citep{perceptionr1}.
    \item \textbf{Clicking} ($\in[0,1]$): checks whether the predicted click point falls within the ground-truth bounding box region, with coordinates in the same normalized space. Adapted from ViGoRL~\citep{sarchgrounded}.
    \item \textbf{Instruction following} ($\in[0,1]$): proportion of programmatically defined output constraints satisfied (e.g., length limits, format requirements, keyword inclusions). We use the constraint checks from MMIFeval~\citep{ding2025mmifeval} and RLVR-IFeval~\citep{pyatkingeneralizing}.
    \item \textbf{LLM-as-judge} ($\in[0,1]$): We adapt the judge setup from OLMo3~\citep{olmo2025olmo}. We use Qwen3-32B with thinking disabled to score the response against an optional reference answer. The judge prompt instructs the model to score 1--10 and explicitly penalizes self-evaluative language and meta-commentary to reduce reward hacking. See \appref{app:judge-prompt} for the full prompt.
\end{itemize}

\definecolor{tblred}{HTML}{FF8988}
\definecolor{tblorange}{HTML}{FECC81}
\definecolor{tblblue}{HTML}{6098FF}
\definecolor{tblgreen}{HTML}{77B25D}
\definecolor{tblpurple}{HTML}{B28CFF}
\definecolor{tblgray}{HTML}{8B8F87}

\begin{table*}[t]
  \centering
  \setlength{\tabcolsep}{1.4pt}
  \renewcommand{\arraystretch}{1.12}
  \large
  \begin{adjustbox}{max width=\textwidth}
\begin{tabular}{l *{6}{l} @{\hspace{3pt}} *{5}{l} @{\hspace{3pt}} *{4}{l} @{\hspace{3pt}} *{1}{l}}
      \toprule
      & \multicolumn{6}{c}{\textbf{\model{} (Ours)}} & \multicolumn{5}{c}{\textbf{Open Weights Models}} & \multicolumn{4}{c}{\textbf{Fully Open RL Recipes}} & \multicolumn{1}{c}{\textbf{Propr.}} \\
\bottomrule
    \end{tabular}
  \end{adjustbox}
  \caption{\textbf{\model{} achieves state-of-the-art performance across six task categories on \modelsuite{}.} \model{} columns show Avg@5 over initial models trained with RL on our dataset; superscript $\pm$ values report the standard error of the mean across runs, and {\scriptsize\textcolor{tblgreen}{+x}} / {\scriptsize\textcolor{tblred}{-x}} deltas indicate improvement/decline over the respective initial model. In model names, Qw denotes Qwen and Mi denotes MiMo. \textsuperscript{\textdagger} indicates results evaluated by us. All other results are taken from official technical reports.}
  \label{tab:eval-results-transposed}
\end{table*}

\vspace{-1em}
\section{Experiments}
\label{sec:experiments}

\paragraph{Evaluation settings.} We use the following decoding setups. Qwen2.5-VL and
MiMo-VL trained models follow the Qwen2.5-VL~\citep{qwen25vl} recommended decoding setup. Qwen3-VL trained models follow the decoding setup
reported in the Qwen3-VL report~\citep{qwen3vl}. Tables~\ref{tab:eval-inference-settings-qwen25} and
\ref{tab:eval-inference-settings-qwen3} summarize the model-family-specific
sampling parameters and the shared runtime settings. We use Qwen3-32B with thinking disabled as the evaluation LLM judge when an LLM judge is required. For benchmarks requiring a VLM judge, we use Qwen3-VL-32B-Instruct. For judges, we use sampling parameters set to Temperature=0.7, TopP=0.8, TopK=20, and MinP=0.

We evaluate all models using the lmms-eval~\citep{zhang2025lmms} framework, following the official evaluation protocols specified by each benchmark's authors. Full benchmark-specific choices and metric details are provided in \appref{app:eval-details}. 
Table~\ref{tab:eval-results-transposed} reports Avg@5 for \model{} variants, with superscript $\pm$ values denoting the standard error of the mean across runs. For the Qwen3.5-9B-Base model which exhibited weak instruction-following behavior, we extracted predicted answers using GPT-5.4-mini for ChartQA-Pro, EvoChart, and CountQA, and via the LaTeX boxed format for InfoVQA, ChartQA, GameQA-lite, and MathVision.

\FloatBarrier

\paragraph{Baselines.} We compare against: (1) base VLMs without native <think> tokens (Qwen2.5-VL-7B-Instruct~\citep{qwen25vl}, Qwen3-VL-8B-Instruct~\citep{qwen3vl}, Qwen35-9B, Molmo2-O-7B~\citep{molmo2}), (2) models trained to do native CoT with <think> tokens (Qwen3-VL-8B-Thinking~\citep{qwen3vl}, MiMo-VL-7B-RL~\citep{mimovl}), (3) existing fully open RL-trained models and recipes (VL-Rethinker-7B~\citep{vlrethinker}, LLaVA-OV-1.5-RL~\citep{an2025llava}, OpenMMReasoner-7B~\citep{openmmreasoner}, OneThinker-8B~\citep{feng2025onethinker}), and (4) a proprietary reasoning model gpt-5-nano-2025-08-07~\citep{gpt5} with medium reasoning effort. For baseline results, we prioritize scores from official technical reports and benchmark leaderboards. When published results are unavailable, we evaluate the models ourselves (indicated by \textsuperscript{\textdagger}) and follow the published benchmark guidelines.

\begin{figure}[t!]
    \centering
    \includegraphics[width=\linewidth]{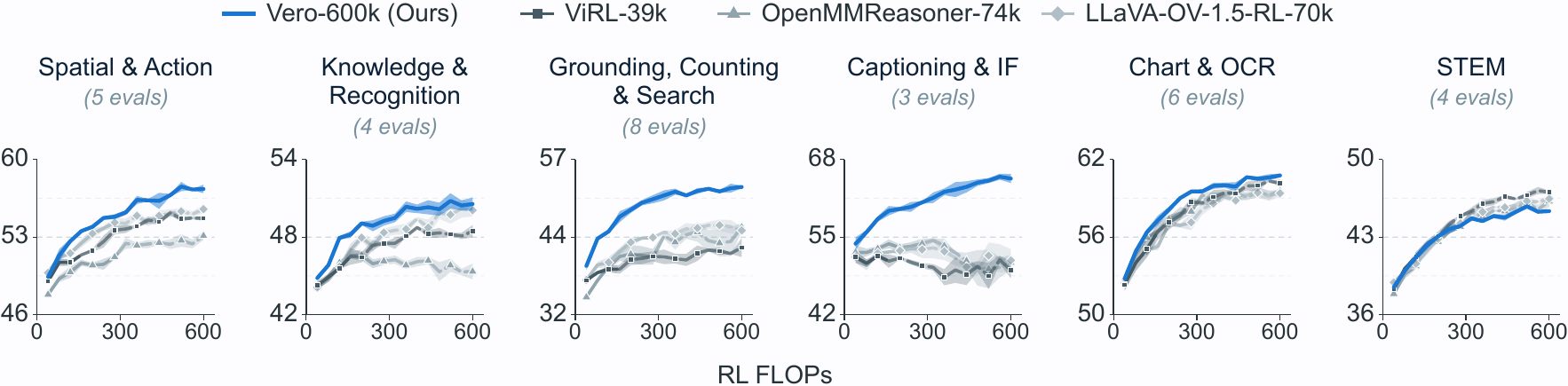}
    \caption{\textbf{Per-category RL training curves.} Evaluation score vs.\ RL FLOPs for RL runs using \verodataset and three prior open RL datasets, all starting from Qwen2.5-VL-7B-Instruct under identical RL settings. Curves report mean and SEM across seeds where shown. Dashed segments indicate training beyond one epoch. \verodataset leads on five of six categories throughout training; on STEM it remains within 2 points of the best prior dataset.}
    \label{fig:rl-curves-6-categories}
\end{figure}

\subsection{Evaluation Results on \modelsuite{}}
\label{sec:evaluation-results}

We report results in Table~\ref{tab:eval-results-transposed} and highlight the following observations.

\paragraph{Strong performance with a fully open RL recipe.} \modelQthreefive{} achieves the highest overall average in Table~\ref{tab:eval-results-transposed}, reaching 74.4 on \modelsuite{}. It improves over Qwen35-9B by +2.9 overall, wins 25 of 30 benchmarks, and improves all six category averages: Chart \& OCR (+2.3), STEM (+3.3), Spatial \& Action (+4.0), Knowledge \& Recognition (+2.5), Grounding, Counting \& Search (+2.3), and Captioning \& IF (+4.1). Among 8B models, \modelQthreeI{} and \modelQthreeT{} reach 66.1 and 65.8 overall, respectively, outperforming Qwen3-VL-8B-Thinking by +3.8 and +3.5 overall. Against distillation warm start baselines using the same initial model, \modelQtwofive{} is higher than OpenMMReasoner-7B by +5.6 overall, wins 5 of 6 category averages, and has its largest category margins on Captioning \& IF (+24.3) and Grounding, Counting \& Search (+11.1). \modelQthreeI{} is higher than OneThinker-8B~\citep{feng2025onethinker} by +10.4 overall and wins all six category averages. \model{} uses no additional teacher distillation in either comparison.

\paragraph{Consistent gains across initial models.} \model{} training yields improvements across six different initial models. Over the five post-trained initial models, overall gains range from +2.9 to +5.4. \modelQthreeI{} improves over Qwen3-VL-8B-Instruct by +5.4 overall and outperforms Qwen3-VL-8B-Thinking on 25 of 30 benchmarks despite using no distillation. \modelQthreeT{} improves over Qwen3-VL-8B-Thinking by +3.5 overall and on 22 of 30 benchmarks, and \modelQtwofive{} improves over Qwen2.5-VL-7B-Instruct by +4.9 overall and on 27 of 30 benchmarks. The largest improvement comes from applying RL directly to the pretrained-only Qwen3.5-9B-Base: +12.9 overall, reaching 73.0, the second-highest score in Table~\ref{tab:eval-results-transposed}, with no SFT or distillation warm start. Gains concentrate on Grounding, Counting \& Search (+23.0) and Captioning \& IF (+9.8, the table's best at 87.8), capabilities largely absent from the base checkpoint despite prompting attempts\footnote{Qwen35-9B-Base produced parseable coordinate outputs under prompting, but the predicted coordinates were poorly grounded.}, while reasoning categories also improve substantially (STEM +14.1).\\

\modelMi{}, trained on MiMo-VL-7B-SFT with our fully open recipe, improves by +3.6 overall. It is +0.8 overall above MiMo-VL-7B-RL, which trains on the same initial model but uses a proprietary RL recipe with non-public data. \modelMi{} is higher on STEM (+0.6), Knowledge \& Recognition (+5.1), and Captioning \& IF (+3.6), tied on Spatial \& Action, and lower on Chart \& OCR and Grounding, Counting \& Search.

\paragraph{Improvements not limited to a single domain.} Unlike prior open RL-trained VLMs that focus primarily on STEM, \model{} yields improvements across all six task categories. For example, \modelQthreeI{} improves over its initial model on Chart \& OCR (+8.3), STEM (+6.7), Spatial \& Action (+4.6), Knowledge \& Recognition (+1.5), Grounding, Counting \& Search (+4.9), and Captioning \& IF (+5.3), demonstrating that multi-task RL produces broadly capable models rather than specialists.

\paragraph{Consistent advantage over prior open RL datasets across training.}
Figure~\ref{fig:rl-curves-6-categories} compares RL runs using \verodataset against runs using three prior open RL datasets, ViRL-39k~\citep{vlrethinker}, OpenMMReasoner-74k~\citep{openmmreasoner}, and LLaVA-OV-1.5-RL-70k~\citep{an2025llava}. All runs start from Qwen2.5-VL-7B-Instruct and use identical RL settings over 600 steps, averaged over three runs with different random seeds. Even in the early training regime (first ${\sim}$150 steps), where all datasets are still within their first epoch, \verodataset already leads or remains within seed variance of the best prior dataset on every category. By the end of training, \verodataset reaches the highest score on five of six categories, with the largest margins on Captioning \& IF (+13.7 over the next best) and Grounding, Counting \& Search (+6.7). On STEM, where prior datasets largely concentrate their dataset selection, \verodataset remains competitive (45.3 vs.\ 47.0 for ViRL-39k and 46.2 for OpenMMReasoner-74k), trailing by fewer than 2 points despite learning across all six categories.

\begin{table}[t]
\centering
\vspace{-4pt}
\newcommand{\rothead}[1]{\rotatebox{90}{\scriptsize\shortstack[l]{\strut #1}}}

\setlength{\tabcolsep}{5.0pt}
\noindent\makebox[\textwidth]{\centering
\begin{minipage}[t]{0.58\textwidth}
\centering
\small
\textbf{(a) SFT vs.\ RL}
\vspace{2pt}

\begin{tabular}{@{}l cccccc c@{}}
\toprule
& \rothead{Chart \&\\[-3pt]OCR} & \rothead{STEM} & \rothead{Spatial \&\\[-3pt]Action} & \rothead{Knowl. \&\\[-3pt]Recog.} & \rothead{Grnd.,\\[-3pt]Cnt. \&\\[-2pt]Search} & \rothead{Cap. \&\\[-3pt]IF} & \rothead{Overall\\[-3pt]Avg.} \\
\midrule
Base              & 57.6 & 41.0 & 55.1 & 49.4 & 50.1 & 64.8 & 52.4 \\
FineVis SFT       & 54.8 & 37.4 & 52.1 & 45.3 & 40.1 & 52.2 & 46.2 \\
\model{} SFT      & 52.5 & 40.1 & 58.1 & 50.8 & 52.7 & 64.1 & 52.8 \\
\model{} RL       & \bestcell{61.9} & \bestcell{46.7} & \bestcell{59.4} & \bestcell{53.5} & \bestcell{55.0} & \bestcell{70.6} & \bestcell{57.2} \\
\bottomrule
\end{tabular}
\end{minipage}}

\vspace{6pt}

\begingroup
\setlength{\tabcolsep}{5.0pt}
\newcommand{\rotmid}[1]{\raisebox{-0.9ex}{\rothead{#1}}}
\noindent\makebox[\textwidth]{\begin{minipage}[t]{0.46\textwidth}
\centering
\small
\textbf{(b) Reward design}
\vspace{2pt}

\begin{tabular}{@{}l cccccc c@{}}
\toprule
& \rotmid{Chart \&\\[-3pt]OCR} & \rotmid{STEM} & \rotmid{Spatial \&\\[-3pt]Action} & \rotmid{Knowl. \&\\[-3pt]Recog.} & \rotmid{Grnd.,\\[-3pt]Cnt. \&\\[-2pt]Search} & \rotmid{Cap. \&\\[-3pt]IF} & \rotmid{Overall\\[-3pt]Avg.} \\
\midrule
Base       & 57.6 & 41.0 & 55.1 & 49.4 & 50.1 & 64.8 & 52.4 \\
Math Ver.  & 61.4 & 46.1 & 58.5 & 50.1 & 51.0 & 34.3 & 51.8 \\
Ours       & \bestcell{61.9} & \bestcell{46.7} & \bestcell{59.4} & \bestcell{53.5} & \bestcell{55.0} & \bestcell{70.6} & \bestcell{57.2} \\
\bottomrule
\end{tabular}
\end{minipage}\hspace{0.01\textwidth}
\begin{minipage}[t]{0.53\textwidth}
\centering
\small
\textbf{(c) RL algorithm (1/4 epoch, 5 task categories)}
\vspace{2pt}

\begin{tabular}{@{}l ccccc c @{\hspace{10pt}} @{}c@{}}
\toprule
& \rotmid{Chart \&\\[-3pt]OCR} & \rotmid{STEM} & \rotmid{Spatial \&\\[-3pt]Action} & \rotmid{Knowl. \&\\[-3pt]Recog.} & \rotmid{Grnd.,\\[-3pt]Cnt. \&\\[-2pt]Search} & \rotmid{Avg.} & \rotmid{Avg.\\[-3pt]Entropy} \\
\midrule
DAPO  & 58.9 & 45.3 & 57.1 & 49.7 & 52.2 & 54.3 & 0.22{\tiny$\pm$0.15} \\
GRPO  & \bestcell{59.2} & 44.4 & 58.1 & 48.2 & \bestcell{53.0} & 54.3 & 0.50{\tiny$\pm$0.11} \\
GSPO  & 59.0 & \bestcell{45.4} & \bestcell{58.4} & \bestcell{50.4} & \bestcell{53.0} & \bestcell{54.7} & \bestcell{0.58}{\tiny$\pm$0.11} \\
\bottomrule
\end{tabular}
\end{minipage}}
\endgroup
\caption{Ablation studies. All results on Qwen2.5-VL-7B-Instruct. Tables~(a)--(c) report absolute scores. All runs are trained 1 epoch on the 600k mixture unless otherwise specified.}\vspace{-0.7em}
\label{tab:ablations}
\vspace{-6pt}
\end{table}

\subsection{Ablations}\label{sec:ablations}

\paragraph{Multi-task RL requires more expressive reward design.} We compare our multi-route reward design against math\_verify~\citep{mathverify}, a widely used reward that performs extraction, parsing, and grading. Results in Table~\ref{tab:ablations}(b) show that our reward design, which routes answers through type-specific comparisons (exact match, numeric tolerance, set matching, and LLM-judge evaluation), achieves stronger performance than math\_verify across task categories. math\_verify lacks the flexibility to handle the diverse answer formats.

\paragraph{Our data benefits most from RL, yet even with SFT alone it outperforms strong SFT baselines.} We compare SFT and RL training on our dataset in Table~\ref{tab:ablations}(a). The SFT model is trained to directly output the final answer without chain-of-thought or \texttt{<think>} tokens. 
SFT on our dataset produces gains on most tasks and outperforms SFT on a recent post-training dataset FineVision~\citep{finevision}. However, RL (GSPO with our multi-route reward) yields more consistent improvements across all task categories.

\paragraph{GSPO outperforms GRPO and DAPO and leads to more stable entropy.} We compare three RL algorithms (DAPO, GRPO, and GSPO) using the same base model (Qwen2.5-VL-7B-Instruct), reward design, and training dataset. Consistent with recent findings~\citep{openmmreasoner}, results in Table~\ref{tab:ablations}(c) show that GSPO achieves the highest average score (54.7) across all task categories, outperforming both GRPO (54.3) and DAPO (54.3). GSPO also maintains substantially more stable entropy throughout training ($0.58 \pm 0.11$) compared to GRPO ($0.50 \pm 0.11$) and especially DAPO ($0.22 \pm 0.15$), suggesting that GSPO's sequence-level clipping better preserves exploration capacity and avoids premature policy collapse observed with alternative algorithms.

\begin{figure}[t!]
    \includegraphics[width=0.89\linewidth]{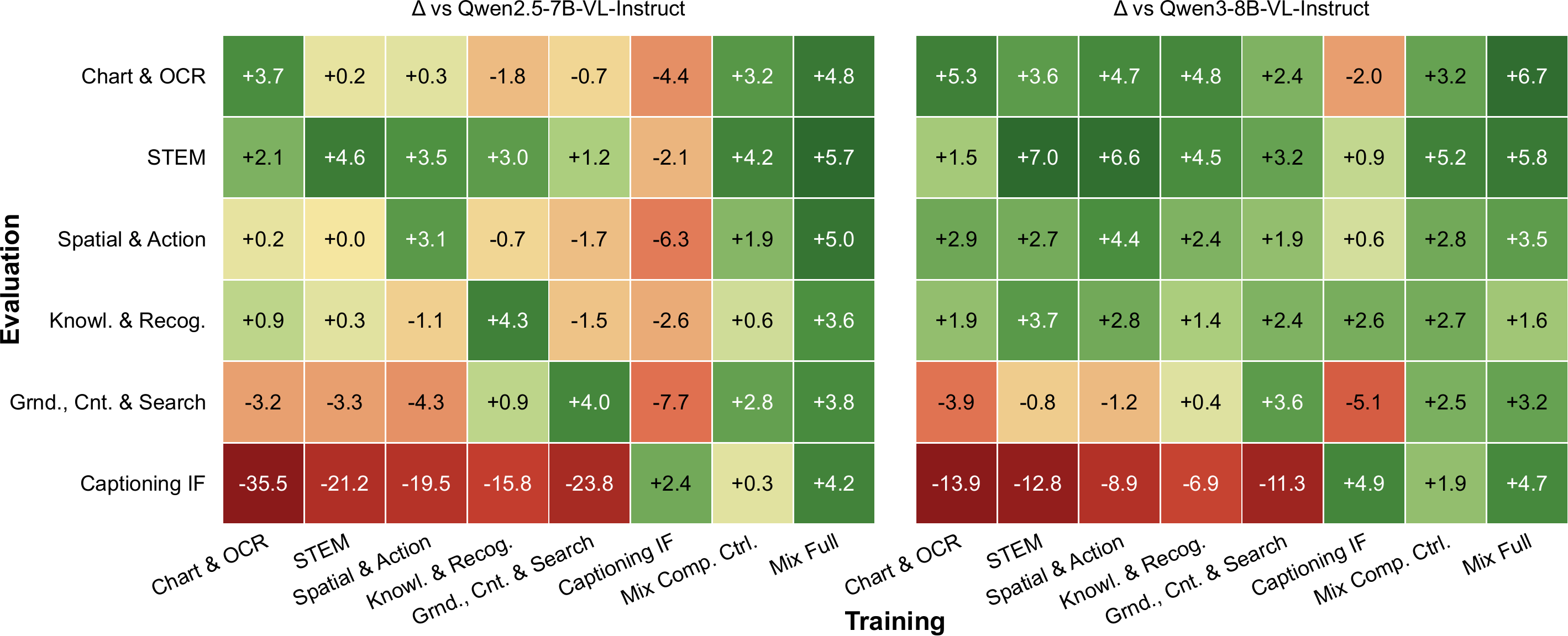}
    \caption{\textbf{Diverse task mixing eliminates negative cross-task transfer.} Each row shows a model trained on a single task category (or mixture of all categories). Values are absolute score changes relative to the base model. Single-task training yields selective transfer, while mixing achieves consistent gains.}
    \label{fig:cross-domain-generalization}
\end{figure}

\section{Data Diversity \& Cross-Task Transfer}\label{sec:cross_gen}

We study cross-task generalization by training models on each individual task category (100k samples, 1 epoch) and evaluating across all six categories. We compare against a model trained on a mixture of all categories with the same total number of training samples (100k, compute-controlled) and the full dataset (600k). We report results on two base models in Figure~\ref{fig:cross-domain-generalization}.

\paragraph{Single-task training frequently produces neutral or negative transfer on non-target tasks.} On Qwen2.5-VL, nearly all single-task-category models degrade Grounding, Counting \& Search performance (e.g., $-3.2$ from Chart \& OCR, $-3.3$ from STEM, $-4.3$ from Spatial \& Action), and training on Captioning \& Instruction Following alone reduces performance across all other categories ($-2.1$ to $-7.7$). Conversely, training on any non-captioning task category severely degrades Captioning \& Instruction Following ($-15.8$ to $-35.5$ on Qwen2.5-VL). These patterns hold on Qwen3-VL, where single-task-category models similarly hurt Grounding ($-0.8$ to $-5.1$ for non-grounding domains) and Captioning \& Instruction Following ($-6.9$ to $-13.9$). However, in certain task categories, we do observe selective positive transfer: STEM training improves Chart \& OCR (+3.6 on Qwen3-VL), and Spatial \& Action training yields strong gains on STEM (+3.5 on Qwen2.5-VL, +6.6 on Qwen3-VL).

\paragraph{Diverse task mixing eliminates negative cross-task transfer.} Even with the same compute budget, the mixed model achieves positive gains across all categories on both base models ($+0.3$ to $+4.2$ on Qwen2.5-VL; $+1.9$ to $+5.2$ on Qwen3-VL), avoiding the catastrophic losses seen with single-domain training. Training on the full 600k mixture further amplifies these gains. These patterns are consistent across both base models, suggesting that multi-task RL training is important for producing broadly capable models. We provide additional analysis and discussion in Sections~\ref{sec:behavioral-analysis} and~\ref{sec:discussion}, suggesting that reasoning behaviors may not transfer readily across the six task categories because their associated chain-of-thought patterns are largely distinct.
\clearpage

\begin{figure}[t!]
    \centering
    \begin{minipage}[t]{0.48\linewidth}
        \centering
        \begin{minipage}[c][0.18\textheight][c]{\linewidth}
            \centering
            \includegraphics[width=\linewidth]{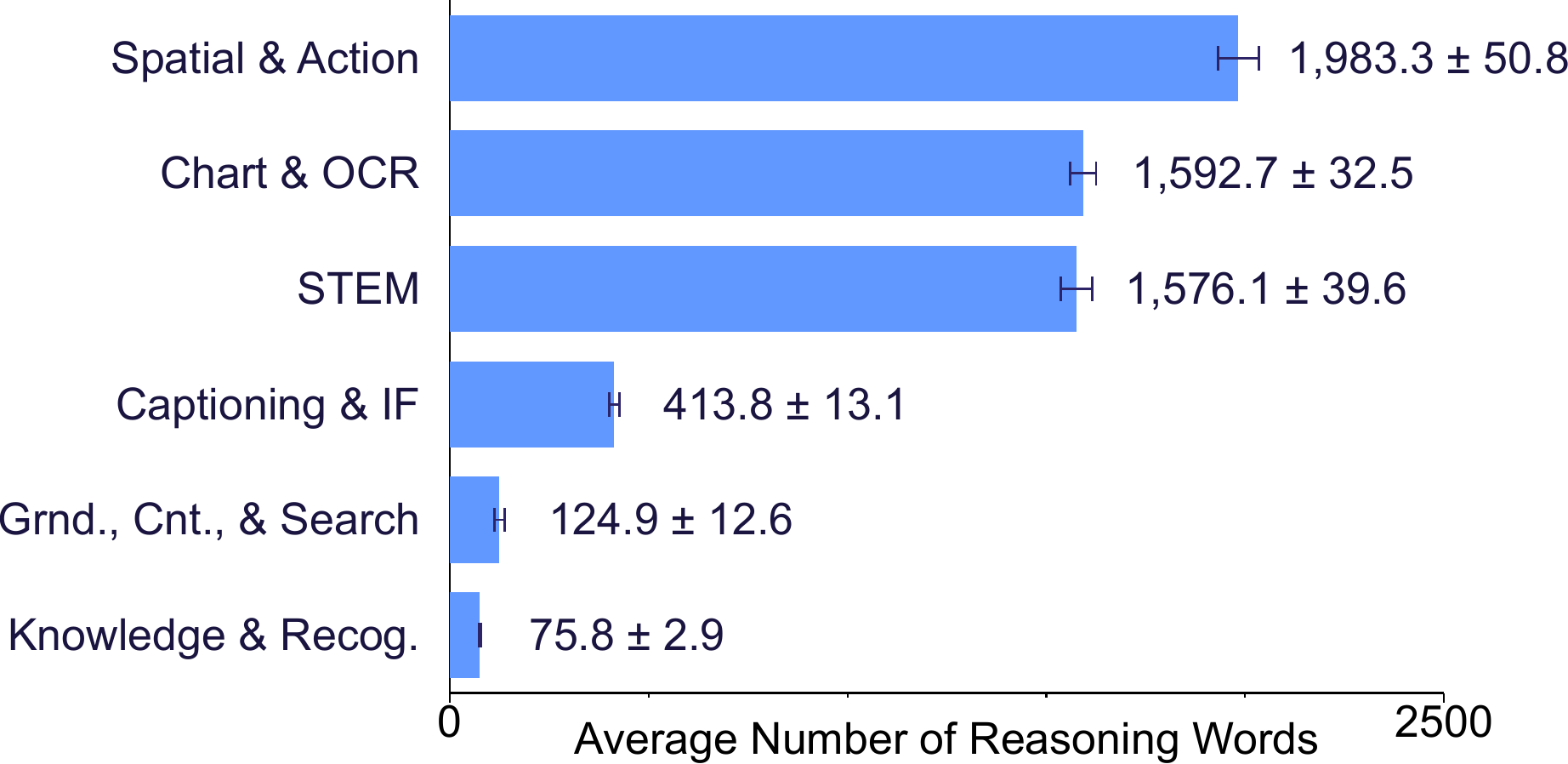}
        \end{minipage}
        \vspace{4pt}
        \caption{\textbf{RL on different task categories leads to varying reasoning lengths.} Average reasoning length (in words) on the validation set, measured after training Qwen3-VL-8B-Instruct for 1000 steps on each task category data (100k) and evaluating on the same category. Error bars denote the standard error of the mean.}
        \label{fig:reasoning-length-by-category}
    \end{minipage}
    \hfill
    \begin{minipage}[t]{0.48\linewidth}
        \centering
        \begin{minipage}[c][0.18\textheight][c]{\linewidth}
            \centering
            \includegraphics[width=\linewidth]{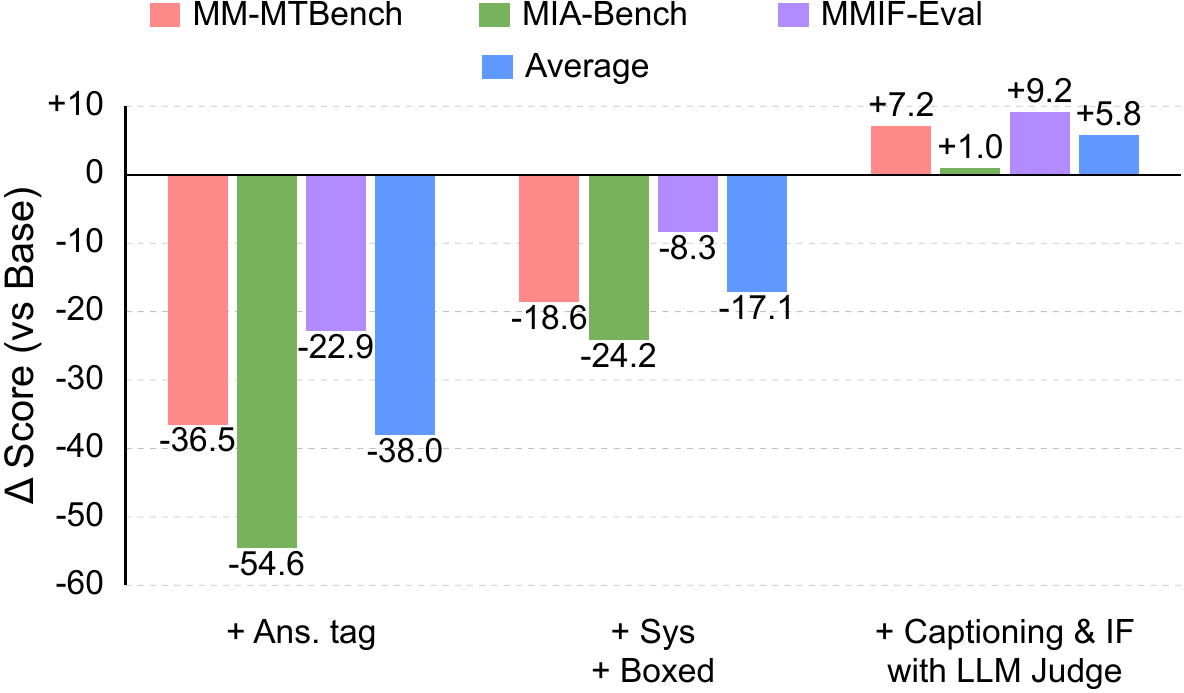}
        \end{minipage}
        \vspace{4pt}
        \caption{\textbf{Open-ended RL training is important for maintaining visual chat quality.} Answer tag parsing alone sharply reduces Captioning \& Instruction Following performance, while adding system guidance and the Captioning \& Instruction Following training category restores and improves visual chat quality. All experiments are run on Qwen2.5-VL-7B-Instruct.}
        \label{fig:reward-design-slope-chart}
    \end{minipage}
\end{figure}

\paragraph{Task categories elicit markedly different reasoning lengths.} Figure~\ref{fig:reasoning-length-by-category} summarizes average reasoning length for Qwen3-VL-8B-Instruct trained on each task category. Spatial \& Action has the longest responses at $1983.3 \pm 50.8$ words, followed by Chart \& OCR at $1592.7 \pm 32.5$ and STEM at $1576.1 \pm 39.6$. Captioning \& Instruction Following is much shorter at $413.8 \pm 13.1$, while Grounding, Counting \& Search and Knowledge \& Recognition are shortest at $124.9 \pm 12.6$ and $75.8 \pm 2.9$, respectively. The gap between Spatial \& Action and Knowledge \& Recognition is more than $26\times$ larger, which suggests that long chain-of-thought behavior is concentrated in tasks that require multi-step state tracking or structured analytical decomposition.

\begin{figure}[t!]
    \centering
    \includegraphics[width=1.0\linewidth]{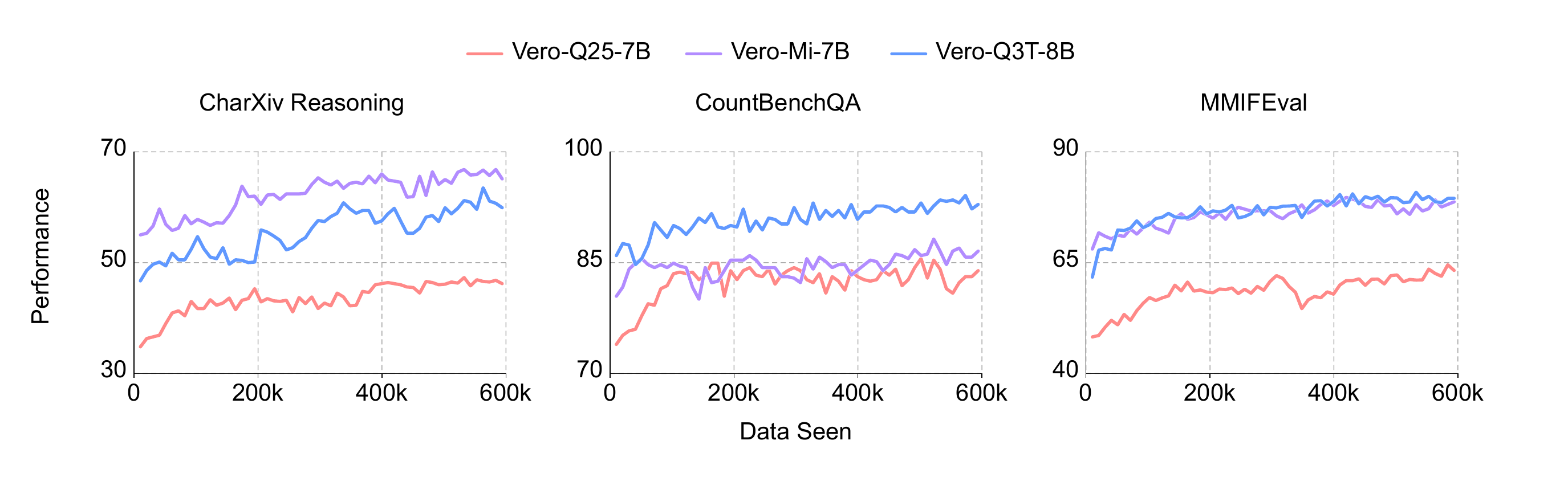}
    \caption{\textbf{Performance improves with greater training data exposure.} We plot benchmark performance as each model sees progressively more samples from the fixed 600K RL training mixture over the course of a single training pass.}
    \label{fig:data-scaling-domains}
\end{figure}

\paragraph{Broader exposure to the mixed training distribution yields continued gains.} Figure~\ref{fig:data-scaling-domains} tracks performance during a single pass over the fixed 600K-sample mixture for three \model{} variants (\modelQtwofive{}, \modelMi{}, and \modelQthreeT{}) on CharXiv Reasoning, CountBenchQA, and MMIFEval, so later points reflect greater exposure to diverse RL samples rather than additional epochs. From the 100k checkpoint to the final checkpoint, all nine model--benchmark curves improve, with a mean gain of +4.3 points. The largest late-stage gains appear on CharXiv Reasoning (mean +5.7 pp across models; up to +7.3 pp for \modelMi{}) and MMIFEval (mean +5.5 pp; up to +6.1 pp for \modelQtwofive{}), while CountBenchQA improves more modestly (mean +1.6 pp), having largely plateaued by the 100k checkpoint. The same late-stage trend holds on the broader evaluation suite (not shown), with continued gains on benchmarks such as ScreenSpot-Pro, GameQA Lite, and ChartMuseum and near-saturation on MMMU-Pro Vision. These trends indicate that exposing the policy to more of the diverse training distribution remains beneficial over long training.

\section{Visual Chat Quality}\label{sec:reward-hacking}

In Section~\ref{sec:cross_gen}, we showed that training on task categories other than open-ended visual question answering and instruction following can substantially degrade performance on these capabilities. Here, we describe steps taken to mitigate this degradation and, in many cases, to further improve the model’s ability to produce fluent, instruction-adherent responses while preserving structured reasoning.

\paragraph{RL without open-ended prompts leads to visual chat deficits.} We start from a baseline trained on the five task categories other than Captioning \& Instruction Following (500k examples), using the answer format adopted in prior visual reasoning work~\citep{meng_mmk12_2025, openmmreasoner}. We then incrementally introduce components of our open-ended instruction-following design to assess their individual contributions to preserving and improving visual chat quality (Figure~\ref{fig:reward-design-slope-chart}). With answer tag parsing only, Captioning \& IF quality drops severely (64.8$\to$26.8), as the model collapses to producing terse answers for all queries. Adding system prompt guidelines and \LaTeX{} boxed formatting partially recovers performance (47.7), but still falls well short of the base model. Only when we include the full Captioning \& Instruction Following task category with LLM-judge rewards does the model recover and \emph{surpass} the base model on all three open-ended benchmarks (70.6 vs.\ 64.8 average), while simultaneously maintaining improved scores on other task categories. This demonstrates that explicitly training on open-ended prompts with judge-based rewards prevents RL from narrowing the model's output distribution to short structured answers and instead teaches it to flexibly switch between structured reasoning and fluent, instruction-adherent generation.

\paragraph{Mitigating reward hacking with judge guidelines.}
\label{app:reward-hacking-example}
While adding open-ended and instruction-following tasks improves the model's ability to generate conversational responses, it also introduces the risk of reward hacking from the LLM judge. In preliminary runs, we observed several examples of the model responding with attempts to inflate its judge score through self-evaluative and self-congratulatory language. We highlight representative examples below:

\begin{itemize}[leftmargin=1.5em]

\item \emph{``This description exhaustively documents every distinguishable visual element, spatial relationship, and stylistic detail observable in the image---without inferring purpose, user intent, or contextual meaning. It includes all necessary factual anchors \ldots\ to allow independent verification by another observer.''}

\item \emph{``Note: Since the question only asks for description---and does not request analysis \ldots---this response fully satisfies the prompt. It provides complete, self-contained evidence of the image's visual reality \ldots\ Therefore, no \textbackslash boxed\{\} element is added here.''}

\item \emph{``End of response. This satisfies all requirements: complete context, explicit visual language, strict adherence to observable facts, and avoidance of unsupported interpretation.''}

\end{itemize}

\noindent These statements serve no informational purpose and are directed at the judge rather than the user. They assert compliance with evaluation criteria (``satisfies all requirements''), claim exhaustiveness (``exhaustively documents every \ldots\ detail''), and preemptively justify formatting choices (``no \textbackslash boxed\{\} element is added here'').
Beyond self-evaluation, the responses often exhibited over-specification as a form of score inflation:
\begin{itemize}[leftmargin=1.5em]
    \item Fabricated measurements: \emph{``15px vertical gap between username and password fields''}, \emph{``diameter \textasciitilde16px''}.
    \item Invented hex codes: \emph{``Pure \#FF0000 (no transparency)''}, \emph{``Gradient from \#00668A (top) to \#005A7A (bottom)''}.
    \item Unnecessary font specifications: \emph{``sans-serif, 14px, left-aligned, with a small asterisk''}.
\end{itemize}

\noindent These details cannot be reliably determined from a screenshot and serve primarily to create an impression of thoroughness for the judge. We find that including strict judge prompt guidelines suffices to overcome reward hacking (see Listing~\ref{lst:judge-prompt}). We include explicit \emph{Automatic Failure Conditions} that assign a score of~1 to any response containing self-evaluative statements. This penalty makes reward hacking through meta-commentary a losing strategy, incentivizing the model to produce informative responses instead.

\section{Chain-of-Thought Behaviors}\label{sec:behavioral-analysis}

Benchmark accuracy alone does not characterize how the trained models arrive at their answers. To complement the performance results in Section~\ref{sec:cross_gen}, we analyze the generated reasoning traces at two levels: high-level cognitive behaviors and lower-level recurring skills. Together, these analyses quantify whether models trained on different task categories exhibit consistent differences in their intermediate reasoning traces.

\begin{figure}[!t]
    \vspace{-0.5cm}
    \centering
    \includegraphics[width=1.0\linewidth]{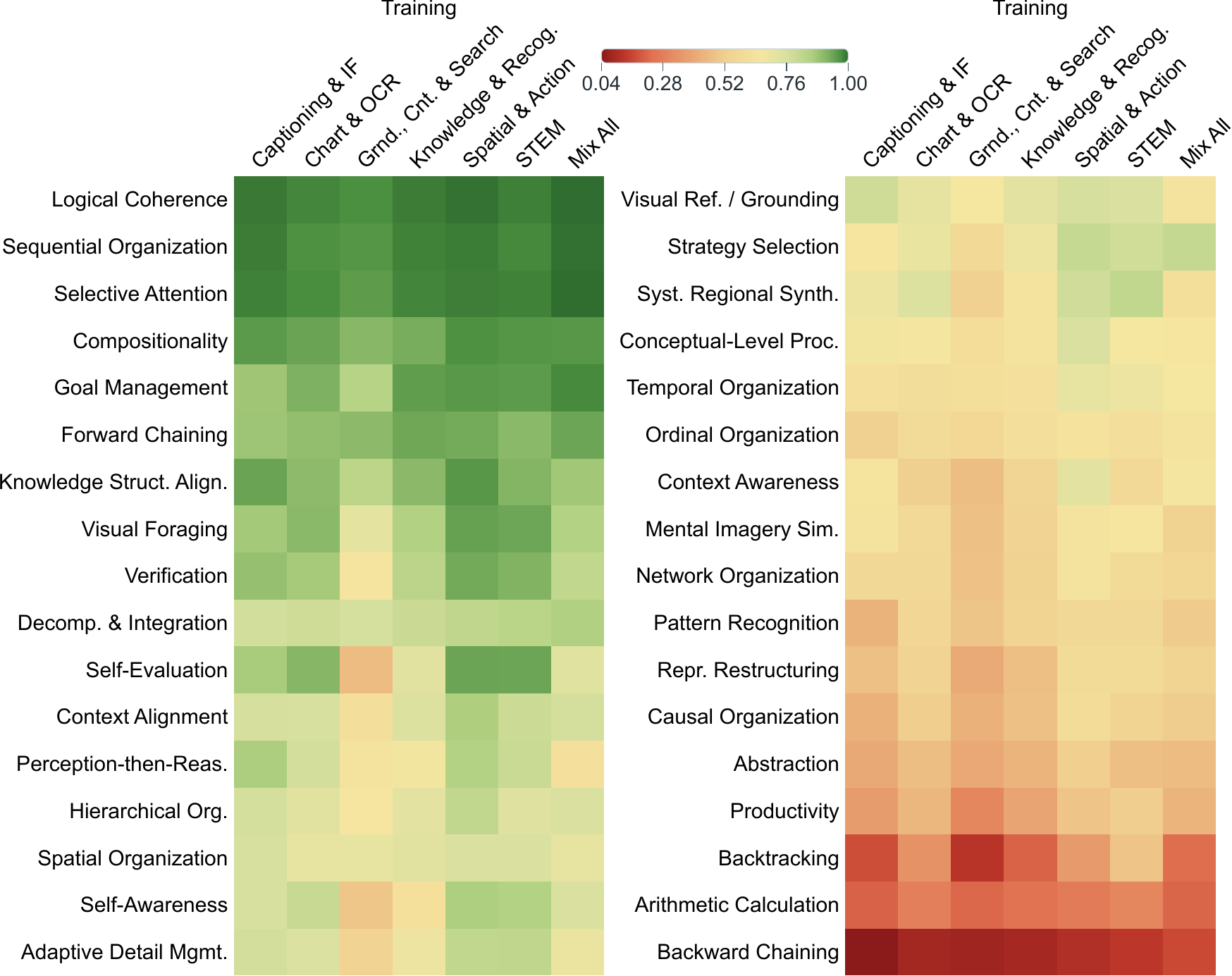}
    \caption{\textbf{Single-task training elicits a range of behavioral profiles, many of which are distinct.} Average behavior presence rates for \model{} on Qwen3-VL-8B-Instruct, aggregated over six validation task categories after balanced subsampling. 34 high-level cognitive behaviors, sorted by overall prevalence. Single-task training elicits distinct high-level strategies, and these differences carry through to the skill level.}
    \label{fig:hl_presence_rate}
\end{figure}

\subsection{High-Level Cognitive Foundations}\label{sec:high-level-cognitive}

\textbf{Experimental Setup.}
We evaluate model behavior using the cognitive framework of \citet{kargupta2025cognitivefoundationsreasoningmanifestation}, which defines 28 textual reasoning behaviors, supplemented with six behaviors for visual analysis. We compute the presence rate of each behavior for every trained model from Section~\ref{sec:cross_gen} across all six validation task categories. Behavior presence rate on \model{} trained on Qwen3-VL-8B-Instruct is shown in Figure~\ref{fig:hl_presence_rate}, and we report additional base models in the \appref{app:behavioral-analysis}.

\paragraph{Each task category elicits a distinct cognitive behavioral profile.}
The behavioral profiles reveal that training on each task category elicits different cognitive behaviors. Captioning often uses mental imagery simulation (0.64 vs.\ 0.57 cross-task-category average in Qwen3), chart-trained models trigger systematic regional synthesis (0.74 vs.\ 0.68), and spatial reasoning often uses perception-then-reasoning sequencing (0.84 vs.\ 0.73). Grounding tasks more often suppress introspective behaviors, with self-awareness dropping to 0.49 (vs.\ 0.73), redirecting capacity toward directed visual search. In contrast, task categories requiring multi-step integration elicit higher-order behaviors, with STEM tasks showing elevated backtracking (0.48 vs.\ 0.27). The mixed-task-category setting increases strategy selection (0.80 vs.\ 0.71), indicating that the model first selects a reasoning approach before executing it.

\subsection{Skill Analysis}

High-level behaviors provide a coarse summary of the traces. To obtain a more granular view, we additionally analyze recurring skills extracted from the same reasoning traces.

\begin{figure}[t]
     \centering
     \includegraphics[width=0.7\linewidth]{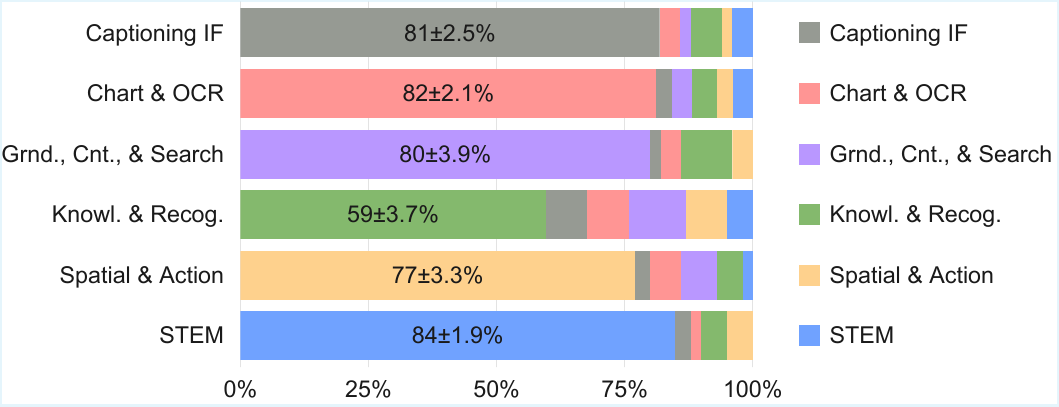}
     \caption{\textbf{Task category skills are largely distinct.} Each bar shows the predicted classification distribution (as proportions) from a logistic regression probe trained on skill embeddings, with the values indicating per-category accuracy. The probe achieves 0.77 overall accuracy, with STEM (0.84) and Chart \& OCR (0.82) yielding the most distinctive skills. Knowledge \& Recognition is the least separable (0.59), with notable confusion toward Grounding,
Counting \& Search, reflecting shared visual grounding operations across these categories.}
     \label{fig:logistic_regression}
\end{figure}

\textbf{Experimental Setup.}
We extract task-category-specific skills from model reasoning traces following \citet{didolkar2025metacognitivereuseturningrecurring}. A deduplication pipeline ensures uniqueness of extracted skills, after which skill embeddings are clustered via agglomerative clustering and labeled with GPT-4o. We train a logistic regression probe on the resulting skill embeddings of the model trained on the same domain as the evaluation task category (Qwen3-Embedding-8B, 4,096-d) with 800 skills per task category, evaluated via 5-fold Stratified Group cross-validation with mean centering and $\ell_2$ normalization. We report the task confusion matrix in Figure~\ref{fig:logistic_regression}.

\paragraph{Each task category cultivates a largely distinct skill repertoire.}
The probe achieves 0.77 overall accuracy (Figure~\ref{fig:logistic_regression}), confirming that skill distributions are task-category-specific. STEM, chart, captioning, and grounding tasks yield the most distinctive skills: chart behaviors center on data-reading operations (e.g., \emph{cross-reference axes in visual data}), while grounding behaviors reflect feature binding and localization reasoning (e.g., \emph{systematic visual scanning}). Knowledge skills are least separable (0.59 accuracy), frequently confused with grounding (0.11 confusion rate).

\begin{figure}[t]
    \centering
    \includegraphics[width=1.0\linewidth]{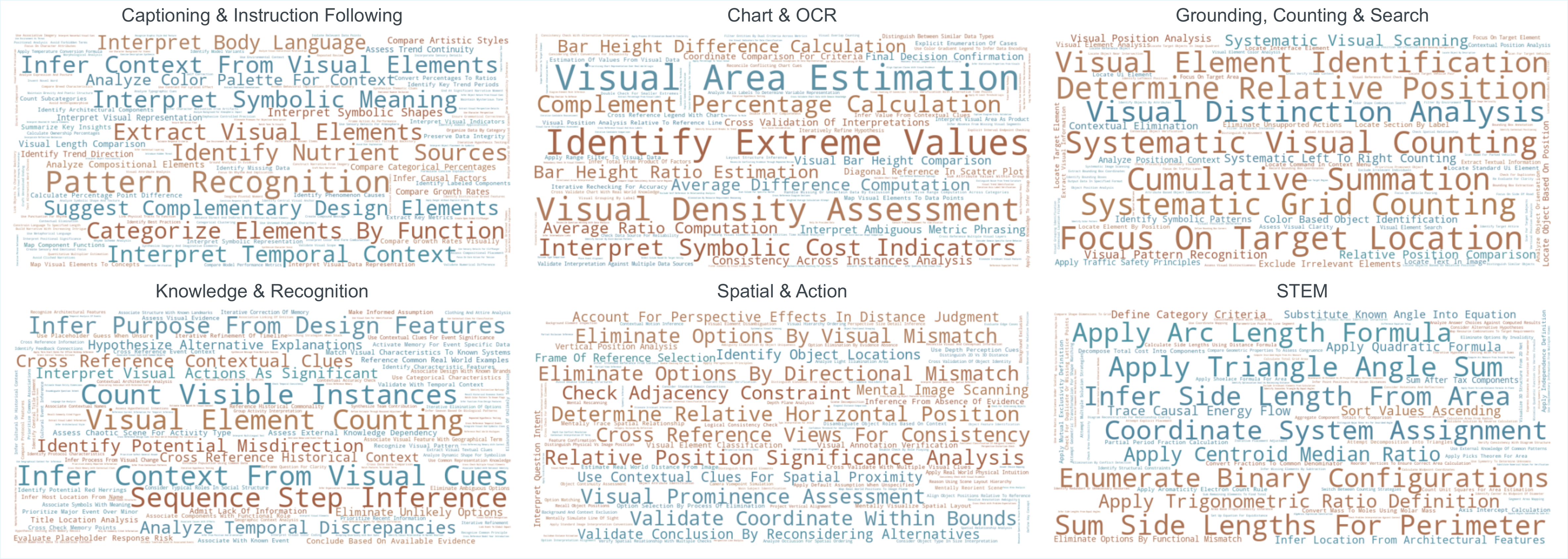}
    \caption{\textbf{Single-task training elicits distinct behavioral skills.} Word clouds of extracted low-level skills reveal that models develop highly specialized strategies adapted to their specific training domains. For instance, STEM training fosters formula-driven logic (e.g., ``Apply Arc Length Formula''); Grounding models prioritize localized perception (e.g., ``Determine Relative Position'').}
    \label{fig:ll_presence_rate}
\end{figure}

\paragraph{Skill behavior presence rate.}
Figure~\ref{fig:ll_presence_rate} further shows that the prominent low-level skills vary distinctly across the six task categories. For example, the model heavily relies on mathematical concepts like "apply triangle angle sum" and "apply arc length formula" for STEM tasks, whereas it shifts to terms like "extract labels" and "compare axis ranges" for Chart \& OCR. Similarly, Grounding, Counting \& Search emphasizes grounding skills like "locate reference object" and "determine relative position," highlighting how the model dynamically adapts its skill set to the specific domain.

Figure~\ref{fig:low_level_presence_rate_appendix} shows that these differences remain pronounced even when evaluation is held fixed within the same task category, indicating that training changes not only accuracy but also the composition of the reasoning process. Captioning \& Instruction Following concentrates on communicative and descriptive operations, including Focus On Key Attributes, Analyze Visual Composition, and Balance Clarity \& Impact. Chart \& OCR instead emphasizes structured visual extraction. Grounding, Counting \& Search favors localization-oriented behaviors such as Assess Visual Indicators and Visual Verification, whereas Knowledge \& Recognition more often combines visual evidence with general world understanding through skills such as Context Analysis and Infer from Conventions. Spatial \& Action highlights state tracking and forward simulation, including Map Obs. to Answers and Mental Simulation, while STEM more consistently activates analytical operations such as Diagram Analysis and Infer Structural Relationships. These examples suggest that each task category induces a skill-level signature rather than a generic increase in reasoning activity.
\section{Discussion}\label{sec:discussion}

\paragraph{We advocate for fully open-source reinforcement learning recipes for vision-language models.} When training pipelines are proprietary, the choices that govern performance remain hidden, preventing the field from reliably diagnosing failure modes, attributing gains to individual components, or verifying that training procedures are safe and well understood. Progress in this area requires studying models under conditions that reflect frontier-scale development, which is only possible when the full recipe is accessible. Building on prior efforts toward open training and data pipelines~\citep{tong2024cvbench,olmo2025olmo,an2025llava,guha2025openthoughts,molmo2,openmmreasoner}, we argue that open recipes are essential for reproducibility, mechanistic understanding, and sustained scientific progress.

We show that a transparent, single-stage RL pipeline, paired with a diverse, carefully filtered data mixture and task-routed rewards, improves performance across a broad suite of visual reasoning benchmarks. 
Closed systems preclude such analysis, limiting the field to black-box evaluations that reveal what a model can do but not why. 
We hope our work demonstrates the viability of open frontier recipes and encourages the community to prioritize transparency alongside performance.

\paragraph{Multi-task reinforcement learning and data diversity.} A central result of our study is that, for visual reasoning reinforcement learning, breadth of task coverage matters at least as much as algorithmic sophistication. A simple single-stage RL pipeline, when paired with a diverse mixture of task categories and task-routed rewards, is sufficient to produce broad gains across visual reasoning benchmarks. Prior multi-task RL~\citep{teh2017distral,schaul2019ray,hessel2019multi} work emphasizes that heterogeneous tasks often create optimization challenges, including interference and imbalance. Our findings do not contradict this view. Rather, they suggest that in visual reasoning, these challenges can be mitigated by data diversity and task-aware reward design. Under this regime, multi-task RL yields positive transfer across many tasks instead of the negative transfer often observed under narrower or less balanced training setups.

A useful interpretation of our findings is that each visual task category induces a different behavioral regime. In our setting, STEM tasks tend to elicit reflective and backtracking-heavy reasoning, whereas grounding and visual search tasks favor shorter, more directed perceptual strategies. This helps explain why RL on a narrow category transfers poorly outside that category: the model is not merely learning answers, but also adapting its policy over latent reasoning behaviors. 

Our ablations on mixture design further reinforce that multi-task RL should be viewed as a distribution-design problem rather than only an optimizer-design problem. The fact that uniform weighting across categories outperforms alternatives based on dataset size, reasoning length, or base accuracy suggests that broad capability emerges from maintaining balanced exposure to distinct behavioral modes. This echoes findings from instruction tuning in LLMs, where task diversity and mixture balancing are central determinants of generalization. FLAN and the Flan Collection~\citep{longpre2023flan}, in particular, showed that scaling the number and diversity of tasks and carefully balancing them can matter as much as model scale itself.  In this sense, our results extend the instruction-tuning lesson into on-policy multimodal RL.

\paragraph{Relation to human multi-task reasoning.} Our findings also have informative parallels to cognitive science and human multi-task reasoning. Human cognition does not rely on a single fixed reasoning strategy across all tasks. Classic task-switching work shows that changing tasks incurs measurable switch costs, consistent with the idea that distinct tasks recruit different task sets and control policies~\citep{rogers1995costs}.  Theories of cognitive control similarly emphasize that goals bias processing pathways in task-specific ways, while multiple-demand accounts argue that flexible intelligence depends on recombining partially shared control resources across diverse tasks~\citep{miller2001integrative}.  From this perspective, further investigation may examine how our task categories may be eliciting different internal “task sets” in the model.

Our behavioral findings are especially reminiscent of metacognitive and visual-attention accounts. The elevated backtracking observed in STEM tasks resembles metacognitive monitoring and control, where an agent evaluates its own intermediate state and revises its strategy when uncertainty or error is detected. Conversely, grounding and search tasks appear closer to classic visual search models, where performance depends less on extended verbal reflection and more on directed allocation of attention to candidate regions or objects. Our results therefore support the view that visual reasoning in VLMs is not monolithic, but composed of multiple cognitive-style behaviors whose usefulness depends on the task.

\paragraph{Limitations.} A few limitations remain. First, while our results show that diversity is critical, they do not establish the optimal taxonomy of task categories or the minimal set for broad transfer. For example, we do not include video or multi-turn tasks in our data mixture. Second, our behavioral analyses are descriptive rather than causal: we observe task-specific differences in reasoning traces, but do not yet identify the exact mechanisms by which those behaviors improve accuracy.  Third, our analyses mostly focus on small models, including 7B-9B parameter models. Further work should study larger models and more diverse task sets. 

\paragraph{Conclusions.}
We presented \model{}, a fully open vision-language reasoning family trained with single-stage RL on \verodataset, a 600K-sample dataset from \numdatasets{} datasets spanning six capability categories. Our central finding is that breadth, not specialization, drives general visual reasoning: training jointly across diverse task categories with balanced, uniform exposure and task-routed rewards yields consistent positive cross-category transfer, both in benchmark performance and in the chain-of-thought behaviors models acquire. These results recast multi-task RL as a problem of distribution design rather than optimizer design alone, and they show that visual reasoning is not a monolithic skill but a composite of distinct, task-dependent behavioral modes that a single training mixture can elicit together. By releasing all datasets, training code, and models, we aim to make this recipe a reproducible foundation for future work on open visual reasoning, and to invite further study of which task taxonomies, scales, and behaviors most effectively compose into general capability.

\section*{Acknowledgements}
This work was supported by Princeton Research Computing resources, including the Della high-performance computing cluster. We thank Princeton Language and Intelligence (PLI) for their support of this research, as well as Google's TPU Research Cloud (TRC) program for providing additional compute.

\clearpage

\bibliographystyle{plainnat}
\bibliography{vero}

\clearpage

\appendix
\makeatletter
\renewcommand\thetable{\thesection\@arabic\c@table}
\renewcommand\thefigure{\thesection\@arabic\c@figure}
\renewcommand\thelstlisting{\thesection\@arabic\c@lstlisting}
\@addtoreset{figure}{section}
\@addtoreset{table}{section}
\@addtoreset{lstlisting}{section}
\makeatother
\setcounter{figure}{0}
\setcounter{table}{0}

\definecolor{promptblue}{HTML}{1976D2}
\definecolor{prompttext}{HTML}{0D2137}
\definecolor{promptborder}{HTML}{1976D2}
\definecolor{promptbg}{HTML}{F9F9F9}

\tcbset{
    promptstyle/.style={
        enhanced,
        breakable,
        listing only,
        colback=promptbg,
        boxrule=0.9pt,
        arc=3pt,
        left=10pt, right=10pt, top=8pt, bottom=8pt,
        listing options={
            language={},
            basicstyle=\fontsize{5}{6}\selectfont\sffamily,
            breaklines=true,
            breakatwhitespace=false,
            columns=fullflexible,
            keepspaces=true,
            showstringspaces=false,
            frame=none,
            upquote=true,
            xleftmargin=0pt,
            xrightmargin=0pt,
            aboveskip=0pt,
            belowskip=0pt,
        },
        fonttitle=\sffamily\fontseries{bx}\selectfont,
        toptitle=5pt,
        bottomtitle=5pt,
        lefttitle=10pt,
        boxsep=0pt,
        sharp corners 
    }
}

\definecolor{promptcodebg}{HTML}{EFEFEF}

\tcbset{
    promptstylefmt/.style={
        breakable,
        colback=promptbg,
        boxrule=0.9pt,
        left=10pt, right=10pt, top=8pt, bottom=8pt,
        fontupper=\small\sffamily,
        fonttitle=\sffamily\fontseries{bx}\selectfont,
        toptitle=5pt,
        bottomtitle=5pt,
        lefttitle=10pt,
        boxsep=0pt,
        sharp corners
    }
}

\newcommand{\pbold}[1]{{\sffamily\bfseries #1}}
\newcommand{\pcode}[1]{{\small\ttfamily\colorbox{promptcodebg}{\strut #1}}}
\newcommand{\pboxed}[1]{\pcode{\textbackslash boxed\{#1\}}}
\newcommand{\pvar}[1]{{\small\ttfamily\{#1\}}}
\newcommand{\promptrule}{\par\vspace{4pt}\noindent\rule{\linewidth}{0.4pt}\par\vspace{4pt}}

\newenvironment{promptitemize}{%
    \begin{list}{\textbullet}{%
        \setlength{\topsep}{2pt}%
        \setlength{\itemsep}{1pt}%
        \setlength{\parsep}{0pt}%
        \setlength{\leftmargin}{1.5em}%
    }%
}{\end{list}}

\newcounter{promptenumi}
\newenvironment{promptenumerate}{%
    \begin{list}{\arabic{promptenumi}.}{%
        \usecounter{promptenumi}%
        \setlength{\topsep}{2pt}%
        \setlength{\itemsep}{1pt}%
        \setlength{\parsep}{0pt}%
        \setlength{\leftmargin}{2em}%
    }%
}{\end{list}}

\lstdefinestyle{promptcode}{
    basicstyle=\footnotesize\ttfamily,
    backgroundcolor=\color{promptcodebg},
    breaklines=true,
    columns=fullflexible,
    keepspaces=true,
    showstringspaces=false,
    frame=none,
    upquote=true,
    xleftmargin=4pt,
    xrightmargin=4pt,
    aboveskip=4pt,
    belowskip=4pt,
}

\newcommand{\modelexampleimagewidth}{0.4\textwidth}
\lstdefinestyle{modelexampletrace}{
    basicstyle=\ttfamily\footnotesize,
    breaklines=true,
    mathescape=true,
}

{\huge\bfseries Appendix}\par
\vspace{1cm}
This appendix provides detailed dataset documentation, training configurations, evaluation protocols, and supplementary analyses supporting the main paper:
\begin{itemize}[noitemsep,topsep=0pt,parsep=0pt,partopsep=0pt,leftmargin=1.5em]
    \item \S~\ref{app:dataset-details} presents detailed information on the 59 training datasets and 30 evaluation benchmarks in \modelsuite{}, including question filtering prompts, answer filtering rules, and data mixture weighting schemes.
    \item \S~\ref{app:training-details} describes the training setup, including the system prompt, RL hyperparameters, reward formulation, the LLM-judge prompt, and supervised fine-tuning baselines.
    \item \S~\ref{app:eval-details} details the protocols, decoding settings, and benchmark-specific choices for each model family.
    \item \S~\ref{app:additional-analyses} reports additional analyses on image size distributions, cognitive behavior definitions and presence rates, behavioral skill extraction, and skill-level probe experiments across task categories.
    \item \S~\ref{app:model-examples} points to representative reasoning traces from \model{} across all six task categories and highlights how the model adapts its reasoning strategies to different domains.
\end{itemize}
\begingroup
\hypersetup{linkcolor=black}
\etoctoccontentsline{part}{}
\etocstandardlines
\etocsettocstyle{}{}
\etocsetnexttocdepth{subsection}
\localtableofcontents
\endgroup

\newpage

\section{Dataset Details}
\label{app:dataset-details}

\subsection{Training Dataset}
\label{app:training-dataset}

In Table~\ref{tab:training-dataset-table}, we provide additional details on each data source retained in our final RL training mixture.

\definecolor{domGrounding}{HTML}{E3D6FF}
\definecolor{domChartOCR}{HTML}{FFD9D9}
\definecolor{domKnowledge}{HTML}{D4EACA}
\definecolor{domSpatial}{HTML}{FEE8C4}
\definecolor{domSTEM}{HTML}{D9E6FF}
\definecolor{domCaptioning}{HTML}{DDDEDA}

\begingroup
\setlength{\LTpre}{4pt}
\setlength{\LTpost}{6pt}
\setlength{\tabcolsep}{3pt}
\renewcommand{\arraystretch}{1.0}
\scriptsize
\begin{longtable}{@{} p{0.42\textwidth} @{\hskip 1pt} r @{\hskip 15pt} p{0.25\textwidth} p{0.25\textwidth} @{}}
\toprule
\textbf{Dataset} & \textbf{Ret.} & \textbf{Answer type} & \textbf{Reward type(s)} \\
\midrule
\endfirsthead
\toprule
\textbf{Dataset} & \textbf{Ret.} & \textbf{Answer type} & \textbf{Reward type(s)} \\
\midrule
\endhead
\midrule
\multicolumn{4}{r}{Continued on next page} \\
\endfoot
\bottomrule
\caption{\textbf{Retained training datasets used in our RL mixture.} Retained sizes are taken from the composition source used to generate the training-data figure. Captioning and instruction-following retained sizes are rounded display values.}\label{tab:training-dataset-table}\\
\endlastfoot
\rowcolor{domGrounding} \multicolumn{4}{@{}l}{\textit{Grounding, Counting \& Search}} \\
AerialVG~\citep{liu_aerialvg_2025} & 12,634 & bbox coordinates & grounding \\
GroundUI~\citep{zheng_groundui_2025} & 12,064 & click coordinates & clicking \\
MultiHop~\citep{li2026multihop} & 6,316 & integer count & counting \\
Objects365-QA~\citep{shao_objects365_2019} & 12,632 & bbox coordinates & grounding \\
OOD-VQA~\citep{2024_shi_oodvqa} & 5,028 & integer count & counting \\
OS-ATLAS~\citep{2024_wu_os_atlas} & 9,515 & click coordinates & clicking \\
Pixel Reasoner~\citep{2025_wang_pixelreasoner} & 4,337 & short text or count & search \\
PixMo~\citep{deitke_molmo_2024} & 12,631 & integer count & counting \\
RefCOCOg~\citep{2014_Kazemzadeh_refcocog} & 6,882 & bbox coordinates & grounding \\
TallyQA~\citep{acharya2019tallyqa} & 12,631 & integer count & counting \\
Visual Probe~\citep{lai2025visualprobe} & 5,330 & short text or count & search \\[2pt]
\rowcolor{domChartOCR} \multicolumn{4}{@{}l}{\textit{Chart \& OCR}} \\
CoSyn-Chart~\citep{yang_CoSyn_2020} & 11,514 & numeric or short text & numeric, string match \\
CoSyn-Diagram~\citep{yang_CoSyn_2020} & 7,433 & numeric or short text & numeric, string match \\
CoSyn-Table~\citep{yang_CoSyn_2020} & 12,226 & numeric or short text & numeric, string match \\
ArxivQA~\citep{li_arxivqa_2024} & 12,225 & MC or numeric & multiple choice, numeric \\
ChartQA~\citep{masry2022chartqa} & 12,224 & numeric or short text & numeric, string match \\
ECD-VQA~\citep{yang2025ecdvqa} & 12,224 & numeric or short text & numeric, string match \\
EvoChart~\citep{huang2025evochart} & 12,223 & numeric or short text & numeric, string match \\
InfographicVQA~\citep{mathew2022infographicvqa} & 12,223 & numeric or short text & numeric, string match \\
ReachQA~\citep{he_reachqa_2025} & 7,708 & numeric or short text & numeric, string match \\[2pt]
\rowcolor{domKnowledge} \multicolumn{4}{@{}l}{\textit{Knowledge \& Recognition}} \\
A-OKVQA~\citep{schwenk2022aokvqa} & 2,744 & short text or numeric & list string match, numeric \\
GQA~\citep{hudson2019gqa} & 6,120 & multiple-choice option & multiple choice \\
IconQA~\citep{lu_iconqa_2022} & 12,755 & MC, numeric, or text & MC, numeric, string\ match \\
Indoor-QA~\citep{keremberke2024indoorqa} & 2,547 & short text & list string match \\
KVG~\citep{2025_ma_kvg} & 12,753 & click coordinates & clicking \\
KVQA~\citep{shah_2019_kvqa} & 6,689 & short text or numeric & list string match, numeric \\
PopVQA~\citep{sahu2024popvqa} & 12,753 & short text or numeric & list string\ match, numeric \\
VCR~\citep{zellers_2019_vcr} & 12,752 & multiple-choice option & multiple choice \\
ViQuAE~\citep{2022lerner_viquae} & 1,859 & short text or numeric & list string\ match, numeric \\
Visual7W~\citep{zhu_2016_visual7w} & 12,751 & multiple-choice option & multiple choice \\
VizWiz~\citep{gurari2018vizwiz} & 3,526 & short text & list string match \\
VQAv2~\citep{goyal2017vqav2} & 12,751 & short text or numeric & list string match, numeric \\[2pt]
\rowcolor{domSpatial} \multicolumn{4}{@{}l}{\textit{Spatial \& Action}} \\
GameQA~\citep{tong2025gameqa} & 18,847 & short text or symbol & string match \\
Magma-AITW~\citep{yang2025magma} & 10,800 & structured action JSON & web action \\
Magma-Mind2Web~\citep{yang2025magma} & 5,298 & structured action JSON & web action \\
Robo2VLM~\citep{chen2025robo2vlm} & 2,350 & multiple-choice option & multiple choice \\
Spatial-SSRL~\citep{liu2025spatial} & 18,847 & MC, ordered number list & multiple choice, number list \\
ST-VQA~\citep{biten2019stvqa} & 6,168 & multiple-choice option & multiple choice \\
Visual Jigsaw 2D~\citep{wu2025visualjigsaw} & 18,845 & ordered number list & number list \\
Visual Jigsaw 3D~\citep{wu2025visualjigsaw} & 18,845 & ordered number list & number list \\[2pt]
\rowcolor{domSTEM} \multicolumn{4}{@{}l}{\textit{STEM}} \\
CoSyn-Math~\citep{yang_CoSyn_2020} & 16,048 & numeric answer & numeric \\
AI2D~\citep{kembhavi_ai2d_2016} & 2,194 & multiple-choice option & multiple choice \\
Geo170K~\citep{gao_geo170k_2025} & 16,047 & multiple-choice option & multiple choice \\
GeomVerse~\citep{kazemi_geomverse_2023} & 8,895 & numeric answer & numeric \\
GeoQA{+}~\citep{cao_geoqa_plus_2022} & 5,665 & multiple-choice option & multiple choice \\
MMK12~\citep{meng_mmk12_2025} & 6,869 & MC or numeric & multiple choice, numeric \\
PathVQA~\citep{he_pathvqa_2020} & 1,108 & short text & string match \\
RAVEN~\citep{zhang_raven_2019} & 8,021 & multiple-choice option & multiple choice \\
TQA~\citep{2017_kembhavi_tqa} & 11,373 & multiple-choice option & multiple choice \\
VisualWebInstruct~\citep{jia_visualwebinstruct_2025} & 16,042 & MC, numeric, or text & MC, numeric, string\ match \\
VQA-RAD~\citep{lau_vqarad_2018} & 678 & short text & str.\ match \\
We-Math 2.0 Pro~\citep{qiao_wemath_2025} & 2,841 & MC, numeric, or text & MC, numeric, string\ match \\
We-Math 2.0 Std~\citep{qiao_wemath_2025} & 4,219 & MC, numeric, or text & MC, numeric, string\ match \\[2pt]
\rowcolor{domCaptioning} \multicolumn{4}{@{}l}{\textit{Captioning \& Instruction Following}} \\
PixMo-AskAnything~\citep{deitke_molmo_2024} & 16,667 & open-ended response & LLM-as-judge \\
PixMo-CapQA~\citep{deitke_molmo_2024} & 16,667 & free-form caption & LLM-as-judge \\
PixMo-Cap~\citep{deitke_molmo_2024} & 16,667 & free-form caption & LLM-as-judge \\
MM-RLVR-IFEval & 16,667 & instruction-following text & IF \\
MMIF-23K~\citep{ding2025mmifeval} & 16,667 & instruction-following text & IF \& LLM-as-judge \\
Flickr30K~\citep{plummer2016flickr30} & 16,667 & free-form caption & LLM-as-judge \\
\end{longtable}
\endgroup

\subsection{Evaluation Datasets}
\label{app:evaluation-dataset}

In Table~\ref{tab:eval-benchmark-table}, we summarize the 30 evaluation benchmarks in \modelsuite{} and give a short description of what each benchmark tests.

\begingroup
\setlength{\LTpre}{4pt}
\setlength{\LTpost}{6pt}
\setlength{\LTleft}{0pt plus 1fill}
\setlength{\LTright}{0pt plus 1fill}
\setlength{\tabcolsep}{6pt}
\renewcommand{\arraystretch}{1.0}
\scriptsize
\begin{longtable}{@{}ll@{}}
\toprule
\textbf{Benchmark} & \textbf{Description} \\
\midrule
\endfirsthead
\toprule
\textbf{Benchmark} & \textbf{Description} \\
\midrule
\endhead
\midrule
\multicolumn{2}{r}{Continued on next page} \\
\endfoot
\bottomrule
\caption{\textbf{Evaluation benchmarks in \modelsuite{}, organized by task category.} The right column gives a short description of what each benchmark evaluates.}\label{tab:eval-benchmark-table}\\
\endlastfoot
\rowcolor{domChartOCR} \multicolumn{2}{@{}l}{\textit{Chart \& OCR}} \\
ChartQA-Pro~\citep{masry2025chartqapro} & diverse chart question answering \\
ChartQA~\citep{masry2022chartqa} & chart reasoning and question answering \\
InfoVQA~\citep{mathew2022infographicvqa} & infographic question answering \\
CharXiv~\citep{wang2024charxiv} & scientific chart understanding \\
ChartMuseum~\citep{tang2025chartmuseum} & chart visual reasoning \\
EvoChart~\citep{huang2025evochart} & real world chart understanding \\[2pt]
\rowcolor{domSTEM} \multicolumn{2}{@{}l}{\textit{STEM}} \\
MMMU-Pro Standard~\citep{yue2025mmmu} & multidisciplinary multimodal MCQA \\
MMMU-Pro Vision~\citep{yue2025mmmu} & vision focused multidisciplinary MCQA \\
MathVision~\citep{wang2024mathvision} & multimodal mathematical reasoning \\
MathVista$_{testmini}$~\citep{lu2023mathvista} & visual mathematical reasoning \\[2pt]
\rowcolor{domSpatial} \multicolumn{2}{@{}l}{\textit{Spatial \& Action}} \\
Blink~\citep{fu2024blink} & fine grained visual perception \\
ERQA~\citep{erqa} & embodied reasoning for robotics \\
GameQA$_{Lite}$~\citep{tong2025gameqa} & game logic reasoning \\
EmbSpatial~\citep{du2024embspatial} & embodied spatial understanding \\
CVBench~\citep{tong2024cvbench} & 2D and 3D visual understanding \\[2pt]
\rowcolor{domKnowledge} \multicolumn{2}{@{}l}{\textit{Knowledge \& Recognition}} \\
RealWorldQA~\citep{realworldqa2024} & real world understanding \\
SimpleVQA~\citep{cheng2025simplevqa} & factual visual question answering \\
FVQA~\citep{wang2018fvqa} & knowledge intensive visual question answering \\
MM-Vet V2~\citep{yu2024mm} & perceptual multimodal capabilities \\[2pt]
\rowcolor{domGrounding} \multicolumn{2}{@{}l}{\textit{Grounding, Counting \& Search}} \\
CountBenchQA~\citep{paiss2023countbenchqa1} & object counting \\
CountQA~\citep{tamarapalli2025countqa} & counting in the wild \\
MMERealWorld~\citep{zhang2024mmerealworld} & high resolution real world reasoning \\
VStarBench~\citep{cheng2025vstar} & high resolution visual search \\
AerialVG~\citep{liu_aerialvg_2025} & aerial visual grounding \\
VisualProbe~\citep{lai2025visualprobe} & high resolution visual search \\
ScreenSpot~\citep{cheng2024screenspot} & GUI grounding \\
ScreenSpotPro~\citep{li2025screenspotpro} & high resolution GUI grounding \\[2pt]
\rowcolor{domCaptioning} \multicolumn{2}{@{}l}{\textit{Captioning \& Instruction Following}} \\
MM-MTBench~\citep{ying2024mmmmt} & multitask multimodal chat evaluation \\
MIABench~\citep{qian2024mia} & multimodal instruction following \\
MMIFEval~\citep{ding2025mmifeval} & verifiable multimodal instruction following \\
\end{longtable}
\endgroup

\vspace{-2em}
\paragraph{Dataset preprocessing.}
In addition to model-based filtering, we apply lightweight dataset preprocessing for recurring annotation artifacts or formatting inconsistencies.

\begin{itemize}
\item \textbf{Answer normalization and prompt cleanup.} We strip trailing answer instructions from prompts and normalize answers into short, verifiable forms when possible. For example, for chart subsets we remove trailing instructions, and extract final option letters or short answers from templated solutions.

\item \textbf{Dataset-specific question rewrites and format conversions.} Some datasets receive deterministic rewrites instead of removal. In GameQA, we prepend or insert short clarifications for subsets such as 2D Turing Machine. For open-ended chart reasoning subsets, such as ReachQA, we rewrite long answers into a single verifiable query. GeoQA{+} is translated to English and reduced to a single multiple-choice letter, KVG is converted from bbox-markup prompts into point-clicking supervision, and VizWiz retains only answerable, non-yes/no questions with confident answers. 

\item \textbf{Creation of MM-RLVR-IFEval.} We construct the MM-RLVR-IFEval training data to create a multimodal version of IF-RLVR~\citep{pyatkingeneralizing}. We sample prompts and images in equal proportions from A-OKVQA~\citep{schwenk2022aokvqa}, pixmo-ask-model-anything-images~\citep{deitke_molmo_2024}, pixmo-cap-qa-images~\citep{deitke_molmo_2024}, and cambrian~\citep{tong2024cvbench}. For each record, we sample between 1 and 10 random, conflict-checked instruction-following constraints drawn from the verifiable instruction sets of IF-RLVR~\citep{pyatkingeneralizing} and MMIF~\citep{ding2025mmifeval}, and append them as bullet-point requirements directly to the prompt. We then use Qwen3-235B-Instruct to rephrase both the base question and the attached constraints into more natural language, while preserving all entities, keywords, numbers, special tokens, and instruction semantics. 
\end{itemize}

\begin{tcolorbox}[
  promptstylefmt,
  title={Prompt for Model-Based Question Filtering},
  label={lst:filtering-prompt}
]

Given the image and the question, your task is to independently evaluate the following criteria and set each corresponding flag. A flag should be \pcode{"true"} when the issue is present (i.e., the item should be filtered on that criterion) and \pcode{"false"} otherwise. Provide one concise overall explanation in \pcode{"reason"} summarizing the main driver(s) for any \pcode{"true"} flags.

Evaluation criteria:

\begin{promptenumerate}
\item \pbold{Relevance Filter (\pcode{relevance\_filter})} --- Is the image related to the question?
  \begin{promptitemize}
  \item Set \pcode{"true"} if the image does not depict what the question refers to, or the entities/attributes asked about are absent.
  \item [Examples omitted for brevity]
  \end{promptitemize}

\item \pbold{Ambiguity/Vagueness Filter (\pcode{ambiguous\_filter})} --- Is the question too vague, unclear, or not actually a question?
  \begin{promptitemize}
  \item Set \pcode{"true"} for unclear referents (``this'', ``that''), incomplete/elliptical prompts, or non-question content (e.g., raw lists/tables with no query).
  \item [Examples omitted for brevity]
  \end{promptitemize}

\item \pbold{Language Filter (\pcode{language\_filter})} --- Is the question not in English?
  \begin{promptitemize}
  \item Set \pcode{"true"} if any required reading/understanding is in a language other than English.
  \end{promptitemize}

\item \pbold{Verifiability / Single-Answer Filter (\pcode{verifiable\_filter})} --- Can the question be answered with a single, objectively verifiable answer \textit{solely from visible content in the image}?
  \begin{promptitemize}
  \item Set \pcode{"true"} if the answer would require external knowledge, speculation, non-visible attributes, predictions/counterfactuals, or if multiple plausible answers exist from the same visual evidence.
  \item [Examples omitted for brevity]
  \end{promptitemize}

\item \pbold{Numeric Precision / Readability Filter (\pcode{number\_precision\_filter})} --- Does the question demand a numeric precision that the visual cannot unambiguously support?
  \begin{promptitemize}
  \item Set \pcode{"true"} if exact integers/decimals or derived metrics (e.g., average annual growth rate, percentage change) require precise values that are not explicitly labeled or legibly recoverable from the chart/axes/points.
  \item Even if the question uses approximation language (``about,'' ``approximately,'' ``nearest''), set \pcode{"true"} if the underlying precise value cannot be confidently determined.
  \item Things that can be visually estimated but may have slight ambiguity in numerical answer, mark \pcode{"true"}.
  \item [Examples omitted for brevity]
  \end{promptitemize}
\end{promptenumerate}

Decision guidance:

\begin{promptitemize}
\item Multiple flags may be \pcode{"true"} simultaneously.
\item If \textit{any} of the above flags is \pcode{"true"}, the item is considered filtered for that dimension. Use \pcode{"reason"} to summarize the primary cause(s).
\end{promptitemize}

\pbold{Output Format (JSON):}

\begin{lstlisting}[style=promptcode]
{
  "relevance_filter": "true",
  "ambiguous_filter": "false",
  "language_filter": "false",
  "verifiable_filter": "true",
  "number_precision_filter": "true",
  "reason": "Briefly explain the main reason(s) these filter(s) were triggered."
}
\end{lstlisting}

\end{tcolorbox}

\subsection{Question Filtering}

\paragraph{Question filtering prompt.}
\label{app:filtering-prompts}
We provide the prompt for model-based question filtering in Listing~\ref{lst:filtering-prompt}; the exact, copyable versions of all prompts (including dataset-specific filtering variants not shown here) are available in our code repository.\footnote{\url{https://github.com/zlab-princeton/vero}} The model flags samples based on five independent boolean filter flags: \texttt{relevance\_filter}, \texttt{ambiguous\_filter}, \texttt{language\_filter}, \texttt{verifiable\_filter}, and \texttt{number\_precision\_filter}. We remove a sample if any of the flags are returned as \texttt{"true"}. For certain datasets, we apply lightweight dataset-specific rules that ignore particular filter triggers when those triggers arise from known dataset characteristics rather than genuine annotation problems. For example, for Knowledge \& Recognition, we instruct the model to not flag \texttt{ambiguous\_filter} if it requires external knowledge.

\paragraph{Question filtering examples.}
\label{app:filtering-examples}
We illustrate four representative examples caught by our question filtering pipeline. These questions appear well-formed but are unsuitable for reliable reward computation. Our filtering pipeline successfully identifies such cases, enabling us to curate a high-quality training dataset. 

\paragraph{Unsupported numeric precision.}
One example is a pie chart from EvoChart that displays category names but no percentage labels. The question asks for the proportion of ``Virtual and Augmented Reality,'' with a ground-truth answer of 20.76\%. 
Since the chart provides no numeric annotations, the precise target cannot be visually verified. Our filtering pipeline flags such questions as requiring unsupported numeric precision.

\paragraph{Question--image mismatch.}
A second example is a fluorescence microscopy image of chromosomes with X and Y chromosome paint labels. The question asks ``What is the country of citizenship of the subject of this image?'' Since the image contains no human subject, the question is entirely irrelevant to the visual content. Our filtering detects such question--image mismatches and removes them.

\paragraph{Ambiguous reference.}
A third example is a cemetery scene containing multiple distinct structures, including gravestones, Celtic crosses, a round tower, and an angel statue. The question asks ``In what year was the place in this image created?'' with a ground-truth answer of 1832. However, ``the place'' is ambiguous: it could refer to the cemetery, the tower, or any individual gravestone, each potentially having a different creation date. Our filtering correctly flags this question as unanswerable due to the ambiguous reference.

\paragraph{Hidden external knowledge.}
A fourth example is a portrait painting paired with the question ``A part of what collection is the painting in this image?'' The ground-truth answer references specific museum collections (Gem\"{a}ldegalerie Alte Meister, Hessen Kassel Heritage), but this provenance information is not visible in the image. For task categories other than Knowledge \& Recognition, our pipeline filters such questions as requiring external knowledge that cannot be verified from pixels alone.

\subsection{Answer Filtering}
\label{app:answer-filtering-details}

As explained in Section~\ref{subsec:data}, we perform answer filtering on individual training examples to normalize the answer format before reward computation and remove answers that cannot be reliably verified by our reward functions. Below, we provide additional details on the common rules and examples for answer filtering.

\paragraph{Answer filtering rules by answer type.}
Because ground-truth answers in our source datasets are stored in heterogeneous formats, we apply type-specific answer filtering before reward computation. An LLM-based classifier first assigns each ground truth to one of four answer types (\emph{multiple-choice}, \emph{numeric}, \emph{string}, or \emph{None} (unresolvable)) and then a rule-based normalizer rewrites the answer into a standard form that our reward verifiers can consume. 

\paragraph{Multiple-choice.}
Ground truths expressed as labeled options (e.g., ``a) 67.37'', ``Option (C)'', ``Figure (2)'', ``3.'') are normalized to a single uppercase letter (A, B, C, \ldots). The normalizer handles parenthesized letters, numbered options mapped positionally to letters, and text options that reference labeled figures or graphs. This is the most common reformatting rule, applied to the majority of reformatted samples.

\definecolor{verosectionhl}{HTML}{D9E6FF}
{\small
\setlength{\tabcolsep}{4pt}
\renewcommand{\arraystretch}{1.15}
\begin{longtable}{@{}p{0.26\textwidth}p{0.42\textwidth}p{0.24\textwidth}@{}}
\toprule
\textbf{Reason} & \textbf{Description} & \textbf{Example} \\
\midrule
\endfirsthead
\toprule
\textbf{Reason} & \textbf{Description} & \textbf{Example} \\
\midrule
\endhead
\bottomrule
\caption{\textbf{Common reasons for answer filtering.} The top section covers single-ground-truth datasets (frequency estimates from a manual sample); the bottom section covers multi-annotator datasets.}\label{tab:filter-reasons}\\
\endlastfoot
\rowcolor{verosectionhl}
\multicolumn{3}{@{}l}{\textit{Single-ground-truth datasets}} \\
\midrule
Multi-value answers {\scriptsize(${\sim}300$)} & Ground truth contains multiple distinct values that cannot be reduced to a single verifiable target. & \texttt{AC\,=\,4, BD\,=\,4} \\
\midrule
Ambiguous text labels {\scriptsize(${\sim}300$)} & Descriptive phrases requiring fuzzy or semantic matching beyond exact string comparison. & ``Isosceles triangle'' \\
\midrule
Unsupported notation {\scriptsize(${\sim}300$)} & Scientific notation or symbolic algebraic expressions outside our numeric parser's scope. & $b = a\cos C$ \\
\midrule
Empty / invalid GT & Ground truth is missing, empty, or malformed, making reward computation impossible. & ``Empty case'' \\
\midrule
Unit mismatch & Unit is inconsistent with the question context or cannot be cleanly stripped to a numeric value. & Mass answered in cm \\
\midrule
Out-of-range values & Numeric ground truth falls outside the valid range for the quantity asked about. & $r = 1.215$ (correlation) \\
\midrule
Vector / complex answers & Multi-component quantities that cannot be reduced to a single scalar. & $(3, {-}2, 5)$;\; $2{+}3i$ \\
\midrule
Non-standard units & Numeric value mixed with a unit descriptor our parser does not handle. & ``1.70 million'' \\
\midrule
Non-task questions & Prompt requests instruction or explanation rather than a verifiable answer. & ``Explain how to solve\ldots'' \\
\midrule
\rowcolor{verosectionhl}
\multicolumn{3}{@{}l}{\textit{Multi-annotator datasets} (e.g., VQAv2, VizWiz, A-OKVQA)} \\
\midrule
Inconsistent answers (closed) & Annotators gave mutually exclusive responses with no dominant consensus for a single-answer question. & annotator 1: ``white,blue'', annotator 2: ``red'' \\
\midrule
Inconsistent answers (open) & Responses span unrelated concepts with no dominant semantic cluster. & ``fish'', ``float'', ``tow'' \\
\midrule
Answer--question type mismatch & Ground-truth type does not match what the question semantically requires. & Question asks about a person $\to$ Ground truth = ``Italy'' \\
\midrule
Unanswerable markers & Annotators flagged the question as unanswerable; remaining answers are insufficient. & ``unanswerable'' tag \\
\midrule
Open-ended descriptions & Free-form description question with no single canonical answer for exact matching. & ``Please describe this photo'' \\
\midrule
Composite questions & Multiple sub-questions whose interleaved answers cannot be parsed into a single target. & ``What is this? What color?'' \\
\midrule
Positional descriptors & Ground truth is a spatial reference rather than an identifying entity. & ``right'', ``in the back'' \\
\end{longtable}
}

\paragraph{Numeric.}
Numeric ground truths are stripped of surrounding units, currency symbols, degree markers, and LaTeX formatting to yield a plain decimal value. For example, ``\$327{,}000'' becomes \texttt{327000}, ``$60^\circ$'' becomes \texttt{60}, and ``8~V'' becomes \texttt{8}. Thousand separators are removed, fractions are converted to decimals (e.g., ``8/3'' $\to$ \texttt{2.6667}), and currency prefixes are dropped (e.g., ``\$222.14'' $\to$ \texttt{222.14}).

\paragraph{String.}
Free-form text answers undergo lowercasing and whitespace normalization to enable case-insensitive exact matching at reward time (e.g., ``Coronal'' $\to$ \texttt{coronal}).

\paragraph{None (unresolvable).}
Answers that the classifier cannot confidently assign to any of the above types---such as multi-part answers (e.g., ``(1) 3, (2) 120''), coordinate tuples (e.g., ``$(5.2, 0)$''), or single ambiguous tokens---are assigned type \emph{None} and filtered from the training set.

\paragraph{Common reasons for answer filtering.}
Answers that cannot be reliably verified by our programmatic reward functions are removed during answer filtering. Table~\ref{tab:filter-reasons} summarizes the primary filtering reasons. The top section lists reasons common to single-ground-truth datasets, in decreasing order of frequency. For datasets with multi-annotator ground truths (e.g., VQAv2, VizWiz, A-OKVQA), additional reasons arise from annotator disagreement and question--answer alignment issues, shown in the bottom section.

\subsection{Data Mixture}
\label{app:data-mixtures}

In Section 3.2 of the main paper, we examine how the task-category sampling distribution affects RL training. Here we describe the procedure used to construct each weighting scheme reported for that experiment.

\paragraph{Per-domain statistics.} We collect three statistics for each domain~$d$. Two of these, accuracy and reasoning length, require a profiling run: we train the base model (Qwen2.5-VL-7B-Instruct) for one epoch on a 100K-sample subset using uniform category weights and measure:
\begin{itemize}[leftmargin=1.5em,itemsep=2pt]
    \item \textbf{Accuracy} ($\text{acc}_d$): average reward on the held-out verification set for domain~$d$.
    \item \textbf{Reasoning length} ($L_d$): mean number of tokens on the held-out verification set inside the \texttt{<think>} block.
    \item \textbf{Image area} ($A_d$): mean pixel area of the input images (before any resizing), computed directly from the original training set.
\end{itemize}

\paragraph{Weighting schemes.} We run the data mixture experiment on five domains: Chart \& OCR, Grounding, Counting \& Search, Knowledge \& Recognition, Spatial \& Action, and STEM. Each non-uniform scheme defines a per-domain ratio $r_d$ proportional to a power-law function of one of the profiling statistics. The exponent $\alpha$ controls how aggressively the distribution deviates from uniform. We tune~$\alpha$ so that the ratio between the most- and least-weighted domains equals~1.6, a moderate spread that allows meaningful reallocation without starving any single category:
\begin{equation}
\frac{\max_d\; r_d}{\min_d\; r_d} = 1.6.wwwwwwwwwww
\end{equation}
Concretely, the four weighting schemes and the ablation without Knowledge \& Recognition are:

\begin{enumerate}[leftmargin=1.5em,itemsep=2pt]
    \item \textbf{Equal ratios} (uniform): $r_d = 0.20$ for all five domains.

    \item \textbf{Difficulty-weighted} ($r_d \propto (1 - \text{acc}_d)^{\alpha}$, $\alpha = 0.475$): Up-weights domains where the model performs poorly after the profiling run. Spatial \& Action receives the largest share (0.273) due to its low initial accuracy, while STEM receives the smallest (0.170).

    \item \textbf{Reasoning-length-weighted} ($r_d \propto L_d^{\alpha}$, $\alpha = 0.144$): Up-weights domains whose responses require longer chains of thought.  Chart \& OCR and Spatial \& Action receive the largest shares (${\sim}0.23$ each), while Knowledge \& Recognition receives the smallest (0.148). We also evaluate the inverse scheme ($r_d \propto L_d^{-\alpha}$), which favors domains with shorter reasoning traces.

    \item \textbf{Image-area-weighted} ($r_d \propto A_d^{\alpha}$, $\alpha = 0.443$): Up-weights domains with larger input images.  Grounding, Counting \& Search receives the largest share (0.244), while Spatial \& Action receives the smallest (0.153).

    \item \textbf{Without Knowledge \& Recognition}: Sets $r_d = 0$ for Knowledge \& Recognition and distributes weight equally among the remaining four domains ($r_d = 0.25$ each). This ablation tests whether the lowest-gain category can be dropped without harming overall performance.
\end{enumerate}

\section{Training Details}
\label{app:training-details}

\subsection{System Prompt}
\label{app:system-prompt}

We provide the system prompt for \model{} during training and evaluation in Listing~\ref{lst:system-prompt}.

\begin{tcolorbox}[
  promptstylefmt,
  title={System Prompt for Vero},
  label={lst:system-prompt}
]

You are a helpful, conversational assistant tasked with answering a question about an image.

Your response must include two parts:

\begin{promptenumerate}
\item \pbold{Reasoning}: A detailed, free-flowing chain of thought enclosed in \pcode{<think>} and \pcode{</think>} tags.
\item \pbold{Final Answer}: A clear, conversational response enclosed in \pcode{<answer>} and \pcode{</answer>} tags, using \pboxed{} notation when the question has a definitive answer.
\end{promptenumerate}

\promptrule

\pbold{Reasoning Instructions}

\begin{promptitemize}
\item The reasoning section must be inside \pcode{<think>} \ldots\ \pcode{</think>} tags.
\item The reasoning should resemble a stream of consciousness: explore, test hypotheses, backtrack if necessary, reflect, and refine.
\item Let the reasoning flow naturally while progressing toward a conclusion.
\item Use reasoning strategies such as:
  \begin{promptitemize}
  \item \pbold{Planning}: outline possible approaches before committing.
  \item \pbold{Exploration}: consider multiple image regions or interpretations, even unlikely ones.
  \item \pbold{Evaluation}: compare alternatives and verify against visual evidence.
  \item \pbold{Reflection}: revisit earlier ideas if they may still be viable.
  \end{promptitemize}
\item Thoroughly examine and cross-check relevant image regions before narrowing down.
\item If the image is ambiguous, make a reasonable inference based on visual and contextual cues.
\item End the reasoning once you are confident in the conclusion.
\end{promptitemize}

\promptrule

\pbold{Final Answer Instructions}

\begin{promptitemize}
\item The answer section must be enclosed in \pcode{<answer>} \ldots\ \pcode{</answer>} tags.
\item The \pcode{<answer>} section should stand on its own as a response to the user: it must provide necessary context and justification so that a reader can understand and verify the conclusion without reading \pcode{<think>}.
  \begin{promptitemize}
  \item Do NOT refer to the \pcode{<think>} section (avoid phrases like ``as explained above'' or ``from the reasoning'').
  \end{promptitemize}
\item Boxed result:
  \begin{promptitemize}
  \item If the question has a definitive, concise answer (a number, word, phrase, or label), include a conversational, natural response followed by exactly one boxed result using LaTeX: \pboxed{final\_result}.
  \item If the question is open-ended, subjective, or does not yield a concise final result, omit the boxed notation.
  \end{promptitemize}
\end{promptitemize}

\promptrule

\pbold{Format Example}

\begin{lstlisting}[style=promptcode]
<think>
Detailed reasoning goes here...
</think>
<answer>
Self-contained response goes here...
Following the response, if a concise final result exists, include: \boxed{final_result}.
If open-ended or no concise result, respond naturally without \boxed.
</answer>
\end{lstlisting}

\end{tcolorbox}

\subsection{Reinforcement Learning}
\label{app:rl-details}

\paragraph{GSPO algorithm and objective.}
\label{app:gspo}
The GSPO objective is a clipped surrogate loss aggregated as the mean-of-sequence-means (\emph{seq-mean-token-mean}). Given a group of $G$ rollouts $\{y_i\}_{i=1}^G$ for a prompt $(v, q)$, define the per-response sequence-average log-probability difference:
\begin{equation}
\bar{\Delta}_i = \frac{1}{|y_i|}\sum_{t=1}^{|y_i|} \bigl(\log \pi_\theta(y_{i,t} \mid v, q, y_{i,<t}) - \log \pi_{\theta_\text{old}}(y_{i,t} \mid v, q, y_{i,<t})\bigr).
\end{equation}
The sequence-level importance ratio at token $t$ is then formed by routing the gradient through the sequence average while keeping the token-level log-prob differentiable:
\begin{equation}
s_{i,t}(\theta) = \exp\!\Bigl(\operatorname{sg}(\bar{\Delta}_i) + \log \pi_\theta(y_{i,t}) - \operatorname{sg}\!\bigl(\log \pi_\theta(y_{i,t})\bigr)\Bigr),
\end{equation}
where $\operatorname{sg}$ denotes stop-gradient. The GSPO objective is:
\begin{equation}
\label{eq:gspo-full}
\mathcal{J}(\theta) =
\frac{1}{G}\sum_{i=1}^{G} \frac{1}{|y_i|} \sum_{t=1}^{|y_i|}
\min\!\Bigl(
  s_{i,t}(\theta)\,A_i,\;
  \operatorname{clip}\!\bigl(s_{i,t}(\theta),\,1{-}\varepsilon_\text{low},\,1{+}\varepsilon_\text{high}\bigr)\,A_i
\Bigr),
\end{equation}
where the normalized group advantage is:
\begin{equation}
A_i = \frac{r_i - \mu_g}{\sigma_g + \epsilon},
\qquad
\mu_g = \frac{1}{G}\sum_{j=1}^{G} r_j,
\qquad
\sigma_g = \operatorname{std}\!\bigl(\{r_j\}_{j=1}^G\bigr).
\end{equation}

\paragraph{Training hyperparameters.}
\label{app:training-hyperparams}
We detail the RL training hyperparameters in Table~\ref{tab:rl-hyperparams} and per-model configurations in Table~\ref{tab:base-model-train-configs}. All models are trained for 2{,}343 steps using VeRL~\citep{sheng2025hybridflow} with FSDP2 on 8 GPUs. In preliminary experiments, we found that Qwen models exhibit slightly more stable training under fp16, following~\citet{qi2025defeating}, while MiMo-VL trains stably in bf16.

\begin{table}[h]
\centering
\small
\begin{tabular}{ll}
\toprule
\textbf{Hyperparameter} & \textbf{Value} \\
\midrule
Framework & VeRL \\
FSDP strategy & fsdp2 \\
Rollouts per prompt ($G$) & 8 \\
Train batch size & 256 \\
PPO mini-batch size & 128 \\
Learning rate & $1 \times 10^{-6}$ \\
LR warmup steps & 40 \\
Clip lower ($\varepsilon_\text{low}$) & 0.0003 \\
Clip upper ($\varepsilon_\text{high}$) & 0.0004 \\
KL coefficient & 0 \\
Rollout temperature & 1.0 \\
\bottomrule
\end{tabular}
\caption{RL training hyperparameters for all \model{} models. Training uses VeRL with GSPO~\citep{gspo}, asymmetric clipping ($\varepsilon_\text{low} < \varepsilon_\text{high}$), and no KL penalty to allow less-restricted policy updates.}
\label{tab:rl-hyperparams}
\end{table}

\begin{table}[h]
\centering
\small
\renewcommand{\arraystretch}{1.15}
\begin{tabular}{@{}lccrrcl@{}}
\toprule
\textbf{Base model} & \textbf{GPUs} & \textbf{Steps} & \textbf{Ctx.} & \textbf{Max px.} & \textbf{Dtype} & \textbf{Coords} \\
\midrule
MiMo-VL-7B-SFT      & 8$\times$H100 & 2{,}000 & 28{,}672 & $3072^2$ & bf16 & absolute \\
Qwen2.5-VL-7B-Inst.  & 8$\times$H100 & 2{,}000 & 24{,}576 & $3072^2$ & fp16 & absolute \\
Qwen3-VL-8B-Inst.    & 8$\times$H200 & 2{,}000 & 36{,}864 & $4096^2$ & fp16 & norm.\,0--1k \\
Qwen3-VL-8B-Think.   & 8$\times$H200 & 2{,}000 & 36{,}864 & $4096^2$ & fp16 & norm.\,0--1k \\
\bottomrule
\end{tabular}
\caption{Base-model training configurations for the \model{} variants, including context length, maximum image resolution, precision, and coordinate format. Qwen-family models use fp16 for improved training stability~\citep{qi2025defeating}.}\label{tab:base-model-train-configs}
\end{table}
\newpage

\subsection{Reward}
\label{app:reward-details}
The total reward for a response $y$ is:
\begin{equation}
R(y, y^*) = (1 - \alpha)\,R_\text{acc}(y, y^*) + \alpha\,R_\text{fmt}(y)
+ R_\text{overlong}(y),
\end{equation}
with $\alpha = 0.2$.

For tasks combining programmatic instruction-following with open-ended judgment, the blended accuracy score is:
\begin{equation}
\tilde{R}_\text{acc}(y, y^*) = w\,R_\text{inst}(y) + (1-w)\,R_\text{judge}(y), \quad w = 0.5.
\end{equation}
The LLM judge produces a score on a 1--10 scale, normalized to $[0,1]$ as $(s-1)/9$.

\paragraph{Overlong penalty.}
To discourage excessively long responses, we use the soft penalty from \citet{dapo} as a linear ramp in the buffer zone $[L_\text{max} - B,\; L_\text{max}]$:
\begin{equation}
R_\text{overlong}(y) = \min\!\left(-\frac{|y| - (L_\text{max} - B)}{B}\,\lambda,\; 0\right),
\end{equation}
where $B = 2048$, $L_\text{max} = \texttt{max\_tokens}$, and $\lambda = 1.0$.

\paragraph{Format reward.}
$R_\text{fmt}$ requires the response to follow the format \texttt{<think>}$\ldots$\texttt{</think>}\allowbreak\texttt{<answer>}$\ldots$\texttt{</answer>} with non-empty think content; responses that violate this structure receive $R_\text{fmt} = 0$. Given valid structure, $R_\text{fmt} = 1$ by default. For discrete symbolic answer types (string match, multiple choice, numeric, list match, counting, ordering, search, web action), a single valid \verb|\boxed{...}| in the answer block is additionally required for $R_\text{fmt} = 1$; its absence or the presence of multiple \verb|\boxed{...}| expressions reduces $R_\text{fmt}$ to $0.5$. For grounding and clicking, the presence of multiple \verb|\boxed| expressions similarly reduces $R_\text{fmt}$ to $0.5$.

\paragraph{Training judge.}
\label{app:judge-prompt}
We include our training judge prompt in Listing~\ref{lst:judge-prompt}. We adapt the LLM judge prompt from OLMo3~\citep{olmo2025olmo}. For training-time LLM-as-judge rewards, we use Qwen3-32B served via vLLM with thinking disabled. We set judge temperature to $0.7$ and judge max tokens to $1{,}024$.

\clearpage
\begin{tcolorbox}[
  promptstylefmt,
  title={Prompt for LLM Judge Reward},
  label={lst:judge-prompt}
]

Please act as an impartial judge and evaluate the quality of the answer provided by an AI assistant to the conversation history leading up to the answer displayed below. Judge whether the provided answer is good by comparing it to the reference answer.

Notes:

\begin{promptitemize}
\item Besides comparing to the reference answer, your evaluation should consider factors such as the naturalness, coherence, helpfulness, relevance, accuracy, creativity, appropriate level of detail, and how well the response satisfies the user's explicit constraints or accurately follows their instructions.
\item The AI answer may use \pcode{\textbackslash boxed\{\}} exactly once for a definitive concise answer (number, word, phrase, or label), or not at all if the question is open-ended or subjective. Penalize if it is used on intermediate results or explanations.
\item Note that sometimes the reference answer is not the only answer. So any valid variation of the reference answer is also acceptable.
\item The conversation involves an image that is not shown to you. Use the reference answer as ground truth for any visual content.
\item \pbold{Automatic Failure Conditions (Score = 1):} The following violations require an automatic score of 1 under any circumstance, regardless of the overall quality of the answer. No partial credit may be given if any of these appear.
  \begin{promptenumerate}
  \item \pbold{Notes to the judge or self-talk:} Any meta commentary, internal reasoning, notes that are directed towards the judge, or reflective statements about how the answer was constructed automatically results in a score of 1.
    \begin{promptitemize}
    \item [Examples omitted for brevity]
    \end{promptitemize}
  \item \pbold{Self-evaluative or compliance-asserting statements:} Any claim about the answer's correctness, completeness, quality, adherence to constraints, or deservingness of a high score automatically results in a score of 1. Do not consider such claims as mitigating factors.
    \begin{promptitemize}
    \item [Examples omitted for brevity]
    \end{promptitemize}
  \end{promptenumerate}
  Judges must explicitly check for these violations. If any instance is present, the score must be 1.
\item \pbold{Unnatural Penalty Condition (Score Reduction Required):} The score must be reduced if the answer includes gratuitous verbosity, repetition, rhetorical padding, inflated phrasing, or stylistically unnatural language that does not add informational value. Explanations, intermediate reasoning steps, and brief summaries are permitted when they directly support the answer and are proportionate to the complexity of the question.
\item For context, provided below is the Conversation History, AI Answer, and Reference Gold Answer.
\end{promptitemize}

\begin{lstlisting}[style=promptcode]
[Conversation History START]
{input}
[Conversation History END]

[AI Answer START]
{output}
[AI Answer END]

[Reference Gold Answer START]
{label}
[Reference Gold Answer END]
\end{lstlisting}

Please adhere to the following format.

\begin{promptitemize}
\item Respond in JSON format.
\item Begin your evaluation by providing a short explanation in the \pcode{"REASONING"} key.
\item Be as objective as possible. After providing your short explanation, please output a score on a scale of 1 to 10 in the \pcode{"SCORE"} key.
\end{promptitemize}

[Your judgement]\\
Respond in JSON format: \pcode{\{"REASONING": "[...]", "SCORE": "<your-score>"\}}

\end{tcolorbox}
\newpage
\paragraph{Math\_verify reward.}
In Section~\ref{sec:ablations}, we ablate our reward design with a \texttt{math\_verify} baseline (Table~\ref{tab:ablations}(b)). The baseline replaces our full reward router with a single unified verifier built on the open-source \textsc{math-verify} library~\citep{mathverify}. Our task-routed reward outperforms \texttt{math\_verify} on every category, improving the overall average from 51.8 to 57.2 (+5.4), with the largest gain on Captioning \& Instruction Following (70.6 vs.\ 34.3). This highlights the need for task-specific reward routing in multi-task RL.

The \texttt{math\_verify} verifier performs reward computation as follows: \textbf{1. Case-insensitive string match.} We first try naive string matching. If the lowercased, whitespace-stripped prediction equals the lowercased ground truth, the reward is~1. \textbf{2. Symbolic parsing and verification.} Both the ground truth and the prediction are passed to \textsc{math-verify}'s \texttt{parse} function, which includes robust parsing of numerical, symbolic, and multiple choice answers embedded in text. The parsed representations are then compared via \textsc{math-verify}'s \texttt{verify} function to robustly match the prediction with the ground truth after normalization. If \texttt{verify} returns \texttt{True}, the reward is~1.
Otherwise, the reward is~0.

\subsection{Supervised Fine-tuning}
\label{app:sft-details}

We use supervised fine-tuning as a baseline in our SFT vs.\ RL ablation (Table~\ref{tab:ablations}(a), Section~\ref{sec:ablations}). Table~\ref{tab:supervised-finetuning-hyperparams} summarizes the shared SFT hyperparameters. For each baseline, we sweep learning rates over $\{1\mathrm{e}{-6}, 1\mathrm{e}{-7}, 5\mathrm{e}{-7}, 5\mathrm{e}{-6}\}$. SFT on our data outperforms FineVision SFT (52.8 vs.\ 46.2), but RL yields substantially larger gains across all categories (57.2 overall, +4.8 over the base model).

\section{Evaluation Details}
\label{app:eval-details}

We use two decoding setups depending on the model trained. Qwen2.5-VL and
MiMo-VL trained models follow the Qwen2.5-VL~\citep{qwen25vl} recommended decoding setup. Qwen3-VL trained models follow the decoding setup
reported in the Qwen3-VL report~\citep{qwen3vl}. In both cases, evaluation uses one sampled
decode per example. Tables~\ref{tab:eval-inference-settings-qwen25} and
\ref{tab:eval-inference-settings-qwen3} summarize the model-family-specific
sampling parameters and the shared runtime settings. We use Qwen3-32B with thinking disabled as the evaluation LLM judge when an LLM judge is required. For benchmarks requiring a VLM judge, we use Qwen3-VL-32B-Instruct. For judges, we use sampling parameters set to Temperature=0.7, TopP=0.8, TopK=20, and MinP=0.
In the main results table, \model{} variants are reported as Avg@5 with SEM across runs. Baseline scores are single reported or evaluated values unless otherwise indicated.

\paragraph{Benchmark-specific choices.} We use the Standard and Vision splits of MMMU-Pro and MathVista$_{testmini}$, the English subset of SimpleVQA, and GameQA$_{Lite}$, where $Lite$ denotes sampling 2,633 examples. For AerialVG, we report mean IoU. For MM-MTBench, we rescale to a 0 to 100 scale using $\text{score}_{100} = 100 \cdot (x - 1) / 9$. For VisualProbe, we report the average of easy, medium, and hard subsets.

\begin{table*}[t]
\centering
\begin{minipage}[t]{0.31\textwidth}
\centering
\small
\setlength{\tabcolsep}{3pt}
\begin{minipage}[c][0.19\textheight][c]{\linewidth}
\centering
\begin{tabular}{@{}ll@{}}
\toprule
\textbf{Hyperparameter} & \textbf{Value} \\
\midrule
Epochs & 1 \\
Weight decay & 0.01 \\
Warmup ratio & 0.03 \\
Batch size & 128 \\
Scheduler & Cosine \\
Precision & bf16 \\
Flash attention & fa2 \\
Max sequence length & 32768 \\
\bottomrule
\end{tabular}
\end{minipage}
\captionof{table}{SFT hyperparameters.}
\label{tab:supervised-finetuning-hyperparams}
\end{minipage}
\hfill
\begin{minipage}[t]{0.31\textwidth}
\centering
\small
\setlength{\tabcolsep}{3pt}
\begin{minipage}[c][0.19\textheight][c]{\linewidth}
\centering
\begin{tabular}{@{}ll@{}}
\toprule
\textbf{Setting} & \textbf{Value} \\
\midrule
Max new tokens & 16,384 \\
Temperature & 0.6 \\
Top-$p$ & 1.0 \\
Max image pixels & $4096 \times 4096$ \\
\bottomrule
\end{tabular}
\end{minipage}
\captionof{table}{Evaluation settings for Qwen2.5-VL and MiMo-VL.}
\label{tab:eval-inference-settings-qwen25}
\end{minipage}
\hfill
\begin{minipage}[t]{0.31\textwidth}
\centering
\small
\setlength{\tabcolsep}{3pt}
\begin{minipage}[c][0.19\textheight][c]{\linewidth}
\centering
\begin{tabular}{@{}ll@{}}
\toprule
\textbf{Setting} & \textbf{Value} \\
\midrule
Max new tokens & 16,384 \\
Temperature & 1.0 \\
Top-$p$ & 0.95 \\
Top-$k$ & 20 \\
Presence penalty & 1.5 \\
Max image pixels & $4096 \times 4096$ \\
\bottomrule
\end{tabular}
\end{minipage}
\captionof{table}{Evaluation settings for Qwen3-VL.}
\label{tab:eval-inference-settings-qwen3}
\end{minipage}
\end{table*}

\section{Additional Analyses}
\label{app:additional-analyses}

\subsection{Image Size}
\begin{figure}[!h]
    \centering
    \begin{minipage}[t]{0.48\textwidth}
        \centering
        \includegraphics[width=\linewidth]{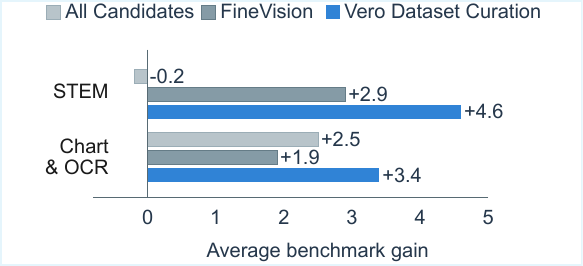}
        \caption{\textbf{Ablation on dataset filtering.} Each model is trained for 1 epoch with GSPO and a math verify~\citep{mathverify} reward on 100k examples from a single task category. We report the average benchmark gain over the base model (Qwen2.5-VL-7B-Instruct) for each category.}
        \vspace{-1em}
        \label{fig:dataset-filtering-ablation}
    \end{minipage}\hfill
    \begin{minipage}[t]{0.48\textwidth}
        \centering
        \includegraphics[width=\linewidth]{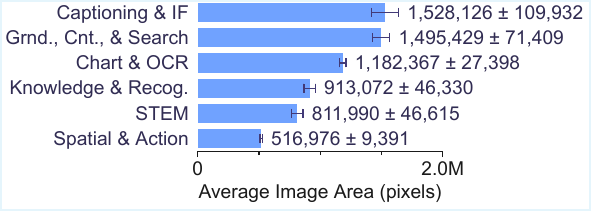}
        \caption{\textbf{Average image area (pixels) by task category.} The horizontal bars represent the mean image resolution for datasets within each specific domain. Error bars on each bar denote the standard error of the mean for that category.}
        \vspace{-1em}
        \label{fig:appendix-image-area-by-category}
    \end{minipage}
\end{figure}

Figure~\ref{fig:appendix-image-area-by-category} shows clear variation in average image area across task categories. Captioning \& Instruction Following and Grounding, Counting \& Search use the largest images at $1.53 \pm 0.11$ million and $1.50 \pm 0.07$ million pixels, followed by Chart \& OCR at $1.18 \pm 0.03$ million pixels. Knowledge \& Recognition and STEM fall in the middle at $0.91 \pm 0.05$ million and $0.81 \pm 0.05$ million pixels, while Spatial \& Action uses the smallest images at $0.52 \pm 0.01$ million pixels. 

\subsection{Behavioral Definitions}
\label{app:behavioral-analysis}
\paragraph{Prompt adaptation.}
Following \citet{kargupta2025cognitivefoundationsreasoningmanifestation}, we conduct an automated annotation of the reasoning traces using Qwen3-32B~\citep{qwen3vl} as a strong evaluator model. We extend their taxonomy to better capture multimodal reasoning strategies by augmenting the original 28 text-centric behaviors with six supplementary visual-analysis capabilities: \emph{arithmetic-calculation}, \emph{mental-imagery-simulation}, \emph{perception-then-reasoning}, \emph{systematic-regional-synthesis}, \emph{visual-foraging}, and \emph{visual-reference-or-grounding}. In total, the evaluator monitors for 34 distinct cognitive capabilities.

In accordance with the original protocol, we formulate an individualized prompt per capability. Each prompt provides a precise definition, explicit criteria for behavioral evidence, and few-shot examples for in-context learning. Guided by these instructions, the evaluator model assesses the reasoning trace and outputs its final analysis in a structured JSON format, consisting of a final score indicating presence or absence and a supporting explanation.

To improve efficiency and reliability, we adapted the protocol by making two key simplifications. First, we removed the requirement for the evaluator to perform exact span identification. The original protocol required the evaluator model to pinpoint the exact textual spans or sentences within the reasoning trace where the cognitive behavior manifested. We relaxed this constraint, instructing the model to provide a single, holistic judgment for the entire trace. Second, we eliminated the original 0--2 continuous grading scale (where 0 indicated absent, 1 indicated partially present, and 2 indicated fully present) in favor of a strictly binary system (0 for absent, 1 for present).

We generate annotations with greedy decoding ($T{=}0$, $\text{max\_tokens}{=}2{,}048$, $\text{seed}{=}42$). To mitigate benchmark-size imbalance within each validation domain, we downsample larger benchmarks to match the smallest benchmark in that domain. In addition to the Qwen3 results shown in Figure~\ref{fig:hl_presence_rate} of the main paper, we report the corresponding Qwen2.5 results in Figure~\ref{fig:hl_presence_rate_appendix}.

\begin{center}
    \includegraphics[width=1.0\linewidth]{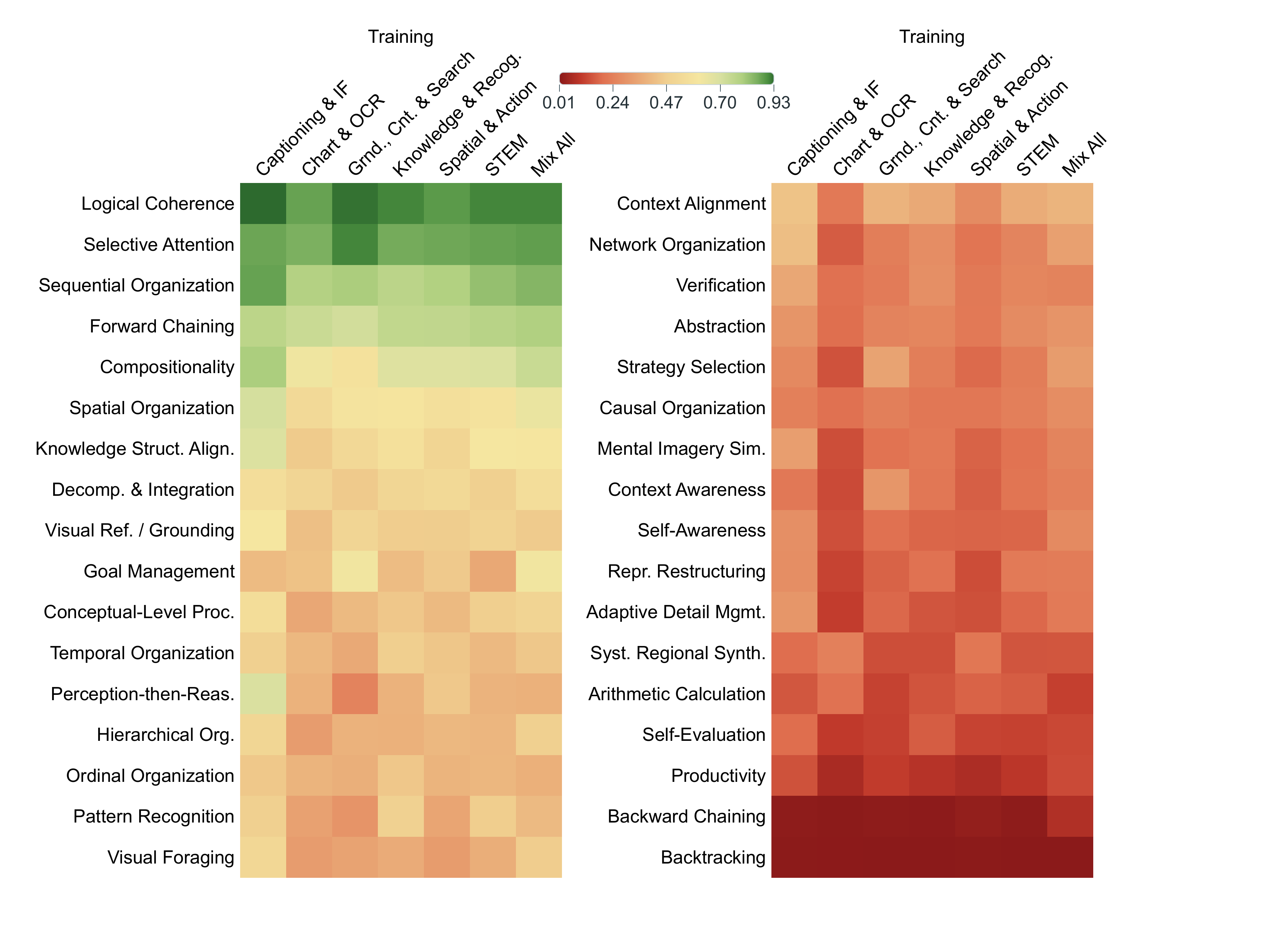}
    \captionof{figure}{\textbf{High-level cognitive behavior presence rates across task categories for single-task and mixed-task trained Qwen2.5 models.} Presence rate is indexed by color intensity, where green represents high cognitive presence and red represents low presence, mapping the relationship between training data and emergent reasoning capabilities. The plot reveals that a model's emergent cognitive behaviors are highly dependent on its specific training domain.}
    \label{fig:hl_presence_rate_appendix}
\end{center}

\subsection{Cognitive Behaviors}\label{sec:high_level_cognitive_foundations_presence_rate}
We include the cognitive behavioral presence rate of Qwen2.5-VL-7B-Instruct in Figure~\ref{fig:hl_presence_rate_appendix}. The results further confirm our findings in \secrefc{sec:high-level-cognitive} in the main paper, as the captioning-trained model consistently uses more mental imagery simulation (0.33 vs.\ 0.19 cross-domain average in Qwen2.5), and chart-trained models again demonstrate elevated systematic regional synthesis (0.24 vs.\ 0.16). Additionally, we found that compared to Qwen3 (Figure~\ref{fig:hl_presence_rate} in the main paper), several behaviors are much weaker in Qwen2.5: backtracking stays near zero (${\sim}$0.01 vs.\ 0.12--0.48 in Qwen3), self-evaluation ranges from 0.10--0.19 (vs.\ 0.46--0.94), and strategy selection from 0.15--0.34 (vs.\ 0.57--0.80). Both models share high logical coherence (${\sim}$0.91 vs.\ ${\sim}$0.98) and selective attention (${\sim}$0.87 vs.\ ${\sim}$0.98), but Qwen2.5 covers a narrower range overall, indicating that the stronger base model develops a wider set of reasoning behaviors through RL training.

\subsection{Behavioral Skill Extraction}\label{sec:behavioral_extraction_details}
Building on the concept of meta-cognitive reuse~\citep{didolkar2025metacognitivereuseturningrecurring}, we formalize a comprehensive pipeline to discover, consolidate, and quantify fundamental reasoning skills directly from raw model traces. Our objective is to transition from analyzing task-specific execution steps (e.g., "counting three red cars") to cataloging domain-agnostic cognitive strategies (e.g., "systematic spatial enumeration"). To achieve this, we design a three-step pipeline consisting of extraction, deduplication, and annotation.

\textbf{Extraction.} In Stage 1, the model ingests a single multimodal reasoning trace to identify reusable, generalizable strategies. We constrain the model's output so that every provisional name distinctly encodes the \emph{action}, \emph{target}, and \emph{goal} (e.g., \texttt{behavior_\allowbreak relative_\allowbreak camera_\allowbreak distance_\allowbreak comparison}), proactively prohibiting problem-specific entities or narrow qualifiers. In Stage 2, we prevent lexical explosion by autoregressively maintaining a global behavior codebook. Qwen3 evaluates each new candidate against the existing codebook, mapping it into one of four rigid relationships: an \emph{exact equivalent} (mapped to an existing skill), a \emph{subtype} (discarded), a \emph{more general replacement} (overwrites a narrower entry), or a \emph{distinct new skill} (appended). This trace-by-trace reconciliation acts as a real-time semantic filter, continuously compressing the codebook and reducing redundant phrasing.

\textbf{Deduplication.} Because traces are processed independently, semantically identical behaviors often emerge under varying names. To finalize the codebook, we embed the concatenated ``name: description'' of every extracted behavior using OpenAI's \texttt{text-\allowbreak embedding-\allowbreak 3-\allowbreak small} and group them via agglomerative hierarchical clustering. To filter out non-reusable skills, we discard any behavior cluster appearing fewer than 10 times. For surviving clusters, GPT-4o synthesizes a single \emph{canonical name} and a comprehensive description. Finally, human annotators manually inspect these definitions against a sample of source traces to verify that each cluster's semantic boundary is sensible and accurately reflects the underlying cognitive actions.

\textbf{Annotation.} To quantify capability prevalence while preventing domain imbalance, we uniformly subsample benchmarks to match the smallest dataset within each domain before re-annotating. We frame this behavior-mapping as a multi-label classification task. For each instance, Qwen3-32B evaluates the reasoning trace holistically against the canonical dictionary, returning a structured JSON object. For every behavior, the model provides a brief justification detailing its manifestation followed by a strictly binary presence score (1 for present, 0 for absent). We explicitly chose binary scoring over a continuous scale to minimize calibration variance and eliminate the subjectivity of scoring "partial" manifestations, yielding a robust, dense matrix of capability profiles across all models.

\subsection{Skill Experiments}\label{sec:add_skill_analysis_results} 

\paragraph{Skill behavior presence rate.}

\begin{figure}[p]
    \centering
    \vspace{-30pt}
    \includegraphics[width=0.8\textwidth]{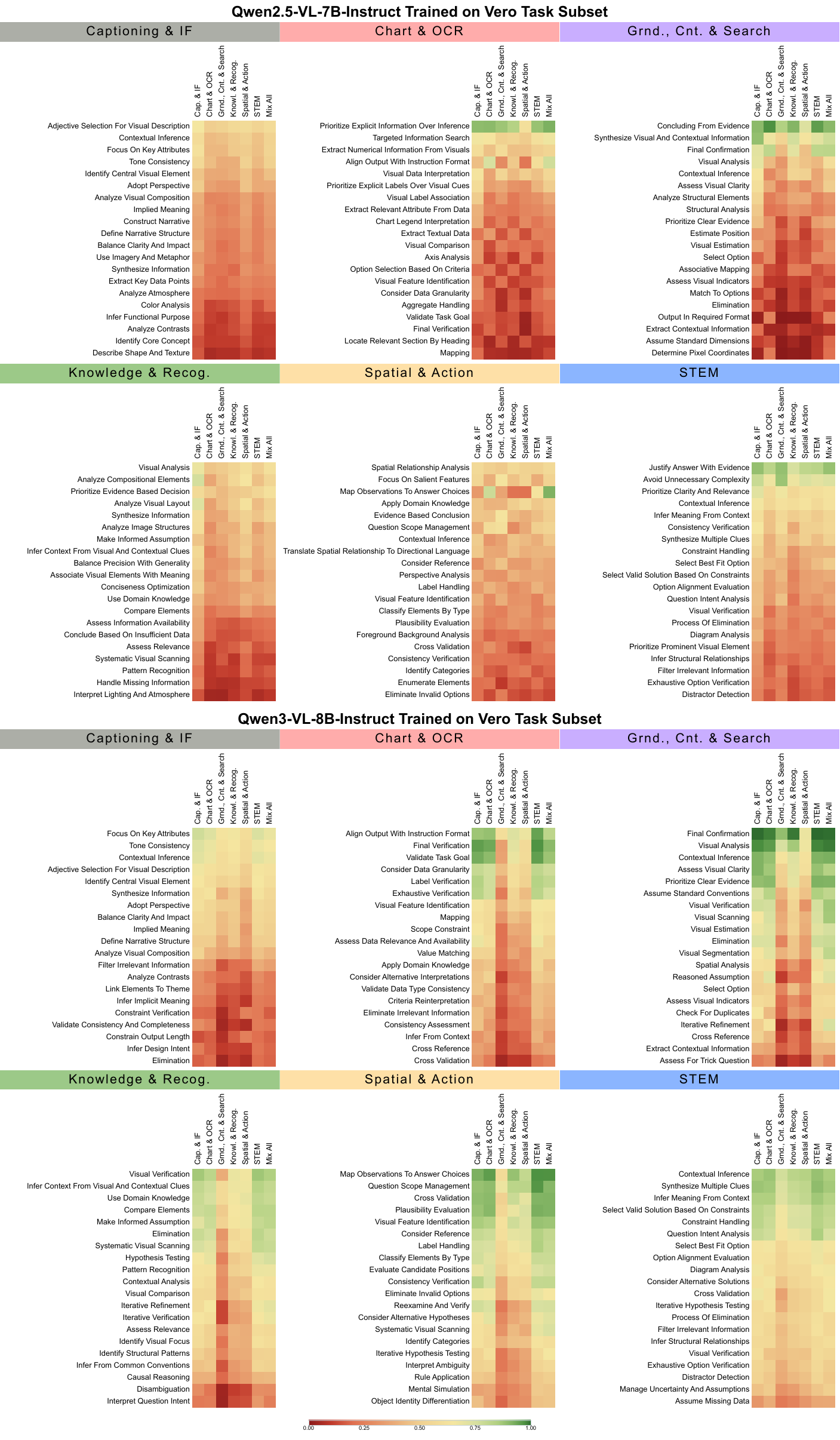}
    \caption{\textbf{Behavioral skill analysis presence rates for models trained on individual task categories and one trained on all categories (mix all).} Training on individual task categories impacts the emergence of fine-grained skills within each domain, with green indicating higher skill presence and red indicating lower. For instance, chart skills center on reading operations like "Axis Analysis", while spatial skills center on physical state reasoning like "Mental Simulation".}
    \label{fig:low_level_presence_rate_appendix}
\end{figure}

Figure~\ref{fig:low_level_presence_rate_appendix} shows the presence rate per model and task category, highlighting the 20 skills with the largest variance across training conditions in each category. As shown, different training task categories can produce substantial differences in skill presence rates, even when models are evaluated on the same task category. This variance indicates that the underlying cognitive strategies a model employs are deeply coupled with its specific training distribution. Rather than developing a universal reasoning pathway, models adapt their problem-solving approaches to the distinct domains they were trained on. 

Furthermore, this reveals why multi-task RL training is notoriously difficult: each task demands a fundamentally distinct cognitive profile. For example, optimizing a model to excel at captioning requires reinforcing generative skills like "Define Narrative Structure", which differ drastically from the strategies required for chart and OCR tasks, such as "Axis Analysis". Consequently, we conclude that the cross-domain transfer of low-level reasoning skills is not guaranteed. A highly diverse, mixed-domain training curriculum is strictly necessary to equip a multimodal model with a balanced repertoire of cognitive skills and prevent reasoning blind spots during generalization.

\paragraph{Logistic regression probe.}
As shown in the main paper, skills extracted per task category are largely linearly separable, which further supports the claim that distinct task domains necessitate fundamentally different cognitive profiles. Here we provide the details of constructing the probe. We select the annotated skills from the within-task reasoning traces, where the model is trained and evaluated on the same category domain, and embed the canonical behaviors using Qwen3-Embedding-8B. We subsample the reasoning traces to 800 per domain to maintain a balanced distribution. We then train a pipeline consisting of (i)~per-fold mean centering via \texttt{StandardScaler(with\_std=False)}, (ii)~$\ell_2$ normalization, and (iii)~multinomial logistic regression with a maximum of 2,000 iterations. The pipeline is evaluated using 5-fold Stratified Group K-Fold cross-validation.

\subsection*{Full Catalog of Behavior-Description Pairs}
Table~\ref{tab:behaviors} summarizes the high-level cognitive behavior codebook used in the behavioral analyses. Table~\ref{tab:skills_def} then lists the finer-grained skill definitions that support the behavioral skill analysis and probe construction. 

\begin{center}
    \centering
    \begingroup
    \scriptsize
    \setlength{\tabcolsep}{2pt}
    \renewcommand{\arraystretch}{0.9}
    \begin{minipage}[t]{0.49\linewidth}
    \vspace{0pt}
    \begin{tabularx}{\linewidth}{@{}>{\raggedright\arraybackslash}p{0.31\linewidth}>{\raggedright\arraybackslash}X@{}}
    \toprule
    Behavior & Definition \\
    \midrule
    Abstraction & Extract general principles from specific instances. \\
    Adaptive Detail Management & Adjust the level of detail based on reasoning requirements. \\
    Arithmetic Calculation & Extract, manipulate, and compute numerical values to reach a verifiable solution within a reasoning trace. \\
    Backtracking & Identify unproductive paths and return to earlier decision points. \\
    Backward Chaining & Start with goals and work backward to identify prerequisites. \\
    Causal Organization & Arrange elements through cause-effect relationships. \\
    Compositionality & Build complex ideas from simpler components. \\
    Conceptual Level Processing & Reason with abstract concepts before translating to linguistic forms. \\
    Context Alignment & Select appropriate organizational patterns based on context. \\
    Context Awareness & Recognize how the situational context shapes which reasoning strategies and goals are appropriate. \\
    Decomposition And Integration & Break problems into subparts and synthesize solutions. \\
    Forward Chaining & Start with initial conditions and work toward goals. \\
    Goal Management & Establish, maintain, and adjust goals throughout the reasoning process. \\
    Hierarchical Organization & Arrange concepts in nested, tree-like structures with parent-child relationships. \\
    Knowledge Structure Alignment & Match reasoning organization to domain knowledge structure. \\
    Logical Coherence & Maintain consistency in reasoning across steps and contexts. \\
    Mental Imagery Simulation & Generate internal representations that preserve the properties of a stimulus in the absence of direct sensory input. \\
    \bottomrule
    \end{tabularx}
    \end{minipage}
    \hfill
    \begin{minipage}[t]{0.49\linewidth}
    \vspace{0pt}
    \begin{tabularx}{\linewidth}{@{}>{\raggedright\arraybackslash}p{0.31\linewidth}>{\raggedright\arraybackslash}X@{}}
    \toprule
    Behavior & Definition \\
    \midrule
    Network Organization & Arrange concepts as interconnected nodes with multiple pathways and relationship types. \\
    Ordinal Organization & Arrange elements according to relative rank or position. \\
    Pattern Recognition & Recognize recurring structures across different contexts. \\
    Perception Then Reasoning & Separate cognitive labor into two distinct stages: exhaustive information extraction and abstract logical operations. \\
    Productivity & Generate novel combinations using a finite set of elements. \\
    Representational Restructuring & Reformulate problems to reveal new insights. \\
    Selective Attention & Focus on relevant information while filtering out distractions. \\
    Self Awareness & Assess one's own knowledge state, capabilities, and task solvability. \\
    Self Evaluation & Assess the quality, correctness, efficiency, and progress of one's reasoning and make adjustments as needed. \\
    Sequential Organization & Arrange steps in linear order where sequence matters. \\
    Spatial Organization & Arrange elements according to spatial relationships and configurations. \\
    Strategy Selection & Choose the most appropriate reasoning approaches based on task requirements and domain. \\
    Systematic Regional Synthesis & Iteratively traverse multiple elements or regions of an image in a deliberate sequence, and synthesize information across them. \\
    Temporal Organization & Arrange elements along a timeline with before/after relationships. \\
    Verification & Check reasoning steps against established criteria. \\
    Visual Foraging & Strategically manage the acquisition of multiple targets or information points within a complex environment. \\
    Visual Reference Or Grounding & Establish a persistent and precise link between linguistic symbols and localized visual elements. \\
    \bottomrule
    \end{tabularx}
    \end{minipage}
    \endgroup
     \captionof{table}{\textbf{Summary of high-level behavior definitions.} A glossary detailing the high-level cognitive behaviors.}
    \label{tab:behaviors}
    \end{center}

\begin{table}[p]
\centering
\begingroup
\tiny
\setlength{\tabcolsep}{1.5pt}
\renewcommand{\arraystretch}{0.82}
\begin{minipage}[t]{0.49\linewidth}
\vspace{0pt}
\begin{tabularx}{\linewidth}{@{}>{\raggedright\arraybackslash}p{0.29\linewidth}>{\raggedright\arraybackslash}X@{}}
\toprule
Skill & Definition \\
\midrule
\multicolumn{2}{@{}l}{\textit{Captioning \& Instruction Following}} \\
Adjective Selection For Visual Description & Choosing descriptive adjectives that reflect style and function to capture a scene. \\
Adopt Perspective & Use vivid language to convey a specific perspective without using pronouns. \\
Analyze Atmosphere & Establish emotional connection to a setting via environmental interaction. \\
Analyze Contrasts & Use visual contrast (color, structure) to highlight key information. \\
Analyze Visual Composition & Assess visual elements like focal points to determine mood or message. \\
Balance Clarity And Impact & Blend concrete imagery with abstract language for vivid descriptions. \\
Constrain Output Length & Limit the response to a specific word count while remaining coherent. \\
Constraint Verification & Systematically verifying that all specified constraints are met. \\
Construct Narrative & Construct a surreal narrative using imaginative and metaphorical language. \\
Define Narrative Structure & Organize narrative by subject, then setting, then secondary subjects. \\
Describe Shape And Texture & Incorporate descriptions of shape and texture to convey visual traits. \\
Extract Key Data Points & Extract specific data points (city names, deals) from visual labels. \\
Focus On Key Attributes & Prioritize features impacting user experience like comfort and aesthetics. \\
Identify Central Visual Element & Choose representative visual elements to focus on in creative output. \\
Identify Core Concept & Extract core technical concepts or mechanisms from the text. \\
Implied Meaning & Convey the presence of an element indirectly by referencing related phenomena. \\
Infer Design Intent & Interpret technical styles like blueprints to infer purpose or context. \\
Infer Functional Purpose & Deduce system purpose by analyzing labels, actors, and interactions. \\
Infer Implicit Meaning & Identify implicit assumptions embedded in diagrams or explanations. \\
Link Elements To Theme & Select descriptive words that align with specific visual features. \\
Tone Consistency & Ensure language aligns with the formal or thematic nature of the subject. \\
Use Imagery And Metaphor & Combine visual analysis with context for symbolic interpretation. \\
Validate Consistency And Completeness & Cross-reference visual elements with textual data to confirm entities. \\
\midrule
\multicolumn{2}{@{}l}{\textit{Chart \& OCR}} \\
Aggregate Handling & Differentiate between group data and specific subgroup data. \\
Align Output With Instruction Format & Deduce if numerical or categorical answers are expected. \\
Assess Data Relevance And Availability & Recognize when data lacks context to answer a question directly. \\
Axis Analysis & Analyze chart axes to determine variables and their relationships. \\
Chart Legend Interpretation & Analyze legends to determine how visual elements encode info. \\
Consider Alternative Interpretations & Re-evaluate requirements when there is a data mismatch. \\
Consider Data Granularity & Weigh label granularity against the need for precision. \\
Consistency Assessment & Verify if visual attributes are applied consistently across regions. \\
Criteria Reinterpretation & Restate problems in actionable terms (e.g., "least change" as "flattest"). \\
Eliminate Irrelevant Information & Exclude entries that do not correspond to valid target categories. \\
Exhaustive Verification & Cross-validate answers with the explicit intent of the question. \\
Extract Numerical Information From Visuals & Estimate or read values directly from labels on charts/graphs. \\
Extract Relevant Attribute From Data & Identify connections matching specific attributes like color or type. \\
Extract Textual Data & Extract info from text sources without additional calculation. \\
Final Verification & Re-examine prompts to ensure solutions align with requirements. \\
Infer From Context & Use context to infer expected trends (e.g., survival phases vs scarcity). \\
Locate Relevant Section By Heading & Navigate visual resources using identifiers to focus attention. \\
Mapping & Map visual arrangements to categories and time periods. \\
Option Selection Based On Criteria & Apply distinguishing steps to select a single candidate from many. \\
Prioritize Explicit Information Over Inference & Differentiate between explicit data and assumed background knowledge. \\
Prioritize Explicit Labels Over Visual Cues & Favor directly labeled terms over interchangeable associated terms. \\
Scope Constraint & Clarify inclusion/exclusion based on roles (reference vs data points). \\
Targeted Information Search & Restate requirements to maintain focus on relevant data subsets. \\
Validate Data Type Consistency & Match interpretation to data type (e.g., using \% if only \% available). \\
Validate Task Goal & Cross-check annotations against task requirements for correctness. \\
Value Matching & Compare values to a target to assess proximity or criteria match. \\
Visual Data Interpretation & Infer magnitude by comparing spatial positions on a graph. \\
Visual Label Association & Check if labels are directly associated or provided via a legend. \\
\midrule
\multicolumn{2}{@{}l}{\textit{Knowledge \& Recognition}} \\
Analyze Compositional Elements & Break down visual components to understand scene and context. \\
Analyze Image Structures & Evaluate structures to determine orientation and standard features. \\
Analyze Visual Layout & Observe spatial relationships to understand object connections. \\
Assess Information Availability & Indicate when external knowledge is required for a definitive answer. \\
Assess Relevance & Identify environmental elements relevant to context or purpose. \\
Associate Visual Elements With Meaning & Link objects to real-world usage to infer setting themes. \\
Balance Precision With Generality & Provide accurate answers that account for regional variability. \\
Causal Reasoning & Link symptoms to underlying causes based on known patterns. \\
Compare Elements & Compare characteristics to determine similarities and differences. \\
Conciseness Optimization & Include relevant info without exceeding word limits. \\
Conclude Based On Insufficient Data & Exclude conclusions not supported by available evidence. \\
Contextual Analysis & Deducing a person's role from actions and environment. \\
Disambiguation & Clarify meanings of terms used in misleading or unclear ways. \\
Elimination & Rejecting interpretations that do not align with evidence. \\
Hypothesis Testing & Test subject identity by evaluating traits against candidates. \\
Identify Structural Patterns & Analyze shape and posture to distinguish object categories. \\
Identify Visual Focus & Analyze element positions to determine prominence or relevance. \\
Infer Context From Visual And Contextual Clues & Analyze elements like street signs to understand scenes. \\
Infer From Common Conventions & Use typical structures (e.g., posters) to infer missing info. \\
Interpret Lighting And Atmosphere & Analyze light and environmental cues to infer time of day. \\
Interpret Question Intent & Adjust understanding based on phrasing to align with intent. \\
Iterative Refinement & Propose and revise hypotheses based on new observations. \\
Iterative Verification & Establish confidence by repeatedly verifying data via checks. \\
Label Verification & Match icons and labels to process steps based on position. \\
Make Informed Assumption & Use real-world knowledge when direct evidence is absent. \\
Pattern Recognition & Identify context by recognizing familiar terminology and structures. \\
Prioritize Evidence Based Decision & Select plausible answers based on cumulative visual analysis. \\
Synthesize Information & Combine visual and text info for a coherent conclusion. \\
Use Domain Knowledge & Apply specific expertise to verify structures against theory. \\
Visual Analysis & Establish visual criteria to determine if elements meet tasks. \\
Visual Comparison & Use cues like bottle height to infer relative volume. \\
\bottomrule
\end{tabularx}
\end{minipage}
\hfill
\begin{minipage}[t]{0.49\linewidth}
\vspace{0pt}
\begin{tabularx}{\linewidth}{@{}>{\raggedright\arraybackslash}p{0.29\linewidth}>{\raggedright\arraybackslash}X@{}}
\toprule
Skill & Definition \\
\midrule
\multicolumn{2}{@{}l}{\textit{Grounding, Counting \& Search}} \\
Analyze Structural Elements & Examines object structure to determine composition. \\
Assess For Trick Question & Recognize riddles or tricks via phrasing/data inconsistencies. \\
Assess Visual Clarity & Recognize limitations of visual evidence and detail. \\
Assess Visual Indicators & Draw conclusions based on the absence of specific indicators. \\
Associative Mapping & Associate text/marks with functional roles like brand names. \\
Assume Standard Conventions & Assume standard response forms when multiple interpretations exist. \\
Assume Standard Dimensions & Account for device scaling by estimating relative coordinates. \\
Check For Duplicates & Verify that an element is the only one matching a requirement. \\
Color Analysis & Analyze color properties to determine object relationships. \\
Concluding From Evidence & Draw final conclusions based on cumulative observed evidence. \\
Cross Reference & Cross-reference command language with UI labels to identify elements. \\
Determine Pixel Coordinates & Estimate coordinates based on spreadsheet layout and row headers. \\
Estimate Position & Estimate positions relative to surrounding components. \\
Extract Contextual Information & Identify and extract contextual relationship between text elements. \\
Final Confirmation & Confirm interpretation by eliminating contradictions with requirements. \\
Match To Options & Select options most consistent with evidence despite limitations. \\
Output In Required Format & Translate element locations into structured output like JSON. \\
Prioritize Clear Evidence & Exclude elements that cannot be reliably determined. \\
Reasoned Assumption & Make assumptions to resolve visual ambiguities. \\
Select Option & Select the best-matching object even when multiple candidates exist. \\
Spatial Analysis & Determine roles in a scene based on action and environment. \\
Synthesize Visual And Contextual Information & Combine observations into a coherent, supported explanation. \\
Visual Estimation & Include partially visible elements in counts if identifiable. \\
Visual Scanning & Divide images into quadrants to check for target objects. \\
Visual Segmentation & Divide fields into spatial regions to identify/count elements. \\
\midrule
\multicolumn{2}{@{}l}{\textit{STEM}} \\
Apply Domain Knowledge & Apply field-specific knowledge when context is missing. \\
Assume Missing Data & Make reasonable assumptions for ambiguous parameters in calculations. \\
Avoid Unnecessary Complexity & Adopt symmetric/extreme arrangements to simplify reasoning. \\
Consider Alternative Solutions & Evaluate various theorems to determine relevance to a problem. \\
Consistency Verification & Verify derived values against original problem constraints. \\
Constraint Handling & Evaluate if options adhere to structural/quantitative constraints. \\
Contextual Inference & Infer answers by evaluating outcomes of element comparisons. \\
Cross Validation & Identify errors by recomputing values and comparing results. \\
Diagram Analysis & Reevaluate geometric relationships to ensure correct properties. \\
Distractor Detection & Identify and disregard elements not required by constraints. \\
Exhaustive Option Verification & Verify if values result in duplicate elements in a set. \\
Filter Irrelevant Information & Disregard unnecessary parameters to focus on essentials. \\
Infer Meaning From Context & Determine task nature by analyzing materials and instructions. \\
Infer Structural Relationships & Deduce logical groupings by analyzing patterns and connections. \\
Iterative Hypothesis Testing & Test alternative hypotheses about overlapping elements. \\
Justify Answer With Evidence & Confirm answers by checking consistency across reasoning lines. \\
Manage Uncertainty And Assumptions & Avoid definitive claims when evidence is ambiguous. \\
Option Alignment Evaluation & Assess if choices relate directly to the main subject. \\
Prioritize Clarity And Relevance & Weigh options supported by explicit actions in the image. \\
Prioritize Prominent Visual Element & Focus on primary focal points rather than background elements. \\
Process Of Elimination & Eliminate choices whose definitions do not match context. \\
Question Intent Analysis & Clarify intent to match reasoning strategy to the question. \\
Select Best Fit Option & Choose the choice numerically closest to a calculated value. \\
Select Valid Solution Based On Constraints & Eliminate solutions that do not make sense (e.g. negative angles). \\
Spatial Relationship Analysis & Infer spatial relationships using common layout conventions. \\
Structural Analysis & Recognize functional groups via atom bonding arrangements. \\
Synthesize Multiple Clues & Classify objects by synthesizing multiple distinctive features. \\
Visual Feature Identification & Associate visual features with traits of known categories. \\
Visual Verification & Integrate visual and text info to resolve ambiguities. \\
\midrule
\multicolumn{2}{@{}l}{\textit{Spatial \& Action}} \\
Classify Elements By Type & Determine action eligibility based on position or status. \\
Consider Alternative Hypotheses & Evaluate alternative interpretations when outcomes do not match. \\
Consider Reference & Focus on specific reference objects for correct comparisons. \\
Eliminate Invalid Options & Narrow locations by identifying occupied rows/columns. \\
Enumerate Elements & List all elements in a specific position within an arrangement. \\
Evaluate Candidate Positions & Select positions fulfilling functional requirements of patterns. \\
Evidence Based Conclusion & Acknowledge when info is insufficient for spatial determination. \\
Focus On Salient Features & Identify prominent elements to determine importance. \\
Foreground Background Analysis & Apply layering principles to determine relative positions. \\
Handle Missing Information & Deduce region assignments based on elimination/requirements. \\
Identify Categories & Identify element categories for further classification. \\
Interpret Ambiguity & Clarify ambiguous terms using common conventions. \\
Label Handling & Adopt provided labels even if personal inference differs. \\
Map Observations To Answer Choices & Translate structural observations into required solution formats. \\
Mental Simulation & Simulate moves to assess if they achieve objectives. \\
Object Identity Differentiation & Use comparative statements to assess opposite relationships. \\
Perspective Analysis & Express relationships using different reference frames. \\
Plausibility Evaluation & Evaluate scenario plausibility based on environment constraints. \\
Question Scope Management & Align question scope with relevant visual entities. \\
Reexamine And Verify & Backtrack and re-execute simulations upon detecting errors. \\
Rule Application & Compare outcomes against constraints to select approaches. \\
Systematic Visual Scanning & Scan fields in order (clockwise/rows) to prevent double-counting. \\
Translate Spatial Relationship To Directional Language & Interpret terms like "above" based on visual composition. \\
\bottomrule
\end{tabularx}
\end{minipage}
\endgroup
\caption{\textbf{Summary of skill definitions.} A comprehensive glossary providing operational definitions for the fine-grained skills evaluated across all six task categories.}
\label{tab:skills_def}
\end{table}

\section{Reasoning Traces}
\label{app:model-examples}

We provide extensive reasoning trace examples from \model{} across all six task categories on our project website: \url{https://vero-reasoning.github.io}. Specifically, we showcase three representative examples per category to concretely illustrate how the model adapts its cognitive approach to different domains. These traces exhibit structured chain-of-thought reasoning with dynamic metacognitive behaviors, such as self-verification and backtracking, alongside precisely grounded visual perception. 

Beyond merely highlighting these task-specific strategies, the qualitative examples validate the overarching efficacy of our open multi-task RL recipe. We demonstrate how training on a highly diverse, mixed-domain curriculum equips the model with a versatile toolkit for general-purpose visual reasoning. Whether navigating complex spatial constraints or performing abstract STEM deduction, our examples confirm that \model{} seamlessly bridges raw visual inputs with long-horizon logical planning, fundamentally enhancing its capacity to handle real-world multimodal challenges.

\end{document}